\documentclass{article}

\PassOptionsToPackage{numbers, compress}{natbib}
\usepackage{graphicx}
\usepackage[preprint]{neurips_2026}

\usepackage[utf8]{inputenc}                                 % allow utf-8 input
\usepackage[T1]{fontenc}                                    % use 8-bit T1 fonts
\usepackage{hyperref}                                       % hyperlinks
\usepackage{xcolor}                                         % colors
\definecolor{linkblue}{HTML}{1565C0}                        % Define modern blue for links
\hypersetup{                                                % Use blue links instead of green boxes
  colorlinks=true,
  linkcolor=linkblue,
  citecolor=linkblue,
  urlcolor=linkblue,
}
\usepackage{url}                                            % simple URL typesetting
\usepackage{booktabs}                                       % professional-quality tables
\usepackage{amsmath}                                        % math environments and \text, \eqref
\usepackage{amsfonts}                                       % blackboard math symbols
\usepackage[nameinlink]{cleveref}                           % references (must come after amsmath)
\newcommand{\crefpanel}[2]{\hyperref[#1]{\Cref*{#1}#2}}     % Allow custom postfix in link, e.g. Figure 6 (a)
\usepackage{nicefrac}                                       % compact symbols for 1/2, etc.
\usepackage{microtype}                                      % microtypography

\usepackage{float}                                          % [H] placement for figures
\usepackage{wrapfig}                                        % figures beside running text (ablation wrap)
\usepackage{duckuments}                                     % example-image-duck placeholder
\usepackage{tikz}
\usepackage{pgfplots}
\pgfplotsset{compat=1.18}
\usepackage{multirow}
\usepackage{placeins}
\usepackage[labelfont=bf]{caption}

\title{Segment to Focus: Guiding Latent Action Models \\ in the Presence of Distractors}

\author{%
  Marcus Fechner\thanks{Equal contribution.}~~\thanks{Correspondence to \texttt{marcus.fechner@kit.edu}.} \\
  {\small Karlsruhe Institute of Technology} \\
  {\small \texttt{marcus.fechner@kit.edu}} \\
  \And
  Hamza Adnan\footnotemark[1] \\
  {\small University of Oxford} \\
  {\small \texttt{hamzaadnan268@outlook.com}} \\
  \And
  Constantin C. Lüth \\
  {\small Karlsruhe Institute of Technology} \\
  {\small \texttt{info@constantin-lueth.de}} \\
  \AND
  Matthew T. Jackson \\
  {\small University of Oxford} \\
  {\small \texttt{jackson@robots.ox.ac.uk}} \\
  \And
  Alexey Zakharov \\
  {\small University of Oxford} \\
  {\small \texttt{alexey.zakharov@cs.ox.ac.uk}} \\
  \And
  J.~Marius Zöllner \\
  {\small Karlsruhe Institute of Technology} \\
  {\small \texttt{marius.zoellner@kit.edu}} \\
}

\begin{document}

\maketitle

% Abstract
% Abstract
% Combination of Alex's and Marcus's abstract versions
\begin{abstract}
  Latent action models (LAMs) offer a promising path to pre-training embodied agents on large amounts of
  action-free video. They infer latent actions between consecutive observations that can later be decoded to
  ground-truth actions using a small number of labels. However, recent work has shown that this recipe fails in the presence of action-correlated visual distractors common
  in real-world video, such as dynamic backgrounds, camera shake, or other moving objects. In these
  scenarios, the standard reconstruction objective drives latent actions to encode exogenous motion instead of
  agent-controlled dynamics, resulting in policies that underperform when fine-tuned.
  We observe, however, that endogenous and exogenous factors are typically spatially separated in pixel space:
  control-relevant change is concentrated on the agent, while distractor motion occurs elsewhere.
  We exploit this observation by restricting the reconstruction objective to agent pixels, forcing latent
  actions to explain agent-controlled dynamics rather than exogenous ones. We call this method
  \textbf{MaskLAM}; it obtains the agent mask zero-shot from off-the-shelf segmentation foundation models
  (e.g., SAM) and requires no architectural changes, auxiliary losses, or action labels during pre-training. Across two continuous-control benchmarks (Distracting Control Suite, Distracting Meta-World), MaskLAM reduces normalized linear-probe MSE by up to $3.51\times$ and improves normalized return by up to $4.97\times$ over LAPO, while narrowing the gap to LAOM-Labels, which relies on ground-truth action supervision.

\end{abstract}

\begin{figure}[h]
    \centering
    \includegraphics[width=\linewidth, trim=0cm 0cm 0cm 0cm, clip]{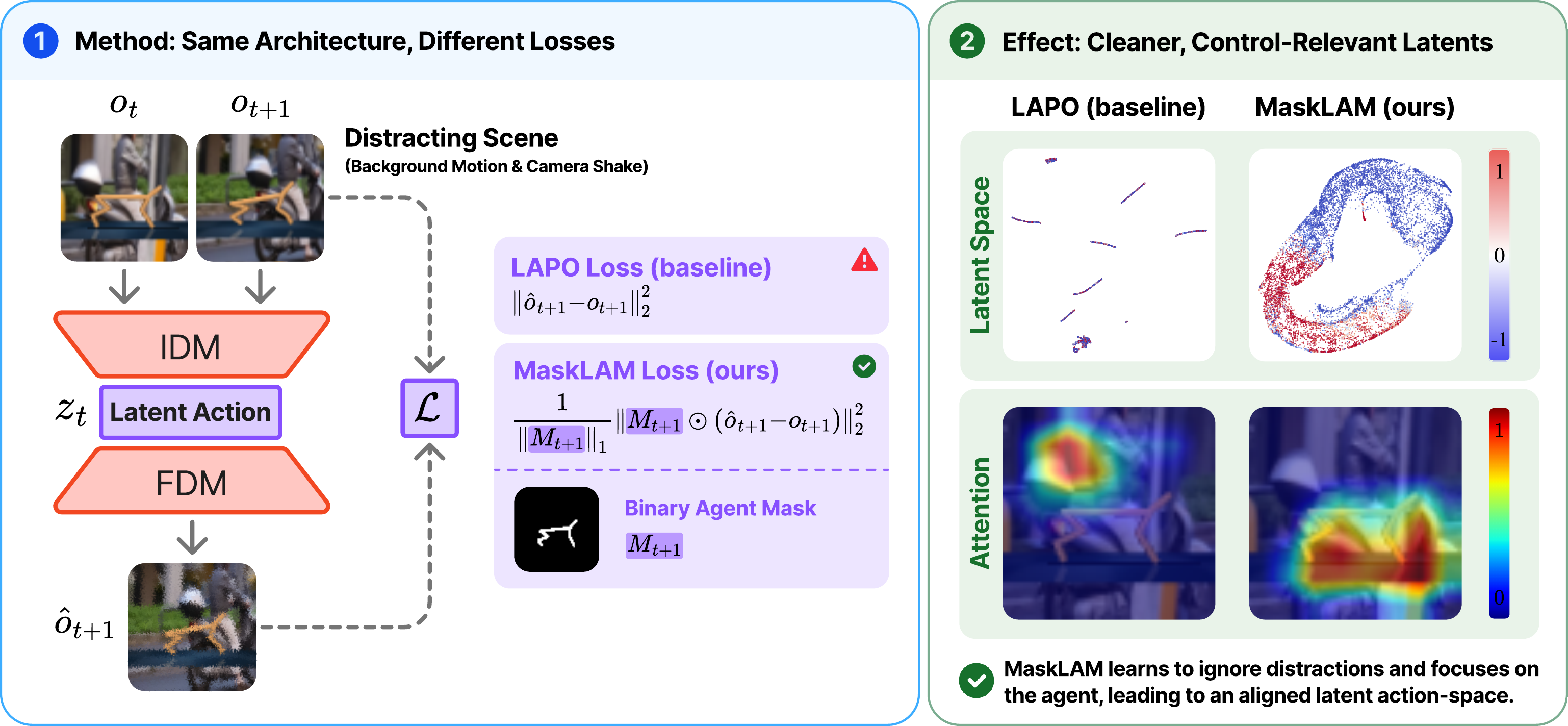}
    \vspace*{-\baselineskip}
    \caption{\textbf{Overview of MaskLAM.} Latent action models like LAPO~\citep{schmidt_learning_2023} infer a latent action $z_t$ from
    observation-only video by jointly training an inverse (IDM) and forward dynamics model (FDM).
    Under visual distractors, LAPO's reconstruction loss pressures $z_t$ to encode exogenous variance and downstream policies
    fail~\citep{nikulin_latent_2025}. Since endogenous and exogenous factors are typically spatially disjoint in
    pixel space, MaskLAM restricts the loss to agent pixels via a binary mask $M_{t+1}$ obtained zero-shot
    from SAM~2.1~\citep{ravi_sam_2024} (1), so distractor pixels contribute no gradient to $z_t$. (2)~On
    cheetah-run under distractors, the UMAP of the latent action space collapses into entangled clusters for LAPO but forms a smooth, monotonic manifold for MaskLAM
    (top); Eigen-CAM \citep{muhammad_eigen-cam_2020} saliency shifts from the background distractors to the agent (bottom). The intervention adds no
    auxiliary losses, action labels, or architectural changes to LAPO.}
    \label{fig:teaser}
\end{figure}

% Introduction
\section{Introduction}
\label{sec:introduction}

The scalability of reinforcement and imitation learning~\citep{argall_survey_2009, levine_end--end_2016} is heavily
limited by the need for action-labeled data. Although the internet contains an abundance of unlabeled video
demonstrations, leveraging this resource is still an unsolved problem when ground-truth control annotations are missing.
Latent Action Models (LAMs)~\citep{edwards_imitating_2019, schmidt_learning_2023} have emerged as a promising way
to overcome this bottleneck: by jointly training an inverse and forward dynamics model on observation-only video, they infer a
\textit{latent action} between successive frames, and a lightweight decoder maps these to actual controls using only a small set of labeled examples. On simple benchmarks (static backgrounds and a single moving agent) this approach works consistently well.

Training robust LAMs in realistic, visually-complex environments remains difficult. Standard LAMs rely on
\textit{global} reconstruction objectives that force the model to encode the entire scene to predict future frames.
With moving backgrounds, camera shake, or additional moving agents, pixel reconstruction has no reason to prefer
agent dynamics over background variation, and the latent encodes whatever varies most~\citep{zhang_what_2025}. The
Exogenous Block MDP (Ex-BMDP) framework~\citep{efroni_provably_2021} formalizes this failure mode: observations
entangle a control-endogenous factor (driven by agent actions) with a control-exogenous factor (evolving independently
of the agent), and standard reconstruction losses pressure the latent to encode both. As a result, the latent action
space of LAPO~\citep{schmidt_learning_2023}, the basis for many subsequent LAMs, no longer recovers the true action, and downstream
policies fail~\citep{nikulin_latent_2025}. Prior work addresses this
through architectural or modeling interventions: information bottlenecks~\citep{ye_become_2022}, improved feature
extractors~\citep{klepach_object-centric_2026}, auxiliary losses~\citep{nikulin_latent_2025, bu_laof_2026}, or explicit
distractor models~\citep{wang_ad3_2024}, at the cost of model complexity, action-label dependence, or extra
hyperparameters.

Our key observation is that the Ex-BMDP factorization, although abstract in latent space, is predominantly
\textit{spatial} in pixel space: agent actions have immediate, local effect on the \textit{agent's} pixels, while
exogenous changes are spatially disjoint from them. Pre-trained segmentation foundation
models~\citep{ravi_sam_2024} localize agent pixels zero-shot from a single prompt. Combining the two yields
\textbf{MaskLAM}: we leave the LAPO recipe unchanged and restrict the forward-dynamics reconstruction loss to agent
pixels. Distractor pixels contribute no gradient, so the latent action receives signal only from agent-controlled
dynamics. The intervention adds no auxiliary losses, requires no action labels during pre-training, and uses SAM's
prompt interface to specify the agent of interest.

Our contributions are:

\begin{itemize}
    \item We identify spatial separability of agent and distractor pixels as the property that turns the abstract
          Ex-BMDP factorization into a concrete training-time intervention, and propose \textbf{MaskLAM}, a lightweight
          modification of the LAPO objective that eliminates exogenous gradient on the latent action.
    \item Across 14 tasks on two benchmarks (Distracting Control Suite, Distracting Meta-World),
          MaskLAM reduces normalized linear-probe MSE by up to $3.51\times$ and improves normalized return by up
          to $4.97\times$ over LAPO, narrowing the gap to LAOM-Labels~\citep{nikulin_latent_2025}, a recent LAM that
          tackles the same distractor problem by relying on privileged action supervision during pre-training.
    \item We show that MaskLAM admits compact latent action spaces, lower action-label budgets for
          downstream decoding, and degrades gracefully under common segmentation failure modes.
\end{itemize}

% Related Work
\section{Related work}
\label{sec:related_work}

\textbf{Latent action models.} Latent action models (LAMs) aim to recover action-relevant representations from unlabeled state transitions. Early work by ILPO~\citep{edwards_imitating_2019} learned discrete latent actions through a forward dynamics objective, but mode collapse and poor scalability~\citep{struckmeier2023preventing} limited its reach. LAPO~\citep{schmidt_learning_2023} resolved both issues by jointly training an inverse and forward dynamics model, and the resulting framework now underpins large-scale robot manipulation pre-training~\citep{chen2025moto, ye_latent_2024} and interactive world models~\citep{bruce2024genie}. This progress, however, rests on a fragile assumption: that state transitions are fully explained by agent actions. Once visual distractors enter the scene, this assumption breaks and reconstruction-based LAMs entangle agent dynamics with irrelevant scene variation.

To address this challenge, prior work has focused on architectural and modeling interventions, including improved feature extractors that decompose scenes into object slots~\citep{klepach_object-centric_2026, locatello2020object}, auxiliary losses such as action-decoding probes~\citep{nikulin_latent_2025} or masked optical flow consistency~\citep{bu_laof_2026}, and explicit distractor models such as AD3~\citep{wang_ad3_2024} that factorize the world into agent- and distractor-controlled components. While these methods offer improvements, they come at the cost of increased model complexity, action label dependence, or extra hyperparameters, all requiring substantial modifications to the underlying LAM. The closest of them to our work is LAOF~\citep{bu_laof_2026}, which also leverages segmentation masks, but applies them to an auxiliary flow-prediction loss rather than to the LAM's reconstruction objective.

\textbf{Our approach.} MaskLAM keeps the LAM architecture and training pipeline intact, modifying only the
reconstruction objective: errors are reweighted by agent-centric segmentation masks, suppressing gradients from
distractor regions. Unlike LAOM-Labels, we require no action labels. In contrast to LAOF, the most similar earlier approach that also
uses masks, our method applies masks directly to the reconstruction objective rather than on a separate flow-prediction loss.
Unlike AD3, it does not rely on any independence assumptions between the agent and distractor dynamics. Unlike object-centric LAMs
that mask the inputs, our masks are applied to the loss, thereby preserving the Ex-BMDP formulation (\Cref{sec:background}) and
removing the need for slot probing during evaluation.

% Background and Problem Formulation
\section{Background and Problem Formulation}
\label{sec:background}

\textbf{Exogenous Block MDP.} We adopt the Ex-BMDP framework~\citep{efroni_provably_2021} as the formal foundation
for reasoning about distractors. Each observation $o_t \sim Q(\cdot \mid s_t, e_t)$ entangles a
\textit{control-endogenous} state $s_t$, governed by agent actions $a_t \in \mathbb{R}^{d_a}$ via $s_{t+1} \sim T(\cdot \mid s_t, a_t)$, and a
\textit{control-exogenous} state $e_t$, evolving independently via $e_{t+1} \sim T_e(\cdot \mid e_t)$. The goal of
latent action learning is to recover a latent action $z_t \in \mathbb{R}^{d_z}$ that captures only the endogenous transition $s_t \to
s_{t+1}$, encoding nothing about $e_t \to e_{t+1}$.

\textbf{Latent action models.} Most reinforcement and imitation learning methods require action-labeled trajectories
$\tau_n = \{(o_i^n, a_i^n)\}_{i=1}^{T}$, but the vast majority of video data on the internet provides only
observation-only trajectories $\tau_n = \{o_i^n\}_{i=1}^{T}$. LAMs~\citep{edwards_imitating_2019,
schmidt_learning_2023} bridge this gap by inferring latent actions $z_i$ from observation-only data, yielding
$\mathcal{D}_z = \{(o_i, z_i)\}$ for downstream policy learning. We build on LAPO~\citep{schmidt_learning_2023}, the basis for many subsequent LAMs~\citep{hu_video_2025, cui_play_2022, ye_latent_2024}. LAPO jointly trains an \textit{inverse dynamics model}
(IDM) $z_t \sim p_{\text{IDM}}(\cdot \mid o_t, o_{t+1})$ and a \textit{forward dynamics model} (FDM) $\hat{o}_{t+1}
\sim p_{\text{FDM}}(\cdot \mid o_t, z_t)$ to minimize the next-observation prediction loss:
\begin{equation}
	\mathcal{L}_{\text{FDM}}^{\text{LAPO}} = \mathbb{E}\left[\| \hat{o}_{t+1} - o_{t+1} \|_2^2\right].
	\label{eq:lapo_loss}
\end{equation}
The latent acts as an information bottleneck: since both models observe $o_t$ but only the IDM sees $o_{t+1}$, the
IDM is forced to compress the difference between consecutive frames into $z_t$. In distractor-free settings, this
difference is most efficiently explained by the agent's true actions~\citep{shah_pitfalls_2020}.

\textbf{Latent action learning pipeline.} The standard pipeline~\citep{schmidt_learning_2023} proceeds in three stages:
\begin{itemize}
    \item \textit{Stage~1} (LAM pre-training): train IDM and FDM on the full unlabeled dataset to learn latent actions.
    \item \textit{Stage~2} (relabeling and behavior cloning): use the trained IDM to assign latent actions to all
        transitions, producing $\mathcal{D}_z = \{(o_i, z_i)\}$, and train a latent policy
        $\pi_{\text{latent}}(z_t \mid o_t)$.
    \item \textit{Stage~3} (action decoder fine-tuning): using a small set of labeled trajectories, train a decoder
        $d\colon \mathcal{Z} \to \mathcal{A}$ that maps from latent to ground-truth actions. The deployable policy is
        $\pi = d \circ \pi_{\text{latent}}\colon \mathcal{O} \to \mathcal{A}$.
\end{itemize}

\textbf{Why LAPO fails under distractors.} Since $o_{t+1} \sim Q(\cdot \mid s_{t+1}, e_{t+1})$, the FDM in
\eqref{eq:lapo_loss} must predict changes from \textit{both} the endogenous and exogenous transitions. Given $o_t$,
$z_t$ is the only varying input to the FDM, so the optimization pressures it to encode exogenous dynamics to reduce
prediction error on distractor pixels. \citet{nikulin_latent_2025} show that this contamination persists even with a
multi-step IDM, larger latents, latent-space FDM, and augmentations, and that ground-truth action supervision during
LAM training can bridge the gap; this reintroduces the action-label dependence that LAMs were designed to eliminate.

\textbf{Problem statement.} We seek a method that restores latent action quality in the presence of visual distractors
\textit{without} requiring ground-truth action supervision during LAM pre-training, and without modifying the LAM
architecture.
% Method
\section{Method}
\label{sec:method}

\begin{figure}[H]
    \centering
    \includegraphics[width=\linewidth, trim=4cm 4.5cm 4cm 4cm, clip]{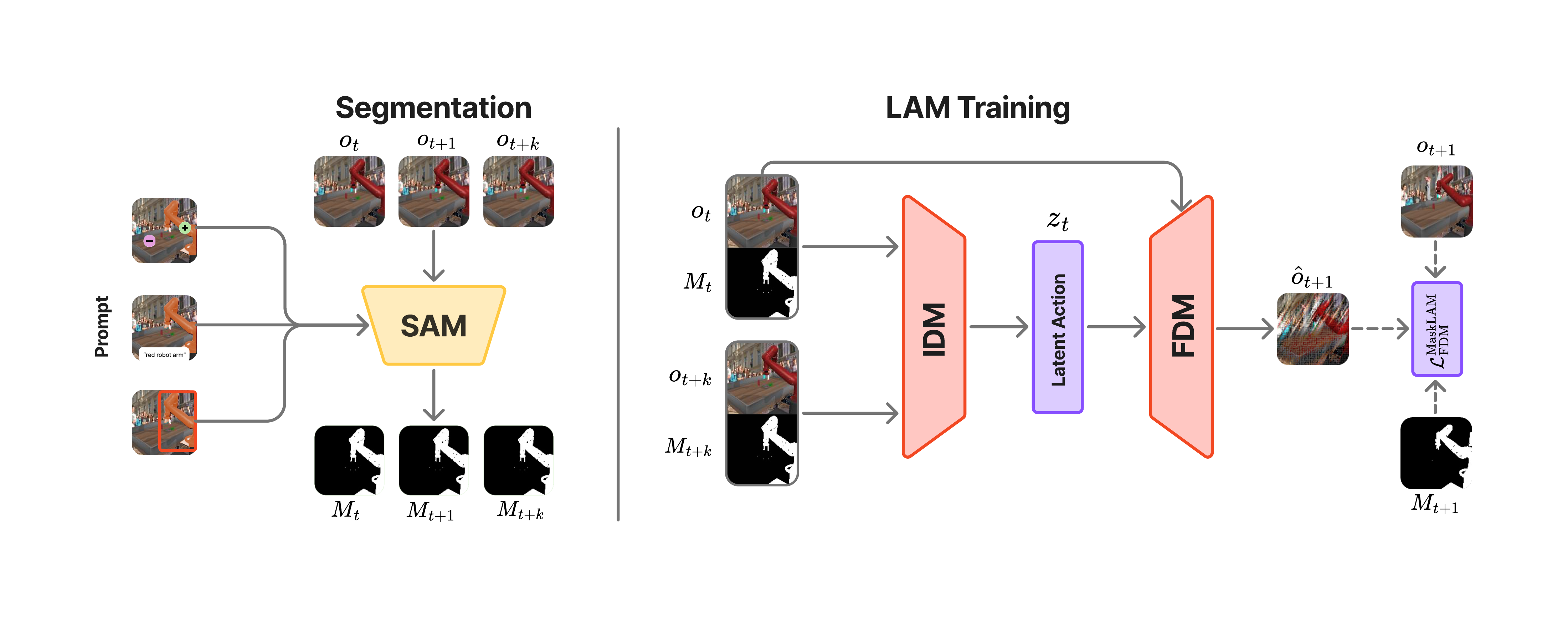}
    \vspace*{-\baselineskip}
    \caption{\textbf{Simplified MaskLAM stage~1 pipeline.} Conditioned on a target agent prompt, a pre-trained SAM~2.1 model extracts
    agent masks $M_t, M_{t+1}, M_{t+k}$ from observations $o_t, o_{t+1}, o_{t+k}$. The IDM consumes
    $(o_t, o_{t+k}, M_t, M_{t+k})$ and predicts the latent action $z_t$; the FDM then reconstructs $\hat{o}_{t+1}$
    from $(o_t, z_t, M_t)$. The forward dynamics loss is spatially restricted to agent pixels (white) via $M_{t+1}$,
    so distractor pixels (black) contribute no gradient. Endogenous prediction errors thus shape $z_t$, while
    exogenous dynamics are filtered out at the loss rather than the encoder. See \Cref{sec:method}.}
    \label{fig:framework}
\end{figure}

As shown in \Cref{sec:background}, the FDM loss in \eqref{eq:lapo_loss} reconstructs the full observation, forcing
$z_t$ to encode endogenous and exogenous dynamics.
Prior work addresses the contamination of $z_t$ at the representation level: action supervision to steer the
encoder~\citep{nikulin_latent_2025}, object-centric slot decomposition~\citep{klepach_object-centric_2026}, or
auxiliary optical flow losses~\citep{bu_laof_2026}. MaskLAM takes a different route. Instead of modifying the encoder
or adding auxiliary losses, we modify the \textit{prediction target} to exclude exogenous dynamics from the learning
signal. The key observation is that in pixel space, the Ex-BMDP factorization is often \textit{spatial}:
endogenous changes correspond to agent pixels, exogenous changes to distractor pixels. Prior methods separate
endogenous from exogenous in latent space~\citep{efroni_provably_2021, lamb_guaranteed_2022,
klepach_object-centric_2026, bu_laof_2026}; MaskLAM performs this separation in pixel space, where pre-trained
segmentation models provide a strong, steerable prior.

\textbf{Spatial separability assumption.} Let $M_t \in \{0,1\}^{H \times W}$ be a binary segmentation mask identifying
agent pixels, obtained from a pre-trained segmentation model. MaskLAM assumes the Ex-BMDP factorization is
approximately spatially separable:
\begin{equation}
    M_t \odot o_t \approx M_t \odot Q_s(\cdot \mid s_t), \quad
    (1 - M_t) \odot o_t \approx (1 - M_t) \odot Q_e(\cdot \mid e_t),
    \label{eq:spatial_sep}
\end{equation}
where $\odot$ denotes the Hadamard (element-wise) product, and $Q_s$ and $Q_e$ denote the endogenous and exogenous
components of the emission function. This holds under a \emph{visual independence} assumption: the rendering of
pixels within $M_t$ must be governed mainly by $s_t$, independent of $e_t$. Spatial overlap between agent and
distractor is allowed.

\textbf{Masked forward dynamics loss.} Given the spatial separability assumption, the most direct intervention is to
restrict the FDM loss to agent pixels:
\begin{equation}
    \mathcal{L}_{\text{FDM}}^{\text{MaskLAM}}
    = \frac{1}{\|M_{t+1}\|_1}\,\| M_{t+1} \odot (\hat{o}_{t+1} - o_{t+1}) \|_2^2,
    \label{eq:MaskLAM_loss}
\end{equation}
where $M_{t+1}$ is the binary mask indicating agent occupancy in the target frame. The loss is zero on distractor
pixels, and normalization by $\|M_{t+1}\|_1$ keeps gradient magnitude consistent across varying mask sizes. To see
why this eliminates exogenous pressure on $z_t$, consider the gradient:
\begin{equation}
    \frac{\partial \mathcal{L}_{\text{FDM}}^{\text{MaskLAM}}}{\partial z_t}
    = \frac{2}{\|M_{t+1}\|_1}
      \left(M_{t+1} \odot (\hat{o}_{t+1} - o_{t+1})\right)^\top
      \frac{\partial \hat{o}_{t+1}}{\partial z_t}.
    \label{eq:MaskLAM_gradient}
\end{equation}
Where $M_{t+1} = 0$, the residual is zeroed before propagating into $z_t$. Since encoding $e_t \to e_{t+1}$ cannot
reduce $\mathcal{L}_{\text{FDM}}^{\text{MaskLAM}}$, the optimization no longer pressures $z_t$ to capture exogenous
dynamics.

\textbf{Mask-conditioned IDM and FDM.} We concatenate the segmentation mask as an extra input channel:
the IDM receives $(o_t, o_{t+k}, M_t, M_{t+k})$ and the FDM receives $(o_t, z_t, M_t)$. The mask channel is a
\textit{hint}, not a hard gate: the models still receive unmasked observations and can represent the full scene
context, including objects the agent interacts with that fall outside $M_t$ (see \Cref{fig:eigencam_cheetah_frame_0}).
In principle, conditioning the IDM on the mask also enables multi-agent control, with different masks yielding
separate latent action streams from the same observation sequence. We leave this to future work.

\textbf{Why loss masking, not input masking.} Feeding $M_t \odot o_t$ instead of $o_t$ aggressively removes exogenous
information but also destroys endogenous context: the endogenous state $s_t$ includes objects the agent interacts
with (a gripper approaching a cup, a hand holding a tool); these are action-dependent but may fall outside $M_t$.
Enumerating such objects a priori is impractical, whereas identifying the agent itself is straightforward with
pre-trained segmentation. Loss-only masking exploits this asymmetry: the encoder receives full observations and is
free to discover the remaining endogenous context from data, while $z_t$ is shaped only by endogenous prediction
errors. We evaluate this design choice in \Cref{sec:ablation}.

\textbf{Implementation} (\Cref{fig:framework})\textbf{.} Following prior work~\citep{nikulin_latent_2025, lamb_guaranteed_2022, levine_multistep_2024}, we use a
multi-step IDM $z_t \sim p_{\text{IDM}}(\cdot \mid o_t, o_{t+k})$ with $k$ sampled uniformly from $\{1, \ldots, 10\}$
during training and $k=1$ at inference, since exogenous noise decorrelates over longer
horizons~\citep{lamb_guaranteed_2022, efroni_provably_2021}. As in \citet{nikulin_latent_2025}, we stack 3
consecutive observations as input to both the IDM and FDM. Masks are obtained zero-shot from SAM~2.1
(hiera-tiny)~\citep{ravi_sam_2024} given a bounding box in the first frame
(see \Cref{sec:mask_generation}), computed in $\sim$10\,ms per frame on
an A100 and pre-computed for the dataset before LAM training. We use the continuous LAPO baseline from 
\citet{nikulin_latent_2025} (IMPALA ResNet encoder~\citep{espeholt_impala_2018}, ResNet
decoder~\citep{he_deep_2016}).

% Experiments
\section{Experiment Setup}
\label{sec:experiments}

We evaluate the masked forward dynamics loss \eqref{eq:MaskLAM_loss} along three axes: latent action quality,
downstream policy performance, and robustness to common segmentation failures (occlusions, imperfect masks).

\textbf{Environments.} Following prior work on latent actions~\citep{nikulin_latent_2025, klepach_object-centric_2026},
we evaluate on two benchmarks.
\begin{itemize}
    \item  \emph{Distracting Control Suite} (DCS)~\citep{stone_distracting_2021}: augments DeepMind Control
           Suite~\citep{tunyasuvunakool_dm_control_2020} with dynamic natural video backgrounds (DAVIS
           dataset~\citep{pont-tuset_2017_2018}), continuously shifting camera pose, and randomized agent and object
           colors. We use the \emph{easy} dynamic setting (magnitude 0.1, 4 background videos, all distractors
           changing smoothly per timestep). Note that color randomization affects the agent itself, so distractors
           are not fully spatially separable. We evaluate four continuous-control tasks (cheetah-run, walker-run,
           hopper-hop, humanoid-walk) at $64{\times}64$ pixels. Training/test: 9M/1M transitions.
    \item  \emph{Distracting Meta-World} (DMW): augments Meta-World~\citep{yu_meta-world_2020} with dynamic natural
           video backgrounds (full DAVIS train/val split, 90 videos) at the \emph{easy} dynamic setting. We evaluate
           the MT10 subset at $128{\times}128$ pixels. DMW introduces richer agent geometry (arm, gripper,
           manipulated objects), testing mask quality under complex interactions. Training/test: 1M/0.1M transitions.
\end{itemize}

\begin{wrapfigure}[22]{r}{0.39\textwidth}
    \centering
    \setlength{\tabcolsep}{2pt}
    \begin{tabular}{@{}cc@{}}
        \includegraphics[width=0.48\linewidth]{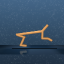} &
        \includegraphics[width=0.48\linewidth]{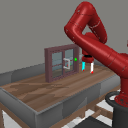} \\
        \includegraphics[width=0.48\linewidth]{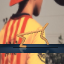} &
        \includegraphics[width=0.48\linewidth]{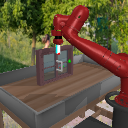} \\
        \shortstack{a) cheetah-run} & \shortstack{b) window-open}
    \end{tabular}
    \caption{\textbf{Environments (top: vanilla, bottom: distractor).} DCS (a) adds dynamic background videos, agent color randomization, and camera shake; DMW (b) adds dynamic background video distractors only.}
    \label{fig:environment}
\end{wrapfigure}

For all environments we provide both ground-truth masks (from simulation) and SAM~2.1
masks~\citep{ravi_sam_2024}; all datasets will be released upon acceptance.

\textbf{Baselines.} \emph{LAPO}~\citep{schmidt_learning_2023} trains the IDM-FDM pair without auxiliary losses; following
LAOM~\citep{nikulin_latent_2025} we remove the VQ-VAE for better performance on continuous control. \emph{LAOM-Labels} adds an auxiliary
linear probe loss decoding ground-truth actions during Stage~1 training, establishing an upper bound for what
privileged label access can achieve; we use $128\text{k}$ ground-truth labels unless stated otherwise. \emph{LAOF}~\citep{bu_laof_2026} adds a masked optical flow auxiliary loss, the
closest prior method that also leverages segmentation masks, but on flow rather than reconstruction.
\emph{LAPO-Slots}~\citep{klepach_object-centric_2026} decomposes observations into object slots and masks the
\emph{input} rather than the loss. We report MaskLAM in two configurations, \textbf{MaskLAM@GT} with simulator
ground-truth masks and \textbf{MaskLAM@SAM} with SAM~2.1 masks~\citep{ravi_sam_2024}; unsuffixed \textbf{MaskLAM}
refers to both jointly. All baselines use their
published hyperparameters, adapted only for convergence on our datasets, with dimension $d_z=128$ for the latent action space unless stated otherwise; full configurations in \Cref{sec:appendix_implementation_details,sec:appendix_hyperparameters}.

\textbf{Metrics.} Three metrics capture different aspects of latent action quality, all normalized
against the expert policy that collected the dataset (\Cref{sec:appendix_expert_results}) and computed over 3 random
seeds as mean $\pm$ standard deviation; see \Cref{sec:appendix_evaluation_details} for the full protocol.
\begin{itemize}
    \item  \emph{Normalized linear action probe MSE} (NMSE): a linear regressor on frozen latent actions predicts
           ground-truth actions, with no gradient flowing back~\citep{alain_understanding_2018,zhang_what_2025},
           normalized by the expert policy's action variance.
    \item  \emph{Normalized success rate} (NSR): mean success fraction over 100 episodes, normalized by the expert collector's rate; reported
           on DMW following~\citep{yu_meta-world_2020}.
    \item  \emph{Normalized return} (NR): mean return over 100 episodes, normalized so the expert equals $100\%$;
           reported on DCS following~\citep{tunyasuvunakool_dm_control_2020}.
\end{itemize}

% Results and Discussion
\section{Results and Discussion}
\label{sec:results}

\begin{figure}[H]
    \centering
    \setlength{\tabcolsep}{1pt}
    \begin{tabular*}{\linewidth}{@{\extracolsep{\fill}}cccccc@{}}
        \includegraphics[width=0.18\linewidth]{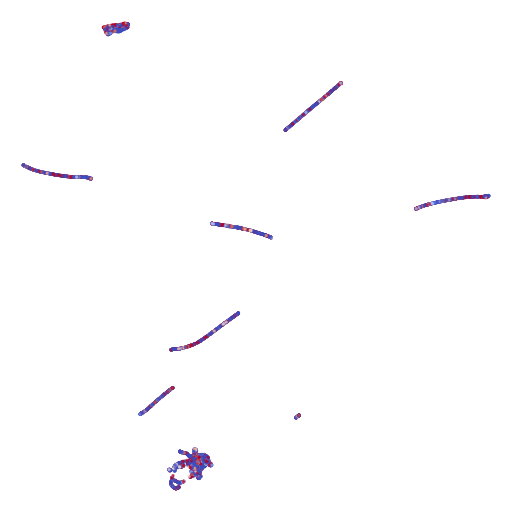} &
        \includegraphics[width=0.18\linewidth]{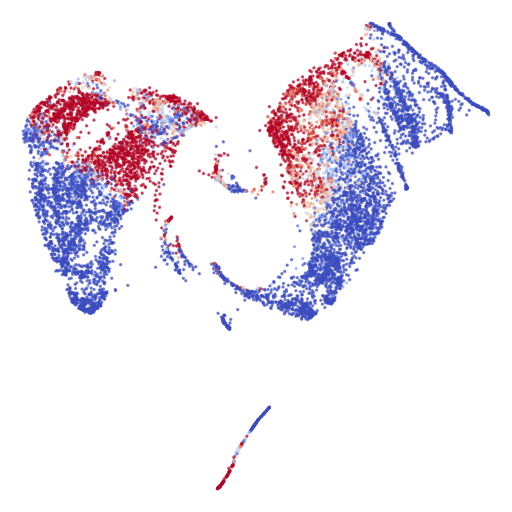} &
        \includegraphics[width=0.18\linewidth]{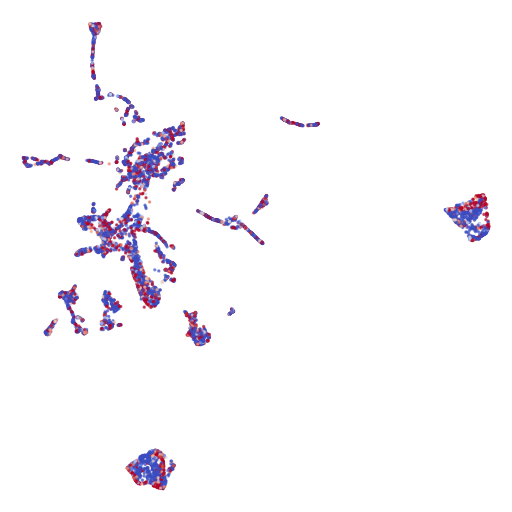} &
        \includegraphics[width=0.18\linewidth]{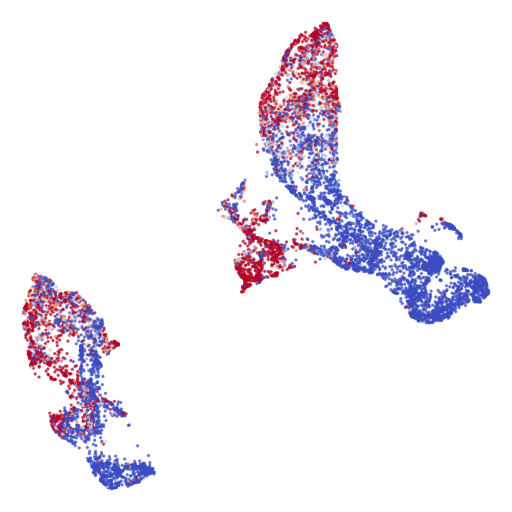} &
        \includegraphics[width=0.18\linewidth]{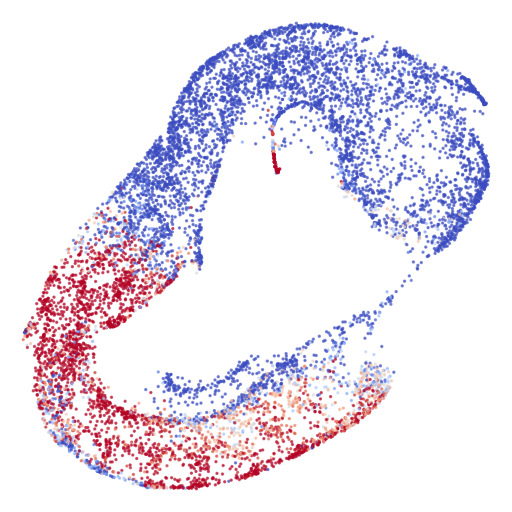} &
        \begin{tikzpicture}
            \shade[bottom color=blue!70,top color=red!70,middle color=white] (0,0) rectangle (0.018\linewidth,0.18\linewidth);
            \node[font=\scriptsize] at (0.009\linewidth,0.02\linewidth) {-1};
            \node[font=\scriptsize] at (0.009\linewidth,0.09\linewidth) {0};
            \node[font=\scriptsize] at (0.009\linewidth,0.16\linewidth) {1};
        \end{tikzpicture} \\
        \makebox[0.18\linewidth][c]{\scriptsize LAPO} &
        \makebox[0.18\linewidth][c]{\scriptsize LAOM-Labels} &
        \makebox[0.18\linewidth][c]{\scriptsize LAOF} &
        \makebox[0.18\linewidth][c]{\scriptsize LAPO-Slots} &
        \makebox[0.18\linewidth][c]{\scriptsize MaskLAM} &
        \makebox[0.018\linewidth][c]{}
    \end{tabular*}
    \caption{\textbf{UMAP projection of the latent action space} on cheetah-run under the distracting setting of the DCS,
    colored by the 4th ground-truth action dimension. A well-aligned latent space exhibits a smooth color gradient
    along the projection. MaskLAM produces a clean, monotonically structured manifold, while LAPO and LAOF collapse
    into entangled clusters with no clear correspondence to the ground-truth action. Per-dimension projections for
    all joints are provided in \Cref{sec:appendix_umap}.}
    \label{fig:umap_action_dim4_methods}
\end{figure}

We frame our analysis through the lens of \citet{zhang_what_2025}, who show that linear latent action models trained by
next-frame reconstruction behave like PCA on the change between consecutive frames: the
latent encodes whatever varies most across the data, regardless of source. In distractor-free benchmarks the
dominant variance is action-driven, and LAPO~\citep{schmidt_learning_2023} is able to recover the true actions with almost no additional cost.
Once distractors enter the frame, the same loss steers the latent toward exogenous variance instead. MaskLAM
intervenes at exactly this point: by zeroing the FDM gradient on distractor pixels, it leaves only action-driven
variance in the supervisory target. The rest of this section traces the consequences: loss masking restores
alignment and downstream performance under distractors, admits smaller latents and label budgets, and remains
effective even when the supervisory mask is degraded by occlusion or geometric error.

\textbf{MaskLAM recovers action-aligned latents under distractors.}
Across both vanilla and distractor settings, MaskLAM achieves the lowest probe error
among label-free methods (LAPO, LAPO-Slots, LAOF), and the gap widens when distractors
are added (\Cref{tab:q1_combined_aggregate}). MaskLAM@GT and MaskLAM@SAM are within noise of
each other, indicating that off-the-shelf segmentation models are sufficient for
MaskLAM to be effective. The aligned latent actions are visible in
\Cref{fig:umap_action_dim4_methods}: MaskLAM produces a smooth monotonic gradient over
the ground-truth action dimension, while LAPO and LAOF collapse into entangled clusters.

\textbf{Better alignment translates to better policies.}
\Cref{tab:q1_combined_aggregate} reports the corresponding downstream return. MaskLAM@GT matches or exceeds every
label-free baseline in both settings, with the largest gap under distractors. MaskLAM@GT also improves over LAPO in the
distractor-free setting; we hypothesize that even in clean scenes the unmasked FDM loss diverts a portion of the
latent's capacity toward reconstructing background structure rather than action-driven change, an effect we make
explicit in the latent-dimension sweep of \crefpanel{fig:q3_q5_aggregate_normalized}{a}.

\begin{table}[t]
    \centering
    \renewcommand{\arraystretch}{1.15}
\setlength{\tabcolsep}{4pt}
\resizebox{\textwidth}{!}{%
\begin{tabular}{lcccccccc}
 & \multicolumn{4}{c}{\textbf{Distracting Control Suite}} & \multicolumn{4}{c}{\textbf{Distracting Meta-World}} \\
\cmidrule(lr){2-5}\cmidrule(lr){6-9}
 & \multicolumn{2}{c}{\textbf{Mean NMSE~$\downarrow$}} & \multicolumn{2}{c}{\textbf{Mean NR~$\uparrow$}} & \multicolumn{2}{c}{\textbf{Mean NMSE~$\downarrow$}} & \multicolumn{2}{c}{\textbf{Mean NSR~$\uparrow$}} \\
\cmidrule(lr){2-3}\cmidrule(lr){4-5}\cmidrule(lr){6-7}\cmidrule(lr){8-9}
\textbf{Method} & \textbf{Vanilla} & \textbf{Distractor} & \textbf{Vanilla} & \textbf{Distractor} & \textbf{Vanilla} & \textbf{Distractor} & \textbf{Vanilla} & \textbf{Distractor} \\
\midrule
LAPO & $0.2971_{\pm 0.1169}$ & $0.5114_{\pm 0.0336}$ & $0.2364_{\pm 0.0277}$ & $0.0759_{\pm 0.0092}$ & $\mathbf{0.0742_{\pm 0.0036}}$ & $0.3093_{\pm 0.0094}$ & $\mathbf{0.8510_{\pm 0.0257}}$ & $0.6362_{\pm 0.0147}$ \\
LAOM-Labels & $0.2106_{\pm 0.0060}$ & $0.2160_{\pm 0.0034}$ & $\mathbf{0.6083_{\pm 0.0165}}$ & $\mathbf{0.5663_{\pm 0.0186}}$ & $0.1137_{\pm 0.0099}$ & $0.1421_{\pm 0.0077}$ & $0.7270_{\pm 0.0278}$ & $0.7810_{\pm 0.0328}$ \\
LAOF & $0.6390_{\pm 0.0451}$ & $0.5056_{\pm 0.0194}$ & $0.0721_{\pm 0.0169}$ & $0.0939_{\pm 0.0195}$ & $0.1849_{\pm 0.0370}$ & $0.4529_{\pm 0.0569}$ & $0.7160_{\pm 0.1175}$ & $0.5768_{\pm 0.0791}$ \\
LAPO-Slots & $0.3516_{\pm 0.0122}$ & $0.6570_{\pm 0.0293}$ & $0.1264_{\pm 0.0205}$ & $0.0454_{\pm 0.0078}$ & $0.1090_{\pm 0.0060}$ & $0.2450_{\pm 0.0481}$ & $0.8326_{\pm 0.0323}$ & $0.8274_{\pm 0.0229}$ \\
\midrule
\textbf{MaskLAM@GT} & $\mathbf{0.2049_{\pm 0.0063}}$ & $\mathbf{0.2061_{\pm 0.0093}}$ & $0.5126_{\pm 0.0286}$ & $0.3776_{\pm 0.0298}$ & $0.0815_{\pm 0.0041}$ & $\mathbf{0.0880_{\pm 0.0036}}$ & $0.8276_{\pm 0.0364}$ & $\mathbf{0.8432_{\pm 0.0211}}$ \\
\textbf{MaskLAM@SAM} & $0.2129_{\pm 0.0096}$ & $0.2326_{\pm 0.0372}$ & $0.4934_{\pm 0.0359}$ & $0.3581_{\pm 0.0404}$ & $0.0779_{\pm 0.0030}$ & $0.0903_{\pm 0.0049}$ & $0.8282_{\pm 0.0215}$ & $0.8318_{\pm 0.0314}$ \\
\end{tabular}%
}

    \caption{\textbf{Latent action quality (NMSE) and downstream behavior cloning performance (NR/NSR)} aggregated over environment suites in the vanilla and distractor settings. MaskLAM@GT and MaskLAM@SAM recover latent actions that align more tightly with ground-truth controls than most baselines, leading to improved downstream performance. This gap increases under distractors.}
    \label{tab:q1_combined_aggregate}
\end{table}

\begin{wrapfigure}[14]{R}{0.42\linewidth}
    \centering
    \vspace*{-\baselineskip}
    \setlength{\tabcolsep}{2pt}
    \begin{tabular*}{\linewidth}{@{\extracolsep{\fill}}cc@{}}
        \includegraphics[width=0.48\linewidth]{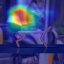} &
        \includegraphics[width=0.48\linewidth]{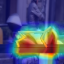} \\
        \scriptsize (a) LAPO & \scriptsize (b) MaskLAM
    \end{tabular*}
    \caption{\textbf{Eigen-CAM saliency on cheetah-run DCS (distractor).} Warmer colors indicate higher influence on the predicted latent action.}
    \label{fig:eigencam_cheetah_frame_0}
\end{wrapfigure}

\textbf{MaskLAM focuses on the agent, not the distractor.}
To verify that the alignment improvement is driven by the intended mechanism rather than by incidental architectural
changes, we visualize the IDM's saliency via Eigen-CAM~\citep{muhammad_eigen-cam_2020} (\Cref{fig:eigencam_cheetah_frame_0}).
LAPO's saliency falls on the distractor background, while MaskLAM's covers the cheetah body and extends to its
contact surface with the ground. Part of the saliency falls outside $M_{t+1}$, indicating that the IDM learns from
data, beyond what the mask directly supervises, to attend to the body and its contact with the ground. Full time
series in \Cref{sec:appendix_eigencam}.

\textbf{MaskLAM enables compact latent action spaces.}
\crefpanel{fig:q3_q5_aggregate_normalized}{a} sweeps the latent action dimension with and without the masked loss. With masking,
a 64-dim latent already matches the alignment of a 256-dim latent without it; the gap is largest in the
low-dimensional regime, where encoding extra exogenous content would push out true action information.
\citet{nikulin_latent_2025} report the same trade-off in LAOM, and \citet{ye_latent_2024} observe it in LAPA. By removing the gradient pressure that ties the latent size to the prediction target, MaskLAM lets the model use a small latent without sacrificing alignment.

\begin{figure}[h]
    \centering
    \includegraphics[width=\textwidth]{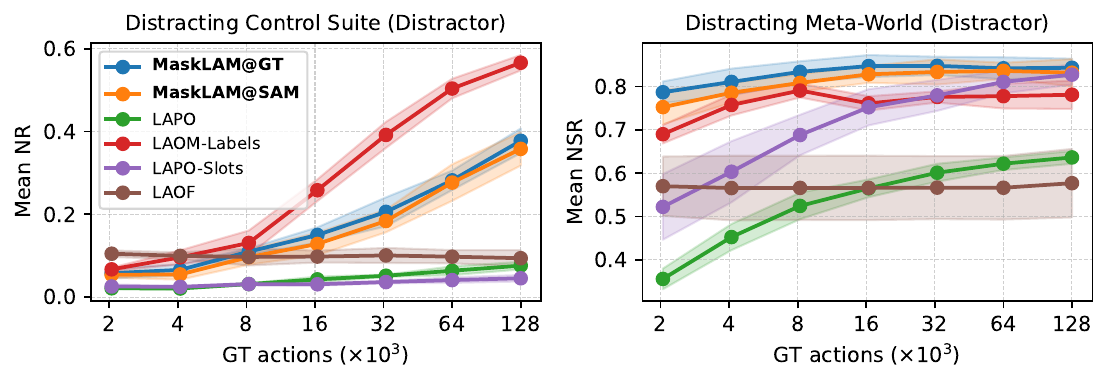}
    \vspace*{-\baselineskip}
    \caption{\textbf{Sample efficiency as a function of the number of ground-truth action labels (log scale) available for action decoder fine-tuning.} MaskLAM@GT peaks with as few as $2\text{k}$ labels on DMW (distractor), while LAPO and LAPO-Slots require nearly two orders of magnitude more labels to close the gap, and on DCS (distractor) they fail to recover useful policies at any label budget.}
    \label{fig:q2_aggregate_suites_normalized}
\end{figure}

\textbf{Cleaner latents allow smaller label budgets.}
\Cref{fig:q2_aggregate_suites_normalized} reports downstream return as a function of the number of action labels available for Stage~3 decoder fine-tuning. On DMW (distractor), MaskLAM reaches near peak performance with as few as $2\text{k}$ labels, while LAPO and LAPO-Slots require nearly two orders of magnitude more labels to close the gap. On DCS (distractor) the asymmetry sharpens: LAPO and LAPO-Slots fail to recover a useful policy at any label budget we tested. MaskLAM is strong under limited labels but LAOM-Labels overtakes it at larger label budgets. Since MaskLAM actually has lower probe NMSE (\Cref{tab:q1_combined_aggregate}), the gap is not a decoding limit; we hypothesize that direct action supervision yields a latent that holds up better under rollout-induced distribution shift, visible on the longer rollouts on DCS.

\textbf{MaskLAM degrades gracefully under agent occlusion.}
Real videos rarely show the full agent at all times. We simulate this on DCS (distractor) by masking out a growing fraction of
the agent at training time and measuring the resulting probe error
(\crefpanel{fig:q3_q5_aggregate_normalized}{b}). With the masked loss, probe error stays nearly flat up to $75\%$ occlusion;
without it, the same level of occlusion more than doubles it. Occlusion erodes the target's action-to-noise ratio
under the unmasked loss, whereas MaskLAM holds it steady by keeping distractor variance out of the target. Further details in \Cref{sec:appendix_q5}

\begin{figure}[b]
    \centering
    \includegraphics[width=\textwidth]{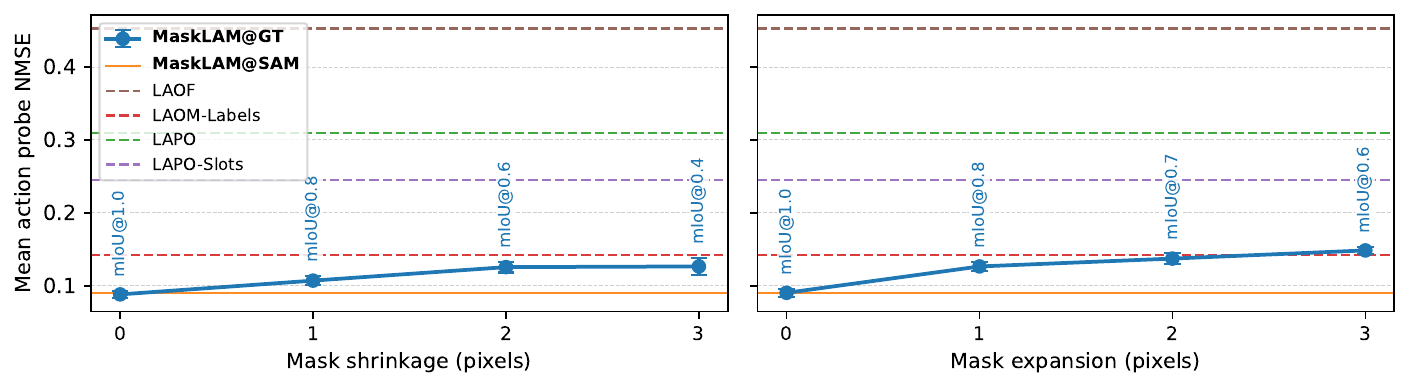}
    \vspace*{-\baselineskip}
    \caption{\textbf{Robustness to imperfect masks on DMW (distractor).} MaskLAM@GT is trained with ground-truth masks perturbed by morphological shrinkage (left) or expansion (right) of $0$--$3$ pixels, which drives the mask mIoU down to $0.4$--$0.6$. Baselines are shown as dashed horizontal references. MaskLAM@GT remains below LAPO, LAPO-Slots, and LAOF across the entire perturbation range, indicating that the method \textbf{requires no} perfect segmentation to recover well-aligned latent actions.}
    \label{fig:q6_aggregate_normalized}
\end{figure}

\textbf{MaskLAM is robust to imperfect segmentation masks.}
In practice the supervisory mask comes from an off-the-shelf segmenter, whose boundaries are never pixel-perfect. We
perturb ground-truth masks by morphological shrinkage or expansion of $0$ to $3$ pixels on DWM (distractor), dropping mIoU as low as
$0.4$, and re-train MaskLAM@GT (\Cref{fig:q6_aggregate_normalized}). Probe error stays below LAPO, LAPO-Slots, and
LAOF across the entire perturbation range. Together with the within-noise agreement between MaskLAM@GT and
MaskLAM@SAM throughout this section, the takeaway is practical: pixel-perfect segmentation is not required. See mask visualization in \Cref{fig:q6_dmw_morphology_examples} for further details.

\begin{figure}[t]
    \centering
    \setlength{\tabcolsep}{2pt}
    \begin{tabular*}{\linewidth}{@{\extracolsep{\fill}}cc@{}}
        \includegraphics[width=0.48\linewidth]{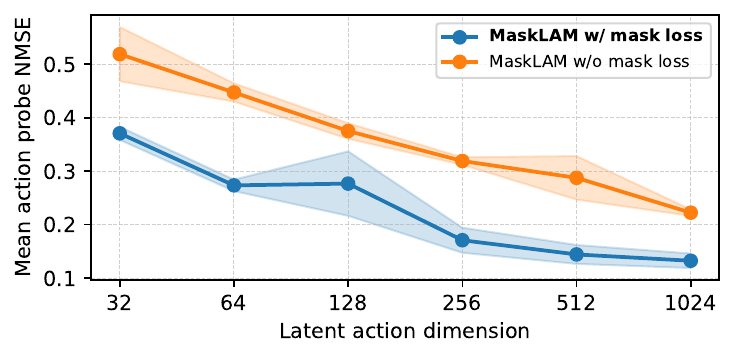} &
        \includegraphics[width=0.48\linewidth]{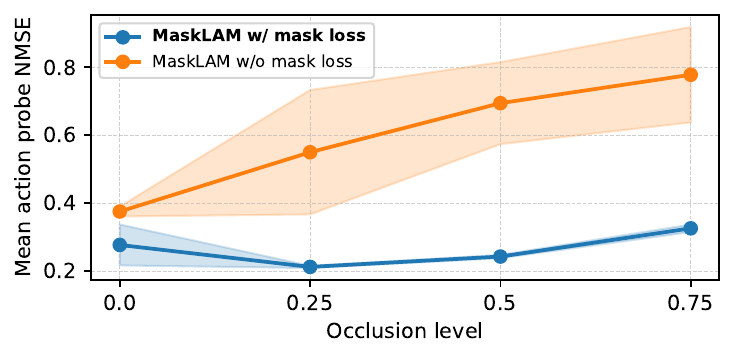} \\
        (a) Latent action dimension & (b) Agent occlusion level
    \end{tabular*}
    \caption{\textbf{Action probe NMSE on DCS (distractor) with and without the mask loss.}
    (a)~Latent action dimension (log scale): masking lets a 64-dim latent match a 256-dim unmasked one, with the
    largest gap in the low-dimensional regime.
    (b)~Agent occlusion level: error stays nearly flat up to $75\%$ occlusion with masking, but more than doubles
    without it.}
    \label{fig:q3_q5_aggregate_normalized}
\end{figure}

\subsection{Ablation Study}
\label{sec:ablation}

\begin{wrapfigure}[21]{R}{0.45\linewidth}
    \centering
    \vspace*{-\baselineskip}
    \includegraphics[width=\linewidth,trim=0 0 0 0.2cm, clip]{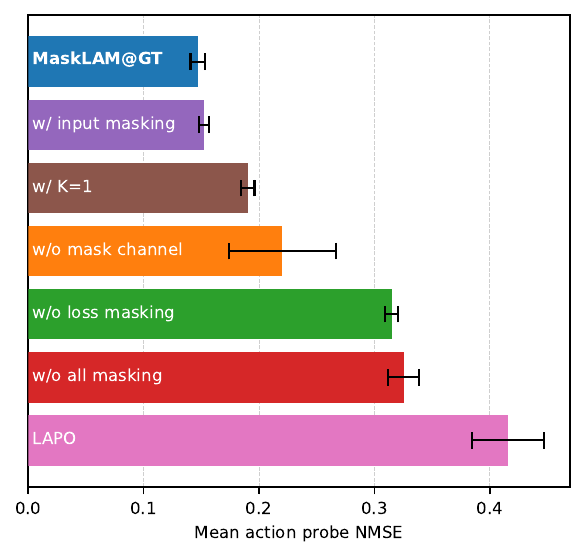}
    \vspace*{-\baselineskip}
    \caption{\textbf{Ablation of MaskLAM components aggregated on DCS and DMW (distractor).}
    Loss masking dominates; the mask channel helps only alongside it; multi-step IDM adds a smaller independent gain.}
    \label{fig:ablation_aggregate_dcs_dmw_combined_normalized}
\end{wrapfigure}

To isolate the contribution of each component, we ablate the masked loss (\textit{w/o loss masking}), the additional mask input
channel (\textit{w/o mask channel}), both together (\textit{w/o all masking}), and the multi-step IDM
(\textit{w/ $k{=}1$}). We further compare against a variant that zeroes distractor pixels in the input rather than in
the loss (\textit{w/ input masking}). Removing every component reduces MaskLAM to LAPO.
\Cref{fig:ablation_aggregate_dcs_dmw_combined_normalized} reports the resulting action probe NMSE aggregated across
DCS and DMW (distractor).

\textbf{Loss masking is the principal contributor.} Removing the masked loss alone more than doubles the alignment
error, while turning every component off recovers LAPO. The four perturbations roughly compose back to the LAPO
baseline, so the gain over LAPO is decomposable. Loss masking accounts for the largest individual share, consistent
with the analysis of \citet{zhang_what_2025}: it is the only component that removes exogenous variance from the
supervisory target itself.

\textbf{The mask channel contributes only in combination with the masked loss.} Dropping the mask input channel hurts
performance when the masked loss is present, but leaves performance largely unchanged when it is not. The channel
acts as a multiplier on loss masking rather than as an independent fix.

\textbf{Multi-step IDM is a smaller, partially independent gain.} Setting $k{=}1$ degrades performance both in
isolation and on top of removing all masking. The effect persists with masking turned on, so multi-step IDM is
orthogonal to our masking intervention.

\textbf{Input masking matches loss masking quantitatively, not qualitatively.} On DCS and DMW (distractor), replacing
the masked loss with input masking is statistically indistinguishable from MaskLAM@GT. The mechanism is not:
Eigen-CAM (\Cref{fig:eigencam_cheetah_frame_0}) shows the IDM's saliency extends to ground-contact pixels outside
$M_{t+1}$, which loss masking keeps visible but input masking would zero at the input. Parity on these benchmarks
reflects that the agent's motion is essentially self-driven, so the contact surface adds little predictive signal
beyond the agent pixels. In real video the agent's motion typically also depends on external factors, e.g., a rider
on an electric scooter or a commuter on an escalator. There, the interaction surface carries action-relevant signal
that input masking discards.

% Limitations
\section{Limitations}
\label{sec:limitations}

MaskLAM improves robustness in visually complex settings but has a few practical limitations. Firstly, MaskLAM depends critically on the upstream segmenter: segmentation errors directly corrupt the learning signal for $z_t$. SAM 2.1 performs well zero-shot but has known failure modes such as prompt ambiguity, out-of-distribution embodiments, and identity drift. In practice the method tolerates substantial mask degradation (\Cref{fig:q6_aggregate_normalized}), so pixel-perfect segmentation is not required; this is especially true for real-world targets, which lie inside SAM's training distribution unlike simulator environments such as MuJoCo~\citep{todorov2012mujoco}. Secondly, this method inherits limitations of pixel-level reconstruction, including sensitivity to high-frequency noise and increased computational cost and influence of confounding factors like textures on the agent's body. Finally, our evaluation relies on synthetic distractor benchmarks. While they may not capture the full complexity of real-world visual noise, they are widely used and provide controlled, reproducible settings that enable precise analysis of robustness to specific distractor types.

\section{Conclusion}
\label{sec:conclusion}

In this work, we revisited the problem of learning latent actions from action-free video in the presence of action-correlated distractors. We showed that LAPO's failure in this regime is driven by its prediction target  
  rather than its architecture: as soon as exogenous variance enters the next-frame reconstruction loss, the latent action absorbs it. We proposed MaskLAM, a lightweight modification of LAPO that restricts the forward-dynamics
  loss to agent pixels using zero-shot masks from an off-the-shelf segmentation model, requiring no auxiliary losses, no architectural changes, and no action labels during pre-training. We found that this single           
  intervention closes the gap to LAOM-Labels, which depends on privileged action supervision. Compared to label-free baselines, MaskLAM additionally yields more compact latent action spaces, lower action-label budgets for 
  downstream decoding, and graceful degradation under agent occlusion and imperfect masks. Our findings suggest that the supervision needed to disentangle agent from distractor dynamics can be re-allocated from action     
  labels, which are scarce, to segmentation masks, which are cheap and already available zero-shot, making latent action pre-training viable on noisy, distractor-rich real-world video.

\begin{ack}
We would like to thank Tim Joseph, Philipp Stegmaier, and Karam Daaboul for many fruitful discussions and detailed feedback on this work. We gratefully acknowledge the computing time provided on the high-performance computer HoreKa by the National High-Performance Computing Center at KIT (NHR@KIT), which is jointly supported by the Federal Ministry of Education and Research and the Ministry of Science, Research and the Arts of Baden-W\"urttemberg as part of the National High-Performance Computing (NHR) joint funding program (\url{https://www.nhr-verein.de/en/our-partners}); HoreKa is partly funded by the German Research Foundation (DFG). This work is further supported by the Helmholtz Association Initiative and Networking Fund on the HAICORE@KIT partition. We declare no competing interests.
\end{ack}

{\small
\bibliographystyle{plainnat}
\bibliography{references}

@article{argall_survey_2009,
  author = {Argall, Brenna D. and Chernova, Sonia and Veloso, Manuela and Browning, Brett},
  title = {A survey of robot learning from demonstration},
  journal = {Robotics and Autonomous Systems},
  volume = {57},
  number = {5},
  pages = {469--483},
  year = {2009},
  doi = {10.1016/j.robot.2008.10.024},
  url = {https://www.sciencedirect.com/science/article/pii/S0921889008001772},
  issn = {0921-8890},
}

@inproceedings{wang_ad3_2024,
  author = {Wang, Yucen and Wan, Shenghua and Gan, Le and Feng, Shuai and Zhan, De-Chuan},
  title = {{AD}3: Implicit Action is the Key for World Models to Distinguish the Diverse Visual Distractors},
  booktitle = {Proceedings of the 41st International Conference on Machine Learning},
  pages = {51546--51568},
  publisher = {{PMLR}},
  year = {2024},
  url = {https://proceedings.mlr.press/v235/wang24bq.html},
  issn = {2640-3498},
}

@inproceedings{ye_become_2022,
  author = {Ye, Weirui and Zhang, Yunsheng and Abbeel, Pieter and Gao, Yang},
  title = {Become a Proficient Player with Limited Data through Watching Pure Videos},
  booktitle = {International Conference on Learning Representations},
  year = {2023},
  url = {https://openreview.net/forum?id=Sy-o2N0hF4f},
}

@article{levine_end--end_2016,
  author = {Levine, Sergey and Finn, Chelsea and Darrell, Trevor and Abbeel, Pieter},
  title = {End-to-End Training of Deep Visuomotor Policies},
  journal = {Journal of Machine Learning Research},
  volume = {17},
  number = {39},
  pages = {1--40},
  year = {2016},
  url = {http://jmlr.org/papers/v17/15-522.html},
  issn = {1533-7928},
}

@inproceedings{cui_play_2022,
  author = {Cui, Zichen Jeff and Wang, Yibin and Shafiullah, Nur Muhammad Mahi and Pinto, Lerrel},
  title = {From Play to Policy: Conditional Behavior Generation from Uncurated Robot Data},
  booktitle = {International Conference on Learning Representations},
  year = {2023},
  url = {https://openreview.net/forum?id=c7rM7F7jQjN},
}

@inproceedings{edwards_imitating_2019,
  author = {Edwards, Ashley and Sahni, Himanshu and Schroecker, Yannick and Isbell, Charles},
  title = {Imitating Latent Policies from Observation},
  booktitle = {Proceedings of the 36th International Conference on Machine Learning},
  pages = {1755--1763},
  publisher = {{PMLR}},
  year = {2019},
  url = {https://proceedings.mlr.press/v97/edwards19a.html},
  issn = {2640-3498},
}

@inproceedings{nikulin_latent_2025,
  author = {Nikulin, Alexander and Zisman, Ilya and Tarasov, Denis and Lyubaykin, Nikita and Polubarov, Andrei and Kiselev, Igor and Kurenkov, Vladislav},
  title = {Latent Action Learning Requires Supervision in the Presence of Distractors},
  booktitle = {Proceedings of the 42nd International Conference on Machine Learning},
  pages = {46427--46447},
  publisher = {{PMLR}},
  year = {2025},
  url = {https://proceedings.mlr.press/v267/nikulin25a.html},
  issn = {2640-3498},
}

@inproceedings{ye_latent_2024,
  author = {Ye, Seonghyeon and Jang, Joel and Jeon, Byeongguk and Joo, Se June and Yang, Jianwei and Peng, Baolin and Mandlekar, Ajay and Tan, Reuben and Chao, Yu-Wei and Lin, Bill Yuchen and Liden, Lars and Lee, Kimin and Gao, Jianfeng and Zettlemoyer, Luke and Fox, Dieter and Seo, Minjoon},
  title = {Latent Action Pretraining from Videos},
  booktitle = {International Conference on Learning Representations},
  year = {2025},
  url = {https://proceedings.iclr.cc/paper_files/paper/2025/hash/45d74e190008c7bff2845ffc8e3facd3-Abstract-Conference.html},
}

@inproceedings{schmidt_learning_2023,
  author = {Schmidt, Dominik and Jiang, Minqi},
  title = {Learning to Act without Actions},
  booktitle = {International Conference on Learning Representations},
  year = {2024},
  url = {https://openreview.net/forum?id=rvUq3cxpDF},
}

@inproceedings{efroni_provably_2021,
  author = {Efroni, Yonathan and Misra, Dipendra and Krishnamurthy, Akshay and Agarwal, Alekh and Langford, John},
  title = {Provably Filtering Exogenous Distractors using Multistep Inverse Dynamics},
  booktitle = {International Conference on Learning Representations},
  year = {2022},
  url = {https://openreview.net/forum?id=RQLLzMCefQu},
}

@inproceedings{shah_pitfalls_2020,
  author = {Shah, Harshay and Tamuly, Kaustav and Raghunathan, Aditi and Jain, Prateek and Netrapalli, Praneeth},
  title = {The Pitfalls of Simplicity Bias in Neural Networks},
  booktitle = {Advances in Neural Information Processing Systems},
  volume = {33},
  pages = {9573--9585},
  publisher = {Curran Associates, Inc.},
  year = {2020},
  url = {https://proceedings.neurips.cc/paper/2020/hash/6cfe0e6127fa25df2a0ef2ae1067d915-Abstract.html},
}

@inproceedings{hu_video_2025,
  author = {Hu, Yucheng and Guo, Yanjiang and Wang, Pengchao and Chen, Xiaoyu and Wang, Yen-Jen and Zhang, Jianke and Sreenath, Koushil and Lu, Chaochao and Chen, Jianyu},
  title = {Video Prediction Policy: A Generalist Robot Policy with Predictive Visual Representations},
  booktitle = {Proceedings of the 42nd International Conference on Machine Learning},
  pages = {24328--24346},
  publisher = {{PMLR}},
  year = {2025},
  url = {https://proceedings.mlr.press/v267/hu25g.html},
  issn = {2640-3498},
}

@misc{zhang_what_2025,
  author = {Zhang, Chuheng and Pearce, Tim and Zhang, Pushi and Wang, Kaixin and Chen, Xiaoyu and Shen, Wei and Zhao, Li and Bian, Jiang},
  title = {What Do Latent Action Models Actually Learn?},
  publisher = {{arXiv}},
  year = {2025},
  doi = {10.48550/arXiv.2506.15691},
  url = {http://arxiv.org/abs/2506.15691},
  eprint = {2506.15691},
  archivePrefix = {arXiv},
  primaryClass = {cs.AI},
  note = {arXiv preprint arXiv:2506.15691},
}

@article{klepach_object-centric_2026,
  author = {Klepach, Albina and Nikulin, Alexander and Zisman, Ilya and Tarasov, Denis and Derevyagin, Alexander and Polubarov, Andrei and Lyubaykin, Nikita and Kiselev, Igor and Kurenkov, Vladislav},
  title = {Object-Centric Latent Action Learning},
  journal = {Proceedings of the {AAAI} Conference on Artificial Intelligence},
  volume = {40},
  number = {27},
  pages = {22626--22634},
  year = {2026},
  doi = {10.1609/aaai.v40i27.39423},
  url = {https://ojs.aaai.org/index.php/AAAI/article/view/39423},
  issn = {2374-3468, 2159-5399},
}

@misc{bu_laof_2026,
  author = {Bu, Xizhou and Lyu, Jiexi and Sun, Fulei and Yang, Ruichen and Ma, Zhiqiang and Li, Wei},
  title = {{LAOF}: Robust Latent Action Learning with Optical Flow Constraints},
  publisher = {{arXiv}},
  year = {2025},
  doi = {10.48550/arXiv.2511.16407},
  url = {http://arxiv.org/abs/2511.16407},
  eprint = {2511.16407},
  archivePrefix = {arXiv},
  primaryClass = {cs.RO},
  note = {arXiv preprint arXiv:2511.16407},
}

@article{lamb_guaranteed_2022,
  author = {Lamb, Alex and Islam, Riashat and Efroni, Yonathan and Didolkar, Aniket Rajiv and Misra, Dipendra and Foster, Dylan J. and Molu, Lekan P. and Chari, Rajan and Krishnamurthy, Akshay and Langford, John},
  title = {Guaranteed Discovery of Control-Endogenous Latent States with Multi-Step Inverse Models},
  journal = {Transactions on Machine Learning Research},
  year = {2023},
  url = {https://openreview.net/forum?id=TNocbXm5MZ},
  issn = {2835-8856},
}

@inproceedings{ravi_sam_2024,
  author = {Ravi, Nikhila and Gabeur, Valentin and Hu, Yuan-Ting and Hu, Ronghang and Ryali, Chaitanya and Ma, Tengyu and Khedr, Haitham and Rädle, Roman and Rolland, Chloe and Gustafson, Laura and Mintun, Eric and Pan, Junting and Alwala, Kalyan Vasudev and Carion, Nicolas and Wu, Chao-Yuan and Girshick, Ross and Dollar, Piotr and Feichtenhofer, Christoph},
  title = {{SAM} 2: Segment Anything in Images and Videos},
  booktitle = {International Conference on Learning Representations},
  year = {2025},
  url = {https://proceedings.iclr.cc/paper_files/paper/2025/hash/45c1f6a8cbf2da59ebf2c802b4f742cd-Abstract-Conference.html},
}

@inproceedings{espeholt_impala_2018,
  author = {Espeholt, Lasse and Soyer, Hubert and Munos, Remi and Simonyan, Karen and Mnih, Vlad and Ward, Tom and Doron, Yotam and Firoiu, Vlad and Harley, Tim and Dunning, Iain and Legg, Shane and Kavukcuoglu, Koray},
  title = {{IMPALA}: Scalable Distributed Deep-{RL} with Importance Weighted Actor-Learner Architectures},
  booktitle = {Proceedings of the 35th International Conference on Machine Learning},
  pages = {1407--1416},
  publisher = {{PMLR}},
  year = {2018},
  url = {https://proceedings.mlr.press/v80/espeholt18a.html},
  issn = {2640-3498},
}

@inproceedings{
levine_multistep_2024,
title={Multistep Inverse Is Not All You Need},
author={Alexander Levine and Peter Stone and Amy Zhang},
booktitle={Reinforcement Learning Conference},
year={2024},
url={https://openreview.net/forum?id=xyrgG4rsqY}
}

@inproceedings{he_deep_2016,
  author = {He, Kaiming and Zhang, Xiangyu and Ren, Shaoqing and Sun, Jian},
  title = {Deep Residual Learning for Image Recognition},
  booktitle = {Proceedings of the {IEEE} Conference on Computer Vision and Pattern Recognition},
  pages = {770--778},
  year = {2016},
  url = {https://openaccess.thecvf.com/content_cvpr_2016/html/He_Deep_Residual_Learning_CVPR_2016_paper.html},
}

@misc{pont-tuset_2017_2018,
  author = {Pont-Tuset, Jordi and Perazzi, Federico and Caelles, Sergi and Arbeláez, Pablo and Sorkine-Hornung, Alex and Gool, Luc Van},
  title = {The 2017 {DAVIS} Challenge on Video Object Segmentation},
  publisher = {{arXiv}},
  year = {2017},
  doi = {10.48550/arXiv.1704.00675},
  url = {http://arxiv.org/abs/1704.00675},
  eprint = {1704.00675},
  archivePrefix = {arXiv},
  primaryClass = {cs.CV},
  note = {arXiv preprint arXiv:1704.00675},
}

@article{tunyasuvunakool_dm_control_2020,
  author = {Tunyasuvunakool, Saran and Muldal, Alistair and Doron, Yotam and Liu, Siqi and Bohez, Steven and Merel, Josh and Erez, Tom and Lillicrap, Timothy and Heess, Nicolas and Tassa, Yuval},
  title = {dm\_control: Software and tasks for continuous control},
  journal = {Software Impacts},
  volume = {6},
  pages = {100022},
  year = {2020},
  doi = {10.1016/j.simpa.2020.100022},
  url = {https://www.sciencedirect.com/science/article/pii/S2665963820300099},
  issn = {2665-9638},
}

@inproceedings{yu_meta-world_2020,
  author = {Yu, Tianhe and Quillen, Deirdre and He, Zhanpeng and Julian, Ryan and Hausman, Karol and Finn, Chelsea and Levine, Sergey},
  title = {Meta-World: A Benchmark and Evaluation for Multi-Task and Meta Reinforcement Learning},
  booktitle = {Proceedings of the Conference on Robot Learning},
  pages = {1094--1100},
  publisher = {{PMLR}},
  year = {2020},
  url = {https://proceedings.mlr.press/v100/yu20a.html},
  issn = {2640-3498},
}

@misc{stone_distracting_2021,
  author = {Stone, Austin and Ramirez, Oscar and Konolige, Kurt and Jonschkowski, Rico},
  title = {The Distracting Control Suite -- A Challenging Benchmark for Reinforcement Learning from Pixels},
  year = {2021},
  url = {https://arxiv.org/abs/2101.02722v1},
  eprint = {2101.02722},
  archivePrefix = {arXiv},
  primaryClass = {cs.LG},
  note = {arXiv preprint arXiv:2101.02722},
  titleaddon = {{arXiv}.org},
}

@misc{alain_understanding_2018,
  author = {Alain, Guillaume and Bengio, Yoshua},
  title = {Understanding intermediate layers using linear classifier probes},
  publisher = {{arXiv}},
  year = {2016},
  doi = {10.48550/arXiv.1610.01644},
  url = {http://arxiv.org/abs/1610.01644},
  eprint = {1610.01644},
  archivePrefix = {arXiv},
  primaryClass = {stat},
  note = {arXiv preprint arXiv:1610.01644},
}

@article{locatello2020object,
  author = {Locatello, Francesco and Weissenborn, Dirk and Unterthiner, Thomas and Mahendran, Aravindh and Heigold, Georg and Uszkoreit, Jakob and Dosovitskiy, Alexey and Kipf, Thomas},
  title = {Object-centric learning with slot attention},
  journal = {Advances in neural information processing systems},
  volume = {33},
  pages = {11525--11538},
  year = {2020},
}

@inproceedings{bruce2024genie,
  author = {Bruce, Jake and Dennis, Michael D and Edwards, Ashley and Parker-Holder, Jack and Shi, Yuge and Hughes, Edward and Lai, Matthew and Mavalankar, Aditi and Steigerwald, Richie and Apps, Chris and others},
  title = {Genie: Generative interactive environments},
  booktitle = {Forty-first International Conference on Machine Learning},
  year = {2024},
}

@inproceedings{todorov2012mujoco,
  author = {Todorov, Emanuel and Erez, Tom and Tassa, Yuval},
  title = {Mujoco: A physics engine for model-based control},
  booktitle = {2012 IEEE/RSJ international conference on intelligent robots and systems},
  pages = {5026--5033},
  year = {2012},
  organization = {IEEE},
}

@inproceedings{chen2025moto,
  author = {Chen, Yi and Ge, Yuying and Tang, Weiliang and Li, Yizhuo and Ge, Yixiao and Ding, Mingyu and Shan, Ying and Liu, Xihui},
  title = {Moto: Latent motion token as the bridging language for learning robot manipulation from videos},
  booktitle = {Proceedings of the IEEE/CVF International Conference on Computer Vision},
  pages = {19752--19763},
  year = {2025},
}

@misc{struckmeier2023preventing,
  author = {Oliver Struckmeier and Ville Kyrki},
  title = {Preventing Mode Collapse When Imitating Latent Policies from Observations},
  year = {2023},
  url = {https://openreview.net/forum?id=Mf9fQ0OgMzo},
}

@inproceedings{muhammad_eigen-cam_2020,
  author = {Muhammad, Mohammed Bany and Yeasin, Mohammed},
  title = {Eigen-{CAM}: Class Activation Map using Principal Components},
  booktitle = {2020 International Joint Conference on Neural Networks ({IJCNN})},
  pages = {1--7},
  year = {2020},
  doi = {10.1109/IJCNN48605.2020.9206626},
  url = {https://ieeexplore.ieee.org/abstract/document/9206626},
  issn = {2161-4407},
  note = {{ISSN}: 2161-4407},
}

@article{huang2022cleanrl,
  author = {Shengyi Huang and Rousslan Fernand Julien Dossa and Chang Ye and Jeff Braga and Dipam Chakraborty and Kinal Mehta and João G.M. Araújo},
  title = {CleanRL: High-quality Single-file Implementations of Deep Reinforcement Learning Algorithms},
  journal = {Journal of Machine Learning Research},
  volume = {23},
  number = {274},
  pages = {1--18},
  year = {2022},
  url = {http://jmlr.org/papers/v23/21-1342.html},
}

@article{stable-baselines3,
  author = {Antonin Raffin and Ashley Hill and Adam Gleave and Anssi Kanervisto and Maximilian Ernestus and Noah Dormann},
  title = {Stable-Baselines3: Reliable Reinforcement Learning Implementations},
  journal = {Journal of Machine Learning Research},
  volume = {22},
  number = {268},
  pages = {1-8},
  year = {2021},
  url = {http://jmlr.org/papers/v22/20-1364.html},
}

@misc{schulman_proximal_2017,
  author = {Schulman, John and Wolski, Filip and Dhariwal, Prafulla and Radford, Alec and Klimov, Oleg},
  title = {Proximal Policy Optimization Algorithms},
  publisher = {{arXiv}},
  year = {2017},
  doi = {10.48550/arXiv.1707.06347},
  url = {http://arxiv.org/abs/1707.06347},
  eprint = {1707.06347},
  archivePrefix = {arXiv},
  primaryClass = {cs},
  note = {arXiv preprint arXiv:1707.06347},
}

@inproceedings{haarnoja_soft_2018,
  author = {Haarnoja, Tuomas and Zhou, Aurick and Abbeel, Pieter and Levine, Sergey},
  title = {Soft Actor-Critic: Off-Policy Maximum Entropy Deep Reinforcement Learning with a Stochastic Actor},
  booktitle = {Proceedings of the 35th International Conference on Machine Learning},
  pages = {1861--1870},
  publisher = {{PMLR}},
  year = {2018},
  url = {https://proceedings.mlr.press/v80/haarnoja18b.html},
  issn = {2640-3498},
}
}

%%%%%%%%%%%%%%%%%%%%%%%%%%%%%%%%%%%%%%%%%%%%%%%%%%%%%%%%%%%%

\appendix

\section{Ethical Considerations and Broader Impacts}
\label{sec:appendix_broader_impacts}

MaskLAM lowers the cost of extracting control-relevant signal from action-free video by replacing privileged action supervision with zero-shot segmentation. We summarize the most likely positive and negative consequences of this shift and the segmenter-level failure modes that condition both.

\textbf{Positive impact.} The principal benefit is a reduction in the action-label budget required to pre-train embodied agents on observation-only video. Action labels are scarce and expensive to collect; segmentation masks are cheap and already available zero-shot from foundation models such as SAM~2.1~\citep{ravi_sam_2024}. By re-allocating the supervision burden from action labels to masks, MaskLAM brings large-scale latent action pre-training closer to the data scales that have driven progress in vision and language, with concrete downstream applications in robot learning, assistive robotics, and any embodied domain where teleoperated demonstrations are difficult to obtain. The same shift broadens the pool of researchers who can experiment with latent action models without privileged hardware or large labeled datasets.

\textbf{Negative impact.} The same property is generic to any video source: a method that recovers control-aligned representations from passively observed video lowers the barrier to building behavioral models of agents that did not consent to that observation. The downstream risks are surveillance-adjacent (an actor with access to public video of a target task can pre-train a policy that imitates the demonstrator without ever instrumenting them, and SAM~2.1's prompt interface makes it straightforward to single out specific individuals or assets within a scene) and automation-displacement-adjacent (any reduction in the cost of imitation from video lowers the marginal cost of replacing the imitated work). We see no technical mitigation at the level of the LAM itself; mitigation belongs at the data and deployment layer, through dataset auditing, restrictions on identifying segmentation prompts, and venue-level norms around consent for video used in embodied pre-training.

\section{Implementation Details}
\label{sec:appendix_implementation_details}

\textbf{Observation and stack.} Observations are $64{\times}64{\times}3$ RGB on DCS and $128{\times}128{\times}3$ on DMW. Following \citet{nikulin_latent_2025} we stack $3$ consecutive frames as input to every module, and the IDM additionally consumes $3$ future frames at offset $k \in \{1, \ldots, 10\}$ sampled uniformly per training step (\Cref{sec:method}).

\textbf{Preprocessing.} Pixel observations are centered to $[-\tfrac{1}{2}, \tfrac{1}{2}]$ via $o \leftarrow o/255 - \tfrac{1}{2}$, matching the FDM output range. Segmentation masks are binarized to $\{0, 1\}$ at a $0.5$ threshold so that ground-truth and SAM masks share the same numerical scale. Ground-truth actions are clamped to the environment action space $[-1, 1]^{d_a}$ before being used as supervision, since the simulator already saturates commands at execution time (\Cref{sec:appendix_evaluation_details}).

\textbf{Encoder backbone.} Every convolutional module uses the IMPALA backbone of \citet{espeholt_impala_2018}: three encoder blocks with base widths $[16, 32, 32]$ scaled by a channel multiplier $m$, and $2$ residual blocks per encoder block. Each encoder block applies a $3{\times}3$ convolution, a $3{\times}3$ max-pool of stride $2$, and the residual stack, halving spatial resolution; a final flatten and linear projection produces a $128$-dimensional embedding. Weights are initialized orthogonally.

\textbf{IDM.} Inputs are the $3$ current and $3$ future frames concatenated along the channel dimension, with each frame augmented by its segmentation mask as an extra input channel ($24$ input channels in total at $H{\times}W$). The IMPALA backbone with multiplier $m{=}6$ maps this stack to the continuous latent action $z_t \in \mathbb{R}^{d_z}$ that conditions the FDM. The latent action dimension is $d_z=128$, if not stated otherwise. No quantizer is used between the encoder and the FDM.

\textbf{FDM.} Implemented as the LAOM-Labels world model of \citet{nikulin_latent_2025}: an IMPALA-style encoder with widths $[16, 32, 32]$, multiplier $6$, and $2$ residual blocks per block, applied to the $3$ current frames stacked with their masks ($12$ input channels). The encoder bottleneck is concatenated channel-wise with $z_t$ projected through a linear layer to the bottleneck spatial shape; a symmetric IMPALA decoder of transposed-convolution upsampling blocks followed by residual blocks reconstructs back to input resolution, and a final $1{\times}1$ convolution maps the result to the $3$ RGB channels of $\hat{o}_{t+1}$, squashed to $[-\tfrac{1}{2}, \tfrac{1}{2}]$ by $\tanh / 2$.

\textbf{Stage~2 latent policy.} A fresh IMPALA backbone with multiplier $m{=}4$ takes the $3$ current frames only as input ($9$ channels; no mask, no future frames) and is trained to regress the frozen IDM output via mean-squared error, producing a $d_z$-dimensional latent action.

\textbf{Stage~3 BC actor.} The Stage~2 latent-policy backbone is loaded and frozen, and a two-layer MLP of hidden size $256$ with ReLU activations followed by a linear action head is appended on top of its $d_z$-dimensional embedding. Only the MLP and the action head are trained, with mean-squared error against the ground-truth action. The actor is therefore fully deterministic at inference.

\textbf{Baselines.} For LAPO, LAOM-Labels, LAOF, and LAPO-Slots we adopt each method's published latent-action backbone unchanged. To keep the comparison fair we set the latent action dimension to $d_z=128$ across all methods and use the shared Stage~2 latent policy and Stage~3 BC actor described above for behavior cloning and action decoding. Every reported policy is therefore fully deterministic at inference, and any gap in downstream return reflects the quality of the upstream latent action rather than differences in policy architecture.

\textbf{Training pipelines.} LAPO~\citep{schmidt_learning_2023}, MaskLAM, LAOM-Labels~\citep{nikulin_latent_2025}, and LAOF~\citep{bu_laof_2026} all follow the standard three-stage latent-action-learning pipeline of \Cref{sec:background}: Stage~1 trains the LAM on observation-only data, Stage~2 distills a latent policy via behavior cloning against the frozen IDM output, and Stage~3 fine-tunes a small action decoder on a labeled subset. The four methods differ only in their Stage~1 design choices, for which we refer the reader to the original papers; per-stage hyperparameters are listed in \Cref{sec:appendix_hyperparameters}.

\textbf{LAPO-Slots pipeline.} LAPO-Slots~\citep{klepach_object-centric_2026} expands the standard pipeline to five stages by inserting an object-centric pre-training step before LAM training:
\begin{itemize}
    \item \textit{Stage~1} (slot pre-training): a VideoSAUR encoder is pre-trained on the unlabeled video corpus to convert each frame into a set of object-centric slot representations that separate controllable objects from background distractors.
    \item \textit{Stage~2} (slot probing): on a small labeled subset, a linear probe regresses ground-truth actions from each slot, and the most action-predictive slot per task is retained.
    \item \textit{Stage~3} (LAM pre-training): a LAM is trained on the selected slots, inferring latent actions $z_t$ from transitions and learning a forward dynamics model in slot space.
    \item \textit{Stage~4} (relabeling and behavior cloning): the inferred latent actions serve as pseudo-labels for a behavior-cloning policy that predicts $z_t$ from observations.
    \item \textit{Stage~5} (action decoder fine-tuning): a small labeled set fine-tunes a decoder that maps the latent policy's output to executable actions.
\end{itemize}

\textbf{Hardware.} All experiments run on a single NVIDIA A100 40\,GB GPU. Peak memory stays below $24$\,GB across every method and stage, so the pipeline reproduces on commodity consumer cards. We use bfloat16 mixed precision and \texttt{torch.compile} where applicable.

\textbf{Wall-clock per training run.} \Cref{tab:appendix_walltime_per_run} reports the total wall-clock for a single end-to-end training run, summed across all stages of the pipeline, for one seed on one task of each benchmark.

\begin{table}[h]
\centering
\small
\caption{\textbf{Wall-clock per end-to-end training} run on a single A100 40\,GB GPU, summed across all stages of the pipeline for one seed on one task. LAPO-Slots is the most expensive because of the additional VideoSAUR pre-training stage; MaskLAM is the cheapest non-LAPO method.}
\label{tab:appendix_walltime_per_run}
\begin{tabular}{lcc}
Method & DCS & DMW \\
\midrule
LAPO        & $\approx 6$\,h\,$21$\,m  & $\approx 5$\,h\,$20$\,m \\
LAOM-Labels & $\approx 23$\,h\,$6$\,m  & $\approx 22$\,h\,$0$\,m \\
LAOF        & $\approx 11$\,h\,$10$\,m & $\approx 7$\,h\,$7$\,m  \\
LAPO-Slots  & $\approx 37$\,h\,$31$\,m & $\approx 18$\,h\,$41$\,m \\
MaskLAM     & $\approx 11$\,h\,$8$\,m & $\approx 7$\,h\,$8$\,m  \\
\end{tabular}
\end{table}

\textbf{Total compute budget.} \Cref{tab:appendix_compute_budget} reports the total GPU-hours used in this work, aggregated over every seed, task, ablation, and sweep that contributed to the reported numbers. The MaskLAM row is the largest single contribution because of the additional latent-dimension sweep (\Cref{sec:appendix_q3}), the occlusion sweep (\Cref{sec:appendix_q5}), the mask-mIoU sweep (\Cref{sec:appendix_q6}), and the ablation study (\Cref{sec:appendix_ablation}). SAM~2.1 mask pre-computation is accounted for separately as a one-time cost amortised across every MaskLAM@SAM evaluation; see \Cref{sec:mask_generation} for the underlying throughput numbers.

\begin{table}[H]
\centering
\small
\caption{\textbf{Total GPU-hours used in this work} on a single A100 40\,GB GPU, aggregated across every seed, task, ablation, and sweep that contributed to the reported numbers. SAM~2.1 mask generation is reported as a separate one-time pre-computation cost.}
\label{tab:appendix_compute_budget}
\begin{tabular}{lrr}
Method & Runs & GPU-hours \\
\midrule
LAPO        &   959 & 1{,}128 \\
LAOM-Labels &   588 & 3{,}862 \\
LAOF        &   745 & 1{,}007       \\
LAPO-Slots  &   672 & 2{,}566 \\
MaskLAM     & 2{,}647 & 6{,}843 \\
\midrule
SAM~2.1 mask generation & --- & 309 \\
\midrule
\textbf{Total} & 5{,}615 & 15{,}715 \\
\end{tabular}
\end{table}

\textbf{Code and data release.} We release the full training and evaluation code and the synchronised distractor-free / distracting datasets with both ground-truth and SAM-extracted masks for both benchmarks.

\section{Hyperparameters}
\label{sec:appendix_hyperparameters}

\Cref{tab:appendix_masklam_hyperparameters,tab:appendix_lapo_hyperparameters,tab:appendix_laom_hyperparameters,tab:appendix_laof_hyperparameters,tab:appendix_lapo_slots_hyperparameters} list per-stage hyperparameters for all five methods.
We initialized each baseline from its published hyperparameters and adjusted learning rates, batch sizes, and training budgets where necessary to ensure convergence on our datasets.
Pipeline descriptions are in \Cref{sec:appendix_implementation_details}.
Where two values appear separated by a slash, the first applies to DCS and the second to DMW; otherwise the value is shared.
A dash (---) indicates the hyperparameter does not apply to that stage.

\begin{table}[h]
\centering
\footnotesize
\caption{\textbf{Per-stage hyperparameters for MaskLAM} (see \Cref{sec:appendix_implementation_details} for the pipeline description).}
\begin{tabular}{@{}llccc@{}}
& \textbf{Hyperparameter} & \textbf{Stage 1: LAM} & \textbf{Stage 2: BC} & \textbf{Stage 3: Decoder} \\
\midrule
\multirow{6}{*}{\textit{Optimization}}
 & Optimizer            & AdamW                                       & AdamW                                       & AdamW \\
 & Learning rate        & $6\!\times\!10^{-4}$ / $3\!\times\!10^{-4}$ & $4\!\times\!10^{-4}$ / $2\!\times\!10^{-4}$ & $1\!\times\!10^{-3}$ \\
 & LR schedule          & Cosine annealing                            & Cosine annealing                            & Cosine annealing \\
 & Batch size           & 512 / 128                                   & 512 / 128                                   & 512 / 128 \\
 & Gradient clipping    & ---                                         & ---                                         & 1.0 \\
 & Mixed precision      & bf16                                        & bf16                                        & bf16 \\
\midrule
\multirow{3}{*}{\textit{Training budget}}
 & Epochs / steps       & 2 / 4 epochs                                    & 2 / 4 epochs                                    & 25{,}000 steps \\
 & Block size           & 13                                          & 13                                          & 3 \\
 & Dataset subset       & full                                        & full                                        & 128{,}000 \\
\midrule
\multirow{7}{*}{\textit{Architecture}}
 & Visual encoder       & IMPALA                                      & IMPALA                                      & IMPALA \\
 & Channel multiplier   & 4                                           & 4                                           & 4 \\
 & Encoder channels     & 6                                           & 4                                           & 4 \\
 & Encoder res blocks   & 2                                           & 2                                           & --- \\
 & Latent action dimension $d_z$       & 128                                         & 128                                         & 128 \\
 & Frame stack          & 3                                           & 3                                           & 3 \\
 & Hidden dim (actor)   & ---                                         & ---                                         & 256 \\
\midrule
\multirow{2}{*}{\textit{MaskLAM-specific}}
 & Future obs.\ sampling & \checkmark                                  & \texttimes                                  & --- \\
 & Future obs.\ offset  & 10                                          & 10                                          & --- \\
\midrule
\multirow{3}{*}{\textit{Loss weights}}
 & Obs.\ reconstruction MSE & 1.0                                         & ---                                         & --- \\
 & Action probe MSE         & 0.01                                        & 0.01                                        & --- \\
 & Lat.\ behavior cloning MSE & ---                                       & 1.0                                         & --- \\
 & Behavior cloning MSE     & ---                                         & ---                                         & 1.0 \\
\end{tabular}
\label{tab:appendix_masklam_hyperparameters}
\end{table}

\begin{table}[H]
\centering
\footnotesize
\caption{\textbf{Per-stage hyperparameters for LAPO}~\citep{schmidt_learning_2023} (see \Cref{sec:appendix_implementation_details} for the pipeline description).}
\begin{tabular}{@{}llccc@{}}
& \textbf{Hyperparameter} & \textbf{Stage 1: LAM} & \textbf{Stage 2: BC} & \textbf{Stage 3: Decoder} \\
\midrule
\multirow{6}{*}{\textit{Optimization}}
 & Optimizer            & AdamW                                       & AdamW                                       & AdamW \\
 & Learning rate        & $6\!\times\!10^{-4}$ / $3\!\times\!10^{-4}$ & $4\!\times\!10^{-4}$ / $2\!\times\!10^{-4}$ & $1\!\times\!10^{-3}$ \\
 & LR schedule          & Cosine annealing                            & Cosine annealing                            & Cosine annealing \\
 & Batch size           & 512 / 128                                   & 512 / 128                                   & 512 / 128 \\
 & Gradient clipping    & ---                                         & ---                                         & 1.0 \\
 & Mixed precision      & bf16                                        & bf16                                        & bf16 \\
\midrule
\multirow{3}{*}{\textit{Training budget}}
 & Epochs / steps       & 2 / 4 epochs                                    & 2 / 4 epochs                                    & 25{,}000 steps \\
 & Block size           & 4                                           & 4                                           & 3 \\
 & Dataset subset       & full                                        & full                                        & 128{,}000 \\
\midrule
\multirow{6}{*}{\textit{Architecture}}
 & Visual encoder       & IMPALA                                      & IMPALA                                      & IMPALA \\
 & Channel multiplier   & 4                                           & 4                                           & 4 \\
 & Encoder channels     & 6                                           & 4                                           & 4 \\
 & Encoder res blocks   & 2                                           & 2                                           & --- \\
 & Latent action dimension $d_z$ & 128                                    & 128                                         & 128 \\
 & Hidden dim (actor)   & ---                                         & ---                                         & 256 \\
\midrule
\multirow{3}{*}{\textit{LAPO-specific}}
 & Number of latents    & 4                                           & ---                                         & --- \\
 & Policy block size    & ---                                         & 3                                           & --- \\
 & Future obs.\ offset  & 1                                           & ---                                         & --- \\
\midrule
\multirow{4}{*}{\textit{Loss weights}}
 & Obs.\ reconstruction MSE & 1.0                                         & ---                                         & --- \\
 & Action probe MSE         & 0.01                                        & 0.01                                        & --- \\
 & Lat.\ behavior cloning MSE & ---                                       & 1.0                                         & --- \\
 & Behavior cloning MSE     & ---                                         & ---                                         & 1.0 \\
\end{tabular}
\label{tab:appendix_lapo_hyperparameters}
\end{table}

\begin{table}[H]
\centering
\footnotesize
\caption{\textbf{Per-stage hyperparameters for LAOM-Labels}~\citep{nikulin_latent_2025} (see \Cref{sec:appendix_implementation_details} for the pipeline description).}
\begin{tabular}{@{}llccc@{}}
& \textbf{Hyperparameter} & \textbf{Stage 1: LAM} & \textbf{Stage 2: BC} & \textbf{Stage 3: Decoder} \\
\midrule
\multirow{6}{*}{\textit{Optimization}}
 & Optimizer            & AdamW                                       & AdamW                                       & AdamW \\
 & Learning rate        & $6\!\times\!10^{-4}$ / $3\!\times\!10^{-4}$ & $3\!\times\!10^{-4}$ / $2\!\times\!10^{-4}$ & $1\!\times\!10^{-3}$ \\
  & LR schedule          & Piecewise linear                            & Piecewise linear                            & Piecewise linear \\
 & Batch size           & 512 / 128                                   & 512 / 128                                   & 512 / 128 \\
 & Warmup (epochs)      & 1                                           & 0                                           & 0 \\
 & Weight decay         & 0                                           & 0                                           & 0 \\
\midrule
\multirow{3}{*}{\textit{Training budget}}
 & Epochs / steps        & 3 / 5 epochs                             & 2 / 5 epochs                                    & 20{,}000 steps \\
 & Frame stack          & 3                                           & 3                                           & --- \\
 & Image augmentation   & \checkmark                                  & ---                                         & --- \\
\midrule
\multirow{6}{*}{\textit{Architecture}}
 & Visual encoder       & IMPALA                                      & IMPALA                                      & IMPALA \\
 & Encoder scale        & 6                                           & 32                                          & --- \\
 & Encoder res blocks   & 2                                           & 2                                           & --- \\
 & Hidden dim (MLP)     & ---                                         & ---                                         & 256 \\
 & Latent action dimension $d_z$ & 128                                    & ---                                         & --- \\
 & Obs.\ head dim       & 1024                                        & ---                                         & --- \\
\midrule
\multirow{6}{*}{\textit{LAOM-Labels-specific}}
 & Labeled subset size  & 128{,}000                                     & ---                                         & 128{,}000 \\
 & Labeled batch size   & 512 / 128                                   & ---                                         & --- \\
 & Labeled loss coef.   & $1\!\times\!10^{-3}$                        & ---                                         & --- \\
 & Future obs.\ sampling & \checkmark                                  & \texttimes                                  & --- \\
 & Future obs.\ offset  & 10                                          & ---                                         & --- \\
 & EMA target $\tau$    & $1\!\times\!10^{-3}$                        & ---                                         & --- \\
\end{tabular}
\label{tab:appendix_laom_hyperparameters}
\end{table}

\begin{table}[H]
\centering
\footnotesize
\caption{\textbf{Per-stage hyperparameters for LAOF}~\citep{bu_laof_2026} (see \Cref{sec:appendix_implementation_details} for the pipeline description).}
\begin{tabular}{@{}llccc@{}}
& \textbf{Hyperparameter} & \textbf{Stage 1: LAM} & \textbf{Stage 2: BC} & \textbf{Stage 3: Decoder} \\
\midrule
\multirow{6}{*}{\textit{Optimization}}
 & Optimizer            & AdamW                                           & AdamW                                         & AdamW \\
 & Learning rate        & $3\!\times\!10^{-4}$ / $1.5\!\times\!10^{-4}$  & $2\!\times\!10^{-4}$ / $1\!\times\!10^{-4}$  & $1\!\times\!10^{-3}$ \\
 & LR schedule          & Piecewise linear                                & Piecewise linear                              & Piecewise linear \\
 & Batch size           & 512 / 128                                       & 512 / 128                                     & 512 / 128 \\
 & Gradient clipping    & 2.0                                             & ---                                           & 2.0 \\
 & Mixed precision      & bf16                                            & bf16                                          & bf16 \\
\midrule
\multirow{3}{*}{\textit{Training budget}}
 & Epochs / steps       & 2 / 4 epochs                                    & 2 / 4 epochs                                  & 25{,}000 steps \\
 & Dataset subset       & 128{,}000                                       & 128{,}000                                     & 128{,}000 \\
 & Frame stack          & ---                                             & 3                                             & 3 \\
\midrule
\multirow{7}{*}{\textit{Architecture}}
 & Encoder backbone     & IMPALA                                          & IMPALA                                        & IMPALA \\
 & Optical-flow encoder & RAFT                                            & ---                                           & --- \\
 & World-model scale    & 24                                              & ---                                           & --- \\
 & IDM encoder scale    & 4                                               & ---                                           & --- \\
 & Policy encoder scale & ---                                             & 4                                             & 4 \\
 & Latent action dimension $d_z$ & 128                                    & 128                                           & 128 \\
 & Hidden dim (actor)   & ---                                             & ---                                           & 256 \\
\midrule
\multirow{1}{*}{\textit{LAOF-specific}}
 & Future obs.\ offset  & 1                                               & ---                                           & --- \\
\end{tabular}

\label{tab:appendix_laof_hyperparameters}
\end{table}

\begin{table}[H]
\centering
\footnotesize
\caption{\textbf{Per-stage hyperparameters for LAPO-Slots}~\citep{klepach_object-centric_2026} (see \Cref{sec:appendix_implementation_details} for the pipeline description). Stage~2 (slot probing) has no learnable parameters.}
\begin{tabular}{@{}llccc@{}}
& \textbf{Hyperparameter} & \textbf{Stage 1: VideoSAUR} & \textbf{Stage 2: Probing} & \textbf{Stage 3: LAM} \\
\midrule
\multirow{6}{*}{\textit{Optimization}}
 & Optimizer            & AdamW                & ---               & AdamW \\
 & Learning rate        & $3\!\times\!10^{-4}$ & ---               & $3\!\times\!10^{-4}$ / $1.5\!\times\!10^{-4}$ \\
 & LR schedule          & exp.\ + warmup       & ---               & --- \\
 & Warmup               & 1{,}250 steps        & ---               & 0 epochs \\
 & Batch size           & 256 / 128            & ---               & 256 / 128 \\
 & Gradient clipping    & 0.05                 & ---               & 1.0 \\
\midrule
\multirow{2}{*}{\textit{Training budget}}
 & Epochs / steps       & 25{,}000 steps       & ---               & 1 / 4 epochs \\
 & Dataset subset       & full                 & ---               & full \\
\midrule
\multirow{10}{*}{\textit{Architecture}}
 & DINO backbone        & ViT-B/14 (DINOv2)    & ---               & ViT-B/14 (DINOv2) \\
 & Slot dim             & 128                  & ---               & 128 \\
 & Num.\ slots          & 4 / 3                & ---               & 4 / 3 \\
 & Slot iterations      & 3                    & ---               & --- \\
 & Sim.\ loss / temp.\  & $0.1$ / $0.075$      & ---               & --- \\
 & Hidden dim           & ---                  & ---               & 1024 \\
 & Residual blocks      & ---                  & ---               & 3 \\
 & Latent action dimension $d_z$ & ---         & ---               & 128 \\
 & Frame stack          & ---                  & ---               & 1 \\
\midrule
\multirow{4}{*}{\textit{LAPO-Slots-specific}}
 & PCA components       & ---                  & 32                & --- \\
 & Probing model        & ---                  & Linear regression & --- \\
 & Cross-validation     & ---                  & 5-fold            & --- \\
 & Future obs.\ offset  & ---                  & ---               & 10 \\
\end{tabular}

\vspace{1em}

\begin{tabular}{@{}llcc@{}}
& \textbf{Hyperparameter} & \textbf{Stage 4: BC} & \textbf{Stage 5: Decoder} \\
\midrule
\multirow{6}{*}{\textit{Optimization}}
 & Optimizer            & AdamW                                         & AdamW \\
 & Learning rate        & $3\!\times\!10^{-4}$ / $1.5\!\times\!10^{-4}$ & $1\!\times\!10^{-3}$ \\
 & LR schedule          & ---                                           & --- \\
 & Warmup               & 0 epochs                                      & 0 epochs \\
 & Batch size           & 256 / 128                                     & 256 / 128 \\
 & Gradient clipping    & 1.0                                           & --- \\
\midrule
\multirow{2}{*}{\textit{Training budget}}
 & Epochs / steps       & 1 / 4 epochs                                  & 25{,}000 steps \\
 & Dataset subset       & 128{,}000                                     & 128{,}000 \\
\midrule
\multirow{5}{*}{\textit{Architecture}}
 & Visual encoder       & IMPALA                                        & IMPALA \\
 & Channel multiplier   & 4                                             & --- \\
 & Residual blocks      & 2                                             & --- \\
 & Latent action dimension $d_z$ & 128                                  & 128 \\
 & Hidden dim (actor)   & 256                                           & 256 \\
\end{tabular}

\label{tab:appendix_lapo_slots_hyperparameters}
\end{table}

\section{Expert Results}
\label{sec:appendix_expert_results}

We report the performance of the expert policies used to collect the offline trajectories that train all latent action models in this work (\Cref{sec:appendix_dataset}). \Cref{tab:appendix_expert_returns_dcs} lists the average undiscounted episode return of each DCS expert, and \Cref{tab:appendix_expert_success_dmw} lists the average task success rate of each DMW expert, both measured over the same evaluation protocol used for the downstream policies (\Cref{sec:appendix_evaluation_details}). These numbers serve as the upper reference point when computing the Normalized Return (NR) on DCS and the Normalized Success Rate (NSR) on DMW: a value of $1.0$ matches expert performance on a given task, and $0.0$ corresponds to a random policy. Reporting expert results explicitly makes the absolute scale of our normalized metrics interpretable and ensures that any apparent gap to expert performance reflects the difficulty of the task rather than an unattainable normalization target.

\begin{table}[H]
\centering
\small
\caption{\textbf{Expert returns of data collection policy used for normalization on the DCS.}}
\label{tab:appendix_expert_returns_dcs}
\begin{tabular}{lc}
Task & Return \\
\midrule
cheetah-run & 837.67 \\
hopper-hop & 307.33 \\
humanoid-walk & 616.52 \\
walker-run & 738.37 \\
\end{tabular}
\end{table}

\begin{table}[H]
\centering
\small
\caption{\textbf{Expert success rates of data collection policy used for normalization on DMW.}}
\label{tab:appendix_expert_success_dmw}
\begin{tabular}{lc}
Task & Success Rate \\
\midrule
button-press-topdown & 1.000 \\
door-open & 0.917 \\
drawer-close & 1.000 \\
drawer-open & 1.000 \\
pick-place & 1.000 \\
peg-insert-side & 0.772 \\
push & 1.000 \\
reach & 1.000 \\
window-open & 1.000 \\
window-close & 1.000 \\
\end{tabular}
\end{table}

\section{Evaluation Details}
\label{sec:appendix_evaluation_details}

All numbers reported in the paper are computed on the held-out evaluation split of the released datasets (\Cref{sec:appendix_dataset}), averaged over three random seeds, with 100 rollout episodes per seed for downstream performance. The remainder of this section makes that protocol explicit, gives the formal definitions of the three metrics summarized in \Cref{sec:experiments}, and lists the per-task action variances that normalize the probe MSE; hardware and wall-clock times are reported in \Cref{sec:appendix_implementation_details}.

\textbf{Evaluation data.} Every metric is computed on the held-out split of the released dataset (\Cref{sec:appendix_dataset}): 1M transitions on DCS and 100k on DMW. Held-out trajectories sample distractor backgrounds exclusively from the DAVIS test split, while training trajectories sample exclusively from the DAVIS train split, so the two distractor-video sets are disjoint at the clip level. LAOM~\citep{nikulin_latent_2025} reports two numbers per task, an in-distribution number that reuses training DAVIS clips at evaluation time and a separate, harder ``OOD'' number on held-out clips; every number we report corresponds to that OOD setting. We treat this as the only fair protocol, since an in-distribution evaluation conflates memorization of the background videos with generalization to distractor variation, which is the property the method is supposed to provide.

\textbf{Action clipping.} As stated in \Cref{sec:experiments}, we clip ground-truth actions to the environment action space $[-1,1]^{d_a}$ prior to probe regression and behavior-cloning fine-tuning. The simulator already clamps commands at execution time, so any two distinct unclipped values that share a clipped image produce identical state transitions; regressing against unclipped targets therefore introduces a many-to-one ambiguity that penalizes models that correctly recover the executed action. Our action-probe MSE values are consequently not directly comparable to LAOM~\citep{nikulin_latent_2025}, which evaluates on unclipped targets.

\textbf{Normalized linear action probe MSE (NMSE).} Let $d_a$ denote the ground-truth action dimension and $d_z$ the latent action dimension. For each task $\tau$ we fit a linear probe consisting of a weight matrix $W \in \mathbb{R}^{d_a \times d_z}$ and a bias vector $b \in \mathbb{R}^{d_a}$ that maps a frozen latent action $z_t \in \mathbb{R}^{d_z}$ to the clipped ground-truth action $a_t \in \mathbb{R}^{d_a}$, with no gradient flowing back into the latent action model. The probe is fit on the training split and evaluated on the held-out split; we report
\begin{equation}
    \mathrm{NMSE}(\tau) \;=\; \frac{\mathbb{E}_{\mathcal{D}_{\mathrm{eval}}^{\tau}}\bigl[\lVert W z_t + b - a_t \rVert_2^2\bigr]}{\mathrm{Var}_{\mathcal{D}^{\tau}}(a)},
    \qquad
    \mathrm{Var}(a) \;=\; \frac{1}{d_a}\sum_{i=1}^{d_a} \mathrm{Var}\!\left(a^{(i)}\right),
    \label{eq:nmse}
\end{equation}
where the denominator is the mean per-dimension variance of the clipped ground-truth actions on that task's training split, listed in \Cref{tab:appendix_action_variance}. Under this normalization, $\mathrm{NMSE}=1$ corresponds to a probe that predicts the mean action and $\mathrm{NMSE}=0$ to perfect recovery, so the metric is comparable across tasks with different action scales.

\textbf{Normalized return (NR).} On DCS we generate rollouts with the decoded policy of Stage~3 for 100 episodes and report
\begin{equation}
    \mathrm{NR}(\tau) \;=\; \frac{R_{\pi}(\tau)}{R_{\mathrm{exp}}(\tau)},
    \label{eq:nr}
\end{equation}
where $R_{\pi}(\tau)$ is the mean undiscounted episode return of the learned policy and $R_{\mathrm{exp}}(\tau)$ is the expert return from \Cref{tab:appendix_expert_returns_dcs}. A value of $1$ matches the data-collection expert.

\textbf{Normalized success rate (NSR).} On DMW we generate rollouts with the decoded policy of Stage~3 for 100 episodes and report
\begin{equation}
    \mathrm{NSR}(\tau) \;=\; \frac{S_{\pi}(\tau)}{S_{\mathrm{exp}}(\tau)},
    \label{eq:nsr}
\end{equation}
where $S_{\pi}(\tau)$ is the success fraction of the learned policy and $S_{\mathrm{exp}}(\tau)$ the expert success rate from \Cref{tab:appendix_expert_success_dmw}.

\textbf{Aggregation.} For every task we first compute the mean and standard deviation across the three seeds. Per-benchmark numbers are obtained by averaging the per-task means, and separately averaging the per-task standard deviations, across the tasks of that benchmark. When we report a single number across both benchmarks, we average the two per-benchmark aggregates, so that DCS and DMW contribute equally regardless of the difference in the number of tasks.

\textbf{Per-task action variance.} \Cref{tab:appendix_action_variance} lists $\mathrm{Var}(a)$ for every evaluated task, which serves as the denominator of \eqref{eq:nmse}. Variances are computed on the clipped ground-truth actions of the released dataset and agree to numerical precision between the training and held-out splits, so the split subscript is dropped in \eqref{eq:nmse}; we report the value computed on the larger training split. They are also shared between the distractor-free and distracting variants of each benchmark, since the action stream is identical by construction in our synchronized data collection (\Cref{sec:appendix_dataset}).

\begin{table}[H]
\centering
\small
\caption{\textbf{Per-task action variance} $\mathrm{Var}(a) = \tfrac{1}{d_a}\sum_{i} \mathrm{Var}(a^{(i)})$, computed on clipped ground-truth actions over the training split of the released dataset. Values are shared between the distractor-free and distracting variants of each benchmark, because the action stream is identical by construction (\Cref{sec:appendix_dataset}). Used as the denominator of NMSE in \eqref{eq:nmse}.}
\label{tab:appendix_action_variance}
\begin{tabular}{lc}
Task & $\mathrm{Var}(a)$ \\
\midrule
\multicolumn{2}{l}{\textit{DCS}} \\
\midrule
cheetah-run & 0.6787 \\
hopper-hop & 0.7454 \\
humanoid-walk & 0.4221 \\
walker-run & 0.6729 \\
\midrule
\multicolumn{2}{l}{\textit{DMW}} \\
\midrule
button-press-topdown & 0.2085 \\
door-open & 0.4266 \\
drawer-close & 0.2466 \\
drawer-open & 0.1483 \\
peg-insert-side & 0.2497 \\
pick-place & 0.2286 \\
push & 0.1311 \\
reach & 0.0372 \\
window-close & 0.3650 \\
window-open & 0.2979 \\
\end{tabular}
\end{table}

\section{Failure Modes of LAOF and LAPO-Slots}
\label{sec:appendix_laof_lapo_slots_result_discussion}

\begin{figure}[H]
    \centering
    \includegraphics[width=0.88\linewidth]{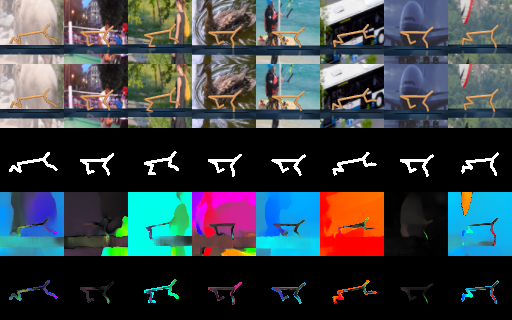}
    \caption{\textbf{LAOF flow target construction on DCS cheetah-run (distractor).} Eight transitions sampled from the dataset. Rows top-to-bottom: current observation $o_t$; next observation $o_{t+1}$; supervisory mask $M_{t+1}$; raw RAFT optical flow between $o_t$ and $o_{t+1}$; masked flow $M_{t+1} \odot \text{RAFT}(o_t, o_{t+1})$, the target for LAOF's auxiliary flow decoder loss. Distractor motion in the dynamic background produces large flow magnitudes that bleed into the cheetah mask along its boundary, so the masked target inside the mask is dominated by exogenous motion rather than agent motion. The flow decoder is consequently asked to predict distractor flow conditioned on $z_t$, which pushes $z_t$ to encode it.}
    \label{fig:appendix_laof_dcs}
\end{figure}

On our distractor benchmarks, LAOF and LAPO-Slots perform on par with or only marginally improve over LAPO (\Cref{tab:q1_combined_aggregate}), in contrast to the strong results reported in the original papers. To explain this gap, we trace each method back to its mask-based prior and identify a distinct failure mode in each case. Both could in principle be alleviated by replacing the underlying component, the optical flow model for LAOF and the object-centric encoder for LAPO-Slots, with a stronger alternative; the conclusions here concern the priors as currently instantiated, not the methods in their idealized form.

\begin{figure}[H]
    \centering
    \includegraphics[width=0.88\linewidth]{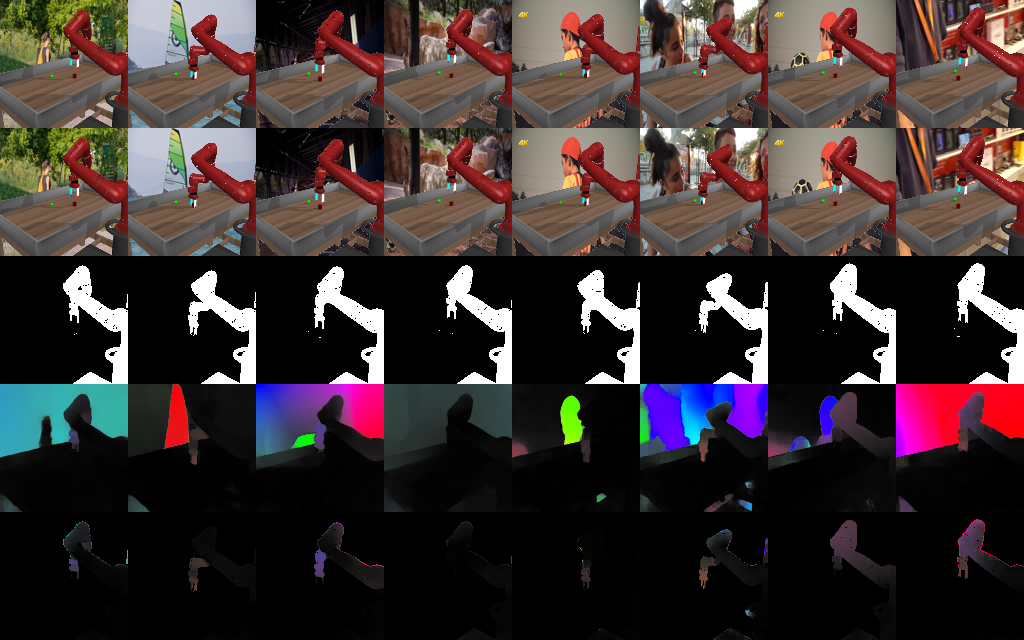}
    \caption{\textbf{LAOF flow target construction on DMW push (distractor).} Layout matches \Cref{fig:appendix_laof_dcs}. The contaminated boundary region covers a smaller fraction of the masked area than on DCS, not because the mask is sharper but because the agent occupies more pixels at DMW's higher rendering resolution. The distractor flow magnitude is still high relative to the slow robot arm, so where bleed does occur it still dominates the local target.}
    \label{fig:appendix_laof_dmw}
\end{figure}

\textbf{LAOF: distractor optical flow bleeds into the agent mask.} LAOF supervises an auxiliary flow decoder on RAFT optical flow restricted to the agent mask, so that only agent-driven motion shapes the latent action $z_t$. \Cref{fig:appendix_laof_dcs} shows the construction on DCS cheetah-run with distractors. Distractor regions adjacent to the cheetah produce large RAFT flow magnitudes that, after multiplying by $M_{t+1}$, contaminate the target inside the mask, so the bottom-row flow the decoder is asked to predict is mostly exogenous motion. The decoder must therefore explain that motion conditioned on $z_t$, which forces $z_t$ to carry distractor information just as in unmasked LAPO. \Cref{fig:appendix_laof_dmw} shows the same construction on DMW push with distractors. The bleed covers a smaller area, but not because the mask is sharper: at DMW's higher rendering resolution the agent simply occupies more pixels, so the contaminated boundary region is a smaller fraction of the masked area. The distractor flow magnitude remains high relative to the slow robot arm, so wherever the bleed does occur it still dominates the local target.

\textbf{LAPO-Slots: agent identity is not stable across slot indices.} The full LAPO-Slots pipeline is described in \Cref{sec:appendix_implementation_details}; here we focus on the three stages relevant to the failure mode. Stage~1 (VideoSAUR) decomposes each frame into $K$ object-centric slots; Stage~2 selects the single most action-predictive slot index per task by linearly probing each slot's features against ground-truth actions; Stage~3 trains the LAM on the features of that one selected slot. The pipeline therefore depends on the agent occupying the same slot index across all samples in the dataset, since the Stage~2 selection is a single fixed index applied uniformly at Stage~3 training time: whenever VideoSAUR routes the agent to a different slot on a sample, the LAM input on that sample contains no agent. We use $K{=}4$ slots on DCS and $K{=}3$ slots on DMW, following the original protocol~\citep{klepach_object-centric_2026}.

\begin{figure}[H]
    \centering
    \includegraphics[width=0.88\linewidth]{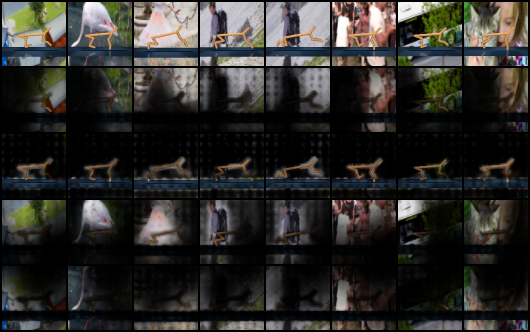}
    \caption{\textbf{LAPO-Slots VideoSAUR slot decomposition on DCS cheetah-run (distractor), stable run.} Eight transitions sampled from the dataset. Top row: input observation $o_t$. Rows 2--5: per-slot masked observations for slots~1--4. Slot~2 (row~3) consistently attends to the cheetah on every sampled frame; the remaining slots absorb distractor background structure. A Stage~2 selection of slot~2 therefore retains the agent on every dataset sample and the downstream LAM input is well defined.}
    \label{fig:appendix_lapo_slots_dcs_good}
\end{figure}

\begin{figure}[H]
    \centering
    \includegraphics[width=0.88\linewidth]{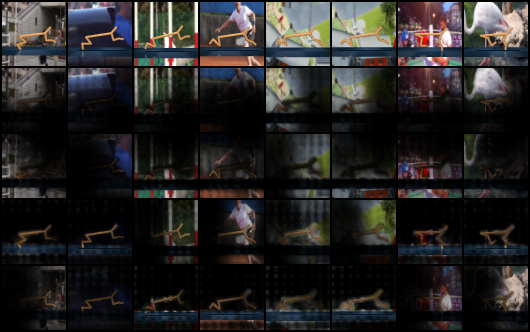}
    \caption{\textbf{LAPO-Slots VideoSAUR slot decomposition on DCS cheetah-run (distractor), failed run.} Layout matches \Cref{fig:appendix_lapo_slots_dcs_good}. Cheetah identity is split between slots~3 and~4 (rows~4 and~5) across the sampled batch, so no single slot index covers the agent on every sample. Because Stage~3 trains the LAM on the features of one fixed slot, the agent is dropped from the LAM input on the samples where VideoSAUR assigns it to the other slot.}
    \label{fig:appendix_lapo_slots_dcs_bad}
\end{figure}

\begin{figure}[H]
    \centering
    \includegraphics[width=0.88\linewidth]{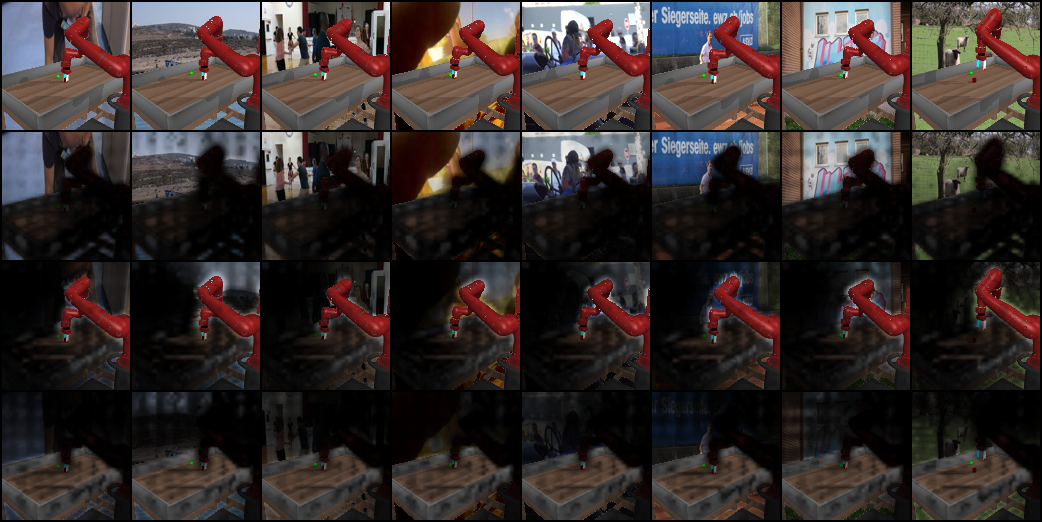}
    \caption{\textbf{LAPO-Slots VideoSAUR slot decomposition on DMW push (distractor), stable run.} Eight transitions sampled from the dataset. Top row: input observation $o_t$. Rows 2--4: per-slot masked observations for slots~1--3. Slot~2 (row~3) consistently attends to the robot arm on every sampled frame; slots~1 and~3 absorb table and distractor regions.}
    \label{fig:appendix_lapo_slots_dmw_good}
\end{figure}

\begin{figure}[H]
    \centering
    \includegraphics[width=0.88\linewidth]{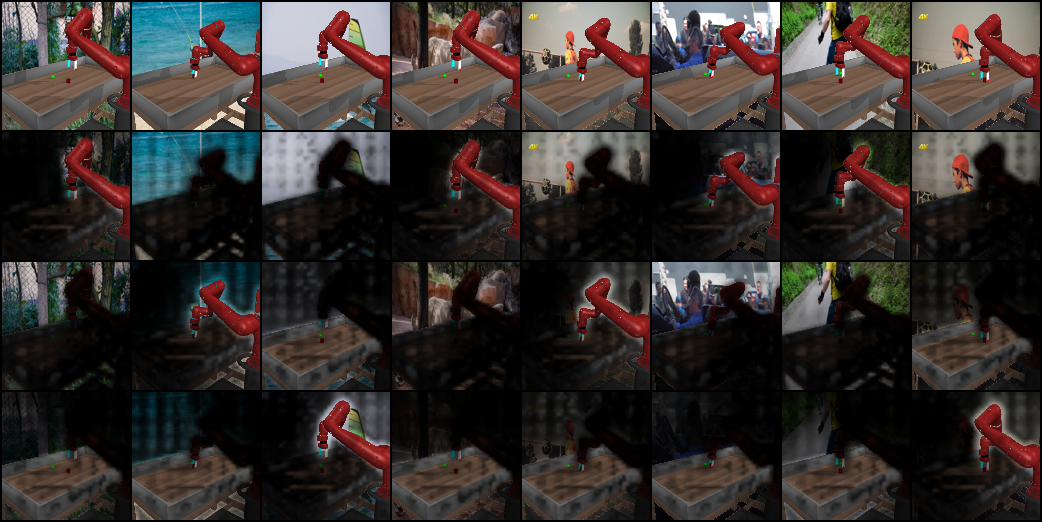}
    \caption{\textbf{LAPO-Slots VideoSAUR slot decomposition on DMW push (distractor), failed run.} Layout matches \Cref{fig:appendix_lapo_slots_dmw_good}. Robot-arm identity is distributed across slots~1, 2, and~3 across the sampled batch, so any fixed Stage~2 slot selection loses the arm on a subset of samples and the downstream LAM is trained on intermittently empty input.}
    \label{fig:appendix_lapo_slots_dmw_bad}
\end{figure}

\Cref{fig:appendix_lapo_slots_dcs_good} shows a representative \emph{stable} run on DCS cheetah-run (distractor): slot~2 attends to the cheetah across all eight sampled transitions while the remaining slots absorb background structure, so a Stage~2 selection of slot~2 retains the agent on every sample. \Cref{fig:appendix_lapo_slots_dcs_bad} shows the failure mode on the same task: cheetah identity is not bound to a single slot index, but is split across slot~3 and slot~4 across the sampled batch, so no fixed slot index covers the agent on every sample. \Cref{fig:appendix_lapo_slots_dmw_good,fig:appendix_lapo_slots_dmw_bad} show the same pattern on DMW push (distractor): a stable run where slot~2 carries the robot arm on every sample, and a failed run where arm identity is distributed across slots~1, 2, and 3 across the sampled batch. Because Stage~3 commits to a single slot index, this instability translates directly into LAM input dropout on the samples where VideoSAUR routes the agent elsewhere.

\section{World Model Visualizations}
\label{sec:appendix_world_model_visualizations}

The FDM (world model) in Stage~1 of the three-stage MaskLAM and LAPO training pipeline is trained jointly with the IDM (\Cref{sec:method}) and provides the only learning signal that shapes the latent action $z_t$: $z_t$ must explain whatever the world model is asked to reconstruct beyond what the prior frames already determine. Visualizing what each method's world model actually predicts therefore exposes what $z_t$ has been forced to encode.

\begin{figure}[H]
    \centering
    \includegraphics[width=\linewidth]{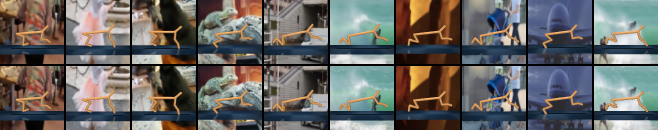}
    \caption{\textbf{LAPO world model predictions on DCS (distractor).} Top row: predicted next observation $\hat{o}_{t+1}$. Bottom row: ground-truth next observation $o_{t+1}$. Ten consecutive transitions sampled from training. The unmasked reconstruction loss yields pixel-accurate reproduction of the dynamic distractor video, camera shake, and agent color randomization, evidence that $z_t$ must carry distractor information for the world model to satisfy its objective.}
    \label{fig:wm_lapo_dcs}
\end{figure}

\begin{figure}[H]
    \centering
    \includegraphics[width=\linewidth]{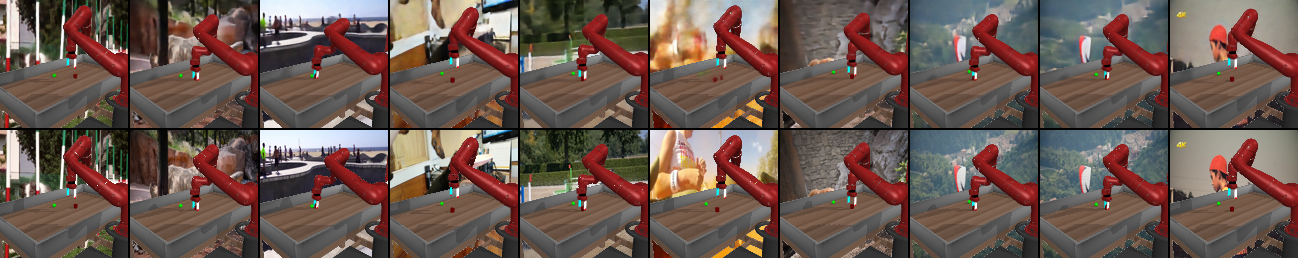}
    \caption{\textbf{LAPO world model predictions on DMW (distractor).} Layout matches \Cref{fig:wm_lapo_dcs}. Reconstructions track the distractor background with the same fidelity as the agent, confirming that the failure mode is benchmark-independent.}
    \label{fig:wm_lapo_dmw}
\end{figure}

\Cref{fig:wm_lapo_dcs,fig:wm_lapo_dmw} show LAPO's world model on DCS (distractor) and DMW (distractor). Reconstructions are nearly indistinguishable from the ground-truth next observation: the agent, the dynamic distractor video behind it, the camera shake, and the color randomization are all recovered at the pixel level. This is exactly what the unmasked reconstruction loss rewards. The world model has no way to know that the distractor video is exogenous, so it spends representational capacity on tracking it, and the IDM is pushed to write frame-to-frame distractor variation that the prior frames cannot predict into $z_t$. The pixel-perfect reconstruction is direct visual evidence of the failure mode that motivates our method.

\begin{figure}[H]
    \centering
    \includegraphics[width=\linewidth]{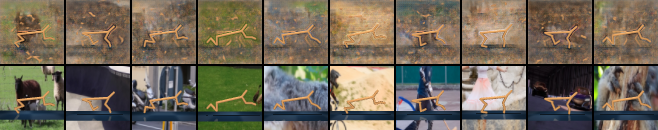}
    \caption{\textbf{MaskLAM world model predictions on DCS (distractor).} Top row: predicted next observation $\hat{o}_{t+1}$. Bottom row: ground-truth next observation $o_{t+1}$. With the reconstruction loss restricted to the supervisory mask $M_{t+1}$, the world model recovers the agent sharply while everything outside the mask collapses into a low-frequency texture with boundary artifacts.}
    \label{fig:wm_slapo_dcs}
\end{figure}

\begin{figure}[H]
    \centering
    \includegraphics[width=\linewidth]{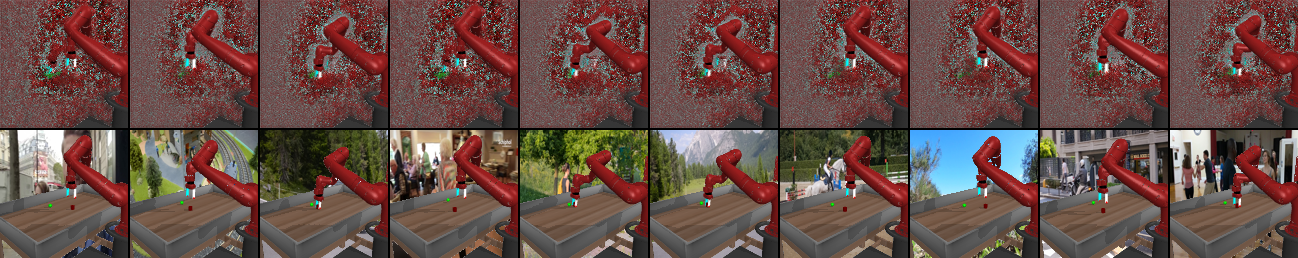}
    \caption{\textbf{MaskLAM world model predictions on DMW (distractor).} Layout matches \Cref{fig:wm_slapo_dcs}. The agent and its contact surfaces are reconstructed cleanly; the masked-out background is reconstructed only to a coarse extent, mirroring the DCS behavior.}
    \label{fig:wm_slapo_dmw}
\end{figure}

\Cref{fig:wm_slapo_dcs,fig:wm_slapo_dmw} show MaskLAM's world model on the same benchmarks. The agent and its immediate vicinity are reconstructed cleanly, while everything outside the supervisory mask collapses into a low-frequency texture with visible artifacts at the agent boundary. This is the intended behavior: with the loss masked to $M_{t+1}$, the world model receives no gradient for distractor pixels and stops modeling them, so $z_t$ is no longer required to carry distractor information. The artifacts outside the mask are not a defect but the visible signature of the bottleneck we impose on the latent action.

\section{Masking Improves Latent Action Quality}
\label{sec:appendix_q1}

Per-task breakdown of the aggregated NMSE in \Cref{tab:q1_combined_aggregate}; metric definition in \Cref{sec:appendix_evaluation_details}.

\textbf{Distractor setting.} MaskLAM@GT and MaskLAM@SAM attain the lowest NMSE among label-free baselines on all four DCS tasks (\Cref{tab:q1_mse_distracting_control_suite}) and all ten DMW tasks (\Cref{tab:q1_mse_meta_world_distractor}), often by a $2$--$3\times$ margin, and match LAOM-Labels on the majority of tasks. The two MaskLAM variants agree within one standard deviation on every task.

\textbf{Distractor-free setting.} On DCS (\Cref{tab:q1_mse_distracting_control_suite}, columns 5--8), MaskLAM matches or exceeds every label-free baseline and lands near the privileged LAOM-Labels. On DMW (\Cref{tab:q1_mse_meta_world_vanilla}), MaskLAM ties LAPO, which is the strongest baseline in the absence of distractors. Without exogenous variance dominating the reconstruction target, the unmasked loss already places most gradient weight on agent-driven change.

\begin{table}[H]
\centering
\renewcommand{\arraystretch}{1.15}
\setlength{\tabcolsep}{4pt}
\resizebox{\textwidth}{!}{%
\begin{tabular}{lcccccccc}
\multicolumn{9}{c}{\textbf{Distracting Control Suite (NMSE~$\downarrow$)}} \\
\midrule
 & \multicolumn{4}{c}{\textbf{Distractor}} & \multicolumn{4}{c}{\textbf{Vanilla}} \\
\cmidrule(lr){2-5}\cmidrule(lr){6-9}
\textbf{Method} & \textbf{cheetah-run} & \textbf{hopper-hop} & \textbf{humanoid-walk} & \textbf{walker-run} & \textbf{cheetah-run} & \textbf{hopper-hop} & \textbf{humanoid-walk} & \textbf{walker-run} \\
\midrule
LAPO & $0.3836_{\pm 0.0220}$ & $0.6433_{\pm 0.0671}$ & $0.6766_{\pm 0.0382}$ & $0.3423_{\pm 0.0071}$ & $0.1484_{\pm 0.0019}$ & $0.1959_{\pm 0.0036}$ & $0.5298_{\pm 0.2591}$ & $0.3142_{\pm 0.2031}$ \\
LAOM-Labels & $0.1652_{\pm 0.0020}$ & $0.1933_{\pm 0.0067}$ & $\mathbf{0.3112_{\pm 0.0025}}$ & $0.1943_{\pm 0.0026}$ & $0.1478_{\pm 0.0009}$ & $\mathbf{0.1904_{\pm 0.0156}}$ & $\mathbf{0.3379_{\pm 0.0072}}$ & $\mathbf{0.1663_{\pm 0.0006}}$ \\
LAOF & $0.3236_{\pm 0.0143}$ & $0.4863_{\pm 0.0303}$ & $0.5719_{\pm 0.0174}$ & $0.6408_{\pm 0.0157}$ & $0.5546_{\pm 0.0162}$ & $0.7345_{\pm 0.0443}$ & $0.6322_{\pm 0.0503}$ & $0.6348_{\pm 0.0696}$ \\
LAPO-Slots & $0.6615_{\pm 0.0212}$ & $0.6950_{\pm 0.0051}$ & $0.7270_{\pm 0.0052}$ & $0.5444_{\pm 0.0859}$ & $0.2892_{\pm 0.0314}$ & $0.3613_{\pm 0.0066}$ & $0.4673_{\pm 0.0088}$ & $0.2884_{\pm 0.0020}$ \\
\midrule
\textbf{MaskLAM@GT} & $0.1171_{\pm 0.0007}$ & $\mathbf{0.1853_{\pm 0.0256}}$ & $0.3531_{\pm 0.0059}$ & $\mathbf{0.1692_{\pm 0.0049}}$ & $0.0910_{\pm 0.0017}$ & $0.2093_{\pm 0.0031}$ & $0.3380_{\pm 0.0041}$ & $0.1811_{\pm 0.0163}$ \\
\textbf{MaskLAM@SAM} & $\mathbf{0.1148_{\pm 0.0040}}$ & $0.2601_{\pm 0.1434}$ & $0.3852_{\pm 0.0007}$ & $0.1704_{\pm 0.0008}$ & $\mathbf{0.0859_{\pm 0.0028}}$ & $0.2147_{\pm 0.0245}$ & $0.3555_{\pm 0.0076}$ & $0.1956_{\pm 0.0035}$ \\
\end{tabular}%
}

\caption{\textbf{Per-task NMSE on DCS in distractor (columns 1--4) and distractor-free (columns 5--8) settings.} MaskLAM@GT and MaskLAM@SAM attain the lowest NMSE among label-free baselines under distractors and surpass LAOM-Labels on three of four tasks; without distractors MaskLAM matches or exceeds every label-free baseline and lands within noise of LAOM-Labels.}
\label{tab:q1_mse_distracting_control_suite}
\end{table}

\begin{table}[H]
\centering
\renewcommand{\arraystretch}{1.15}
\setlength{\tabcolsep}{4pt}
\resizebox{\textwidth}{!}{%
\begin{tabular}{lcccccccccc}
\multicolumn{11}{c}{\textbf{Distracting Meta-World - Vanilla (NMSE~$\downarrow$)}} \\
\midrule
\textbf{Method} & \textbf{reach} & \textbf{push} & \textbf{pick-place} & \textbf{door-open} & \textbf{drawer-open} & \textbf{drawer-close} & \textbf{button-press-topdown} & \textbf{peg-insert-side} & \textbf{window-open} & \textbf{window-close} \\
\midrule
LAPO & $\mathbf{0.0773_{\pm 0.0086}}$ & $0.1332_{\pm 0.0076}$ & $\mathbf{0.1105_{\pm 0.0015}}$ & $\mathbf{0.0557_{\pm 0.0009}}$ & $\mathbf{0.0399_{\pm 0.0018}}$ & $0.0124_{\pm 0.0012}$ & $0.0358_{\pm 0.0034}$ & $0.2174_{\pm 0.0068}$ & $0.0466_{\pm 0.0024}$ & $\mathbf{0.0129_{\pm 0.0014}}$ \\
LAOM-Labels & $0.1130_{\pm 0.0120}$ & $0.1583_{\pm 0.0146}$ & $0.1273_{\pm 0.0073}$ & $0.1047_{\pm 0.0103}$ & $0.0986_{\pm 0.0193}$ & $0.0931_{\pm 0.0065}$ & $0.1068_{\pm 0.0097}$ & $0.1983_{\pm 0.0100}$ & $0.0789_{\pm 0.0034}$ & $0.0584_{\pm 0.0058}$ \\
LAOF & $0.1254_{\pm 0.0049}$ & $0.3574_{\pm 0.0437}$ & $0.2276_{\pm 0.0235}$ & $0.2055_{\pm 0.1099}$ & $0.1483_{\pm 0.0398}$ & $0.1249_{\pm 0.0104}$ & $0.1680_{\pm 0.0855}$ & $0.3118_{\pm 0.0223}$ & $0.1192_{\pm 0.0191}$ & $0.0605_{\pm 0.0113}$ \\
LAPO-Slots & $0.1117_{\pm 0.0046}$ & $0.2166_{\pm 0.0118}$ & $0.1675_{\pm 0.0049}$ & $0.1245_{\pm 0.0081}$ & $0.0757_{\pm 0.0031}$ & $0.0540_{\pm 0.0041}$ & $0.0447_{\pm 0.0031}$ & $\mathbf{0.1979_{\pm 0.0095}}$ & $0.0678_{\pm 0.0053}$ & $0.0292_{\pm 0.0058}$ \\
\midrule
\textbf{MaskLAM@GT} & $0.0837_{\pm 0.0063}$ & $0.1376_{\pm 0.0047}$ & $0.1290_{\pm 0.0093}$ & $0.0763_{\pm 0.0017}$ & $0.0439_{\pm 0.0019}$ & $0.0137_{\pm 0.0026}$ & $0.0341_{\pm 0.0018}$ & $0.2417_{\pm 0.0079}$ & $\mathbf{0.0399_{\pm 0.0014}}$ & $0.0155_{\pm 0.0030}$ \\
\textbf{MaskLAM@SAM} & $0.0794_{\pm 0.0057}$ & $\mathbf{0.1271_{\pm 0.0018}}$ & $0.1259_{\pm 0.0041}$ & $0.0667_{\pm 0.0029}$ & $0.0468_{\pm 0.0021}$ & $\mathbf{0.0108_{\pm 0.0008}}$ & $\mathbf{0.0292_{\pm 0.0021}}$ & $0.2354_{\pm 0.0056}$ & $0.0448_{\pm 0.0038}$ & $0.0131_{\pm 0.0012}$ \\
\end{tabular}%
}

\caption{\textbf{Per-task NMSE on DMW (vanilla).} MaskLAM ties LAPO, the strongest label-free baseline without distractors, confirming that loss masking does not degrade alignment in the absence of exogenous variance.}
\label{tab:q1_mse_meta_world_vanilla}
\end{table}

\begin{table}[H]
\centering
\renewcommand{\arraystretch}{1.15}
\setlength{\tabcolsep}{4pt}
\resizebox{\textwidth}{!}{%
\begin{tabular}{lcccccccccc}
\multicolumn{11}{c}{\textbf{Distracting Meta-World - Distractor (NMSE~$\downarrow$)}} \\
\midrule
\textbf{Method} & \textbf{reach} & \textbf{push} & \textbf{pick-place} & \textbf{door-open} & \textbf{drawer-open} & \textbf{drawer-close} & \textbf{button-press-topdown} & \textbf{peg-insert-side} & \textbf{window-open} & \textbf{window-close} \\
\midrule
LAPO & $0.2906_{\pm 0.0087}$ & $0.4812_{\pm 0.0028}$ & $0.5470_{\pm 0.0110}$ & $0.2312_{\pm 0.0063}$ & $0.2612_{\pm 0.0087}$ & $0.1715_{\pm 0.0067}$ & $0.2978_{\pm 0.0131}$ & $0.5117_{\pm 0.0176}$ & $0.2021_{\pm 0.0140}$ & $0.0987_{\pm 0.0051}$ \\
LAOM-Labels & $0.1610_{\pm 0.0027}$ & $0.2065_{\pm 0.0060}$ & $0.1409_{\pm 0.0030}$ & $0.1407_{\pm 0.0124}$ & $0.1143_{\pm 0.0183}$ & $0.1335_{\pm 0.0138}$ & $0.1075_{\pm 0.0029}$ & $0.2496_{\pm 0.0065}$ & $0.0916_{\pm 0.0067}$ & $0.0751_{\pm 0.0050}$ \\
LAOF & $0.3392_{\pm 0.0244}$ & $0.5003_{\pm 0.0645}$ & $0.4891_{\pm 0.0497}$ & $0.4487_{\pm 0.0802}$ & $0.4893_{\pm 0.1146}$ & $0.5076_{\pm 0.0681}$ & $0.4376_{\pm 0.0176}$ & $0.5331_{\pm 0.0436}$ & $0.4662_{\pm 0.0529}$ & $0.3180_{\pm 0.0530}$ \\
LAPO-Slots & $0.2174_{\pm 0.0049}$ & $0.4550_{\pm 0.2032}$ & $0.2691_{\pm 0.0081}$ & $0.2444_{\pm 0.0319}$ & $0.2927_{\pm 0.0616}$ & $0.1666_{\pm 0.0047}$ & $0.2083_{\pm 0.1377}$ & $0.3184_{\pm 0.0070}$ & $0.1600_{\pm 0.0045}$ & $0.1183_{\pm 0.0179}$ \\
\midrule
\textbf{MaskLAM@GT} & $\mathbf{0.1140_{\pm 0.0066}}$ & $0.1441_{\pm 0.0080}$ & $\mathbf{0.1291_{\pm 0.0005}}$ & $\mathbf{0.0738_{\pm 0.0077}}$ & $0.0548_{\pm 0.0021}$ & $\mathbf{0.0156_{\pm 0.0003}}$ & $\mathbf{0.0391_{\pm 0.0018}}$ & $0.2538_{\pm 0.0068}$ & $\mathbf{0.0409_{\pm 0.0008}}$ & $\mathbf{0.0148_{\pm 0.0016}}$ \\
\textbf{MaskLAM@SAM} & $0.1246_{\pm 0.0077}$ & $\mathbf{0.1366_{\pm 0.0061}}$ & $0.1316_{\pm 0.0077}$ & $0.0789_{\pm 0.0068}$ & $\mathbf{0.0524_{\pm 0.0046}}$ & $0.0184_{\pm 0.0015}$ & $0.0462_{\pm 0.0007}$ & $\mathbf{0.2381_{\pm 0.0100}}$ & $0.0605_{\pm 0.0022}$ & $0.0161_{\pm 0.0018}$ \\
\end{tabular}%
}

\caption{\textbf{Per-task NMSE on DMW (distractor).} MaskLAM@GT and MaskLAM@SAM reduce NMSE by $2$--$3\times$ relative to LAPO and surpass every other baseline on all ten tasks, including the contact-rich pick-place, peg-insert-side, and push.}
\label{tab:q1_mse_meta_world_distractor}
\end{table}

\section{Masking Improves Downstream Returns}
\label{sec:appendix_q1_2}

Per-task breakdown of the aggregated downstream return in \Cref{tab:q1_combined_aggregate}; metric definitions in \Cref{sec:appendix_evaluation_details}. All numbers are mean $\pm$ standard deviation across three seeds.

\textbf{Distractor setting.} MaskLAM@GT and MaskLAM@SAM exceed every label-free baseline on every DCS task (\Cref{tab:q1_return_distracting_control_suite}), and match or exceed LAOM-Labels on 7 of 10 DMW tasks (\Cref{tab:q1_return_meta_world_distractor}) despite using no action supervision in Stage~1.

\textbf{Distractor-free setting.} On DCS (\Cref{tab:q1_return_distracting_control_suite}, columns 5--8), LAOM-Labels is the top-performing method, but MaskLAM stays in reach. On DMW (\Cref{tab:q1_return_meta_world_vanilla}), MaskLAM ties LAPO, which is the strongest baseline in the absence of distractors.

\begin{table}[H]
\centering
\renewcommand{\arraystretch}{1.15}
\setlength{\tabcolsep}{4pt}
\resizebox{\textwidth}{!}{%
\begin{tabular}{lcccccccc}
\multicolumn{9}{c}{\textbf{Distracting Control Suite (Mean NR~$\uparrow$)}} \\
\midrule
 & \multicolumn{4}{c}{\textbf{Distractor}} & \multicolumn{4}{c}{\textbf{Vanilla}} \\
\cmidrule(lr){2-5}\cmidrule(lr){6-9}
\textbf{Method} & \textbf{cheetah-run} & \textbf{hopper-hop} & \textbf{humanoid-walk} & \textbf{walker-run} & \textbf{cheetah-run} & \textbf{hopper-hop} & \textbf{humanoid-walk} & \textbf{walker-run} \\
\midrule
LAPO & $0.2057_{\pm 0.0211}$ & $0.0043_{\pm 0.0005}$ & $0.0375_{\pm 0.0115}$ & $0.0562_{\pm 0.0038}$ & $0.6828_{\pm 0.0431}$ & $0.1729_{\pm 0.0180}$ & $0.0022_{\pm 0.0001}$ & $0.0876_{\pm 0.0496}$ \\
LAOM-Labels & $\mathbf{0.9415_{\pm 0.0018}}$ & $\mathbf{0.5425_{\pm 0.0120}}$ & $0.0448_{\pm 0.0030}$ & $\mathbf{0.7364_{\pm 0.0578}}$ & $0.9219_{\pm 0.0091}$ & $\mathbf{0.7114_{\pm 0.0292}}$ & $0.0019_{\pm 0.0001}$ & $\mathbf{0.7981_{\pm 0.0275}}$ \\
LAOF & $0.2721_{\pm 0.0529}$ & $0.0561_{\pm 0.0189}$ & $0.0082_{\pm 0.0022}$ & $0.0391_{\pm 0.0040}$ & $0.2131_{\pm 0.0403}$ & $0.0281_{\pm 0.0155}$ & $0.0022_{\pm 0.0000}$ & $0.0452_{\pm 0.0119}$ \\
LAPO-Slots & $0.1055_{\pm 0.0141}$ & $0.0020_{\pm 0.0007}$ & $0.0247_{\pm 0.0052}$ & $0.0494_{\pm 0.0113}$ & $0.1801_{\pm 0.0146}$ & $0.1211_{\pm 0.0234}$ & $0.0020_{\pm 0.0002}$ & $0.2023_{\pm 0.0437}$ \\
\midrule
\textbf{MaskLAM@GT} & $0.7588_{\pm 0.0309}$ & $0.2736_{\pm 0.0382}$ & $\mathbf{0.0758_{\pm 0.0121}}$ & $0.4022_{\pm 0.0381}$ & $\mathbf{0.9362_{\pm 0.0286}}$ & $0.6337_{\pm 0.0451}$ & $0.0024_{\pm 0.0000}$ & $0.4782_{\pm 0.0407}$ \\
\textbf{MaskLAM@SAM} & $0.7384_{\pm 0.0342}$ & $0.2164_{\pm 0.1050}$ & $0.0645_{\pm 0.0086}$ & $0.4130_{\pm 0.0136}$ & $0.9167_{\pm 0.0151}$ & $0.5628_{\pm 0.0733}$ & $\mathbf{0.0024_{\pm 0.0002}}$ & $0.4919_{\pm 0.0548}$ \\
\end{tabular}%
}

\caption{\textbf{Per-task NR on DCS in distractor (columns 1--4) and distractor-free (columns 5--8) settings.} MaskLAM is the top-performing label-free method on every task under distractors and on 3 of 4 tasks without distractors, where it also lands near LAOM-Labels.}
\label{tab:q1_return_distracting_control_suite}
\end{table}

\begin{table}[H]
\centering
\renewcommand{\arraystretch}{1.15}
\setlength{\tabcolsep}{4pt}
\resizebox{\textwidth}{!}{%
\begin{tabular}{lcccccccccc}
\multicolumn{11}{c}{\textbf{Distracting Meta-World - Vanilla (Mean NSR~$\uparrow$)}} \\
\midrule
\textbf{Method} & \textbf{reach} & \textbf{push} & \textbf{pick-place} & \textbf{door-open} & \textbf{drawer-open} & \textbf{drawer-close} & \textbf{button-press-topdown} & \textbf{peg-insert-side} & \textbf{window-open} & \textbf{window-close} \\
\midrule
LAPO & $0.5648_{\pm 0.1118}$ & $\mathbf{0.9364_{\pm 0.0348}}$ & $0.7194_{\pm 0.0555}$ & $0.9212_{\pm 0.0212}$ & $\mathbf{1.0000_{\pm 0.0000}}$ & $\mathbf{1.0000_{\pm 0.0000}}$ & $\mathbf{1.0000_{\pm 0.0000}}$ & $0.4011_{\pm 0.0187}$ & $0.9668_{\pm 0.0146}$ & $\mathbf{1.0000_{\pm 0.0000}}$ \\
LAOM-Labels & $0.6133_{\pm 0.0633}$ & $0.1983_{\pm 0.0451}$ & $0.0750_{\pm 0.0328}$ & $0.9400_{\pm 0.0350}$ & $\mathbf{1.0000_{\pm 0.0000}}$ & $0.9967_{\pm 0.0058}$ & $0.9900_{\pm 0.0000}$ & $\mathbf{0.4567_{\pm 0.0961}}$ & $\mathbf{1.0000_{\pm 0.0000}}$ & $\mathbf{1.0000_{\pm 0.0000}}$ \\
LAOF & $\mathbf{0.7297_{\pm 0.2191}}$ & $0.4307_{\pm 0.2137}$ & $0.3347_{\pm 0.1592}$ & $\mathbf{0.9690_{\pm 0.0494}}$ & $\mathbf{1.0000_{\pm 0.0000}}$ & $\mathbf{1.0000_{\pm 0.0000}}$ & $0.8317_{\pm 0.2916}$ & $0.4140_{\pm 0.0233}$ & $0.4507_{\pm 0.2192}$ & $\mathbf{1.0000_{\pm 0.0000}}$ \\
LAPO-Slots & $0.5076_{\pm 0.0798}$ & $0.9242_{\pm 0.0284}$ & $0.6704_{\pm 0.0571}$ & $0.9468_{\pm 0.0150}$ & $\mathbf{1.0000_{\pm 0.0000}}$ & $\mathbf{1.0000_{\pm 0.0000}}$ & $\mathbf{1.0000_{\pm 0.0000}}$ & $0.2811_{\pm 0.1385}$ & $0.9963_{\pm 0.0042}$ & $\mathbf{1.0000_{\pm 0.0000}}$ \\
\midrule
\textbf{MaskLAM@GT} & $0.5430_{\pm 0.0188}$ & $0.9171_{\pm 0.0379}$ & $0.7054_{\pm 0.0556}$ & $0.9213_{\pm 0.0389}$ & $\mathbf{1.0000_{\pm 0.0000}}$ & $\mathbf{1.0000_{\pm 0.0000}}$ & $\mathbf{1.0000_{\pm 0.0000}}$ & $0.3102_{\pm 0.0812}$ & $0.9507_{\pm 0.0304}$ & $0.9281_{\pm 0.1012}$ \\
\textbf{MaskLAM@SAM} & $0.5375_{\pm 0.0591}$ & $0.9143_{\pm 0.0100}$ & $\mathbf{0.7220_{\pm 0.0368}}$ & $0.9378_{\pm 0.0161}$ & $\mathbf{1.0000_{\pm 0.0000}}$ & $\mathbf{1.0000_{\pm 0.0000}}$ & $\mathbf{1.0000_{\pm 0.0000}}$ & $0.2369_{\pm 0.0570}$ & $0.9372_{\pm 0.0324}$ & $0.9958_{\pm 0.0039}$ \\
\end{tabular}%
}

\caption{\textbf{Per-task NSR on DMW (vanilla).} MaskLAM ties LAPO, the strongest baseline in the absence of distractors, indicating that loss masking carries no penalty when its target assumption (exogenous contamination) is absent.}
\label{tab:q1_return_meta_world_vanilla}
\end{table}

\begin{table}[H]
\centering
\renewcommand{\arraystretch}{1.15}
\setlength{\tabcolsep}{4pt}
\resizebox{\textwidth}{!}{%
\begin{tabular}{lcccccccccc}
\multicolumn{11}{c}{\textbf{Distracting Meta-World - Distractor (Mean NSR~$\uparrow$)}} \\
\midrule
\textbf{Method} & \textbf{reach} & \textbf{push} & \textbf{pick-place} & \textbf{door-open} & \textbf{drawer-open} & \textbf{drawer-close} & \textbf{button-press-topdown} & \textbf{peg-insert-side} & \textbf{window-open} & \textbf{window-close} \\
\midrule
LAPO & $0.2100_{\pm 0.0211}$ & $0.1715_{\pm 0.0461}$ & $0.0560_{\pm 0.0207}$ & $0.9118_{\pm 0.0247}$ & $0.9979_{\pm 0.0029}$ & $\mathbf{1.0000_{\pm 0.0000}}$ & $0.9934_{\pm 0.0080}$ & $0.0535_{\pm 0.0140}$ & $0.9698_{\pm 0.0093}$ & $0.9981_{\pm 0.0005}$ \\
LAOM-Labels & $0.5250_{\pm 0.0527}$ & $0.5250_{\pm 0.0482}$ & $0.4433_{\pm 0.1079}$ & $\mathbf{0.9483_{\pm 0.0153}}$ & $\mathbf{1.0000_{\pm 0.0000}}$ & $\mathbf{1.0000_{\pm 0.0000}}$ & $\mathbf{1.0000_{\pm 0.0000}}$ & $\mathbf{0.3683_{\pm 0.1040}}$ & $\mathbf{1.0000_{\pm 0.0000}}$ & $\mathbf{1.0000_{\pm 0.0000}}$ \\
LAOF & $0.3927_{\pm 0.0264}$ & $0.3560_{\pm 0.1863}$ & $0.2853_{\pm 0.1513}$ & $0.9340_{\pm 0.0245}$ & $\mathbf{1.0000_{\pm 0.0000}}$ & $0.9200_{\pm 0.1213}$ & $0.5333_{\pm 0.2197}$ & $0.0847_{\pm 0.0133}$ & $0.2617_{\pm 0.0480}$ & $\mathbf{1.0000_{\pm 0.0000}}$ \\
LAPO-Slots & $0.5321_{\pm 0.0531}$ & $0.7736_{\pm 0.1015}$ & $0.7330_{\pm 0.0292}$ & $0.9124_{\pm 0.0058}$ & $0.9965_{\pm 0.0061}$ & $\mathbf{1.0000_{\pm 0.0000}}$ & $0.9945_{\pm 0.0095}$ & $0.3347_{\pm 0.0212}$ & $0.9970_{\pm 0.0029}$ & $\mathbf{1.0000_{\pm 0.0000}}$ \\
\midrule
\textbf{MaskLAM@GT} & $\mathbf{0.5451_{\pm 0.0455}}$ & $\mathbf{0.9286_{\pm 0.0222}}$ & $0.7771_{\pm 0.0511}$ & $0.9076_{\pm 0.0319}$ & $\mathbf{1.0000_{\pm 0.0000}}$ & $\mathbf{1.0000_{\pm 0.0000}}$ & $\mathbf{1.0000_{\pm 0.0000}}$ & $0.2812_{\pm 0.0542}$ & $0.9921_{\pm 0.0062}$ & $\mathbf{1.0000_{\pm 0.0000}}$ \\
\textbf{MaskLAM@SAM} & $0.4861_{\pm 0.0460}$ & $0.9073_{\pm 0.0469}$ & $\mathbf{0.8574_{\pm 0.0293}}$ & $0.9133_{\pm 0.0347}$ & $\mathbf{1.0000_{\pm 0.0000}}$ & $\mathbf{1.0000_{\pm 0.0000}}$ & $\mathbf{1.0000_{\pm 0.0000}}$ & $0.2966_{\pm 0.0696}$ & $0.8578_{\pm 0.0865}$ & $0.9994_{\pm 0.0010}$ \\
\end{tabular}%
}

\caption{\textbf{Per-task NSR on DMW (distractor).} MaskLAM@GT and MaskLAM@SAM match or exceed LAOM-Labels on 7 of 10 tasks while outperforming every other baseline, despite using no action supervision in Stage~1.}
\label{tab:q1_return_meta_world_distractor}
\end{table}

\section{Masking Improves Sample Efficiency}
\label{sec:appendix_q2}

Per-environment breakdown of \Cref{fig:q2_aggregate_suites_normalized}: NR on DCS (distractor) and NSR on DMW (distractor) as a function of the Stage~3 label budget, swept from $2\text{k}$ to $128\text{k}$ labels with the Stage~1 latent action model held fixed. All numbers in this section are reported in the distractor setting.

\begin{figure}[h]
    \centering
    \includegraphics[width=\textwidth]{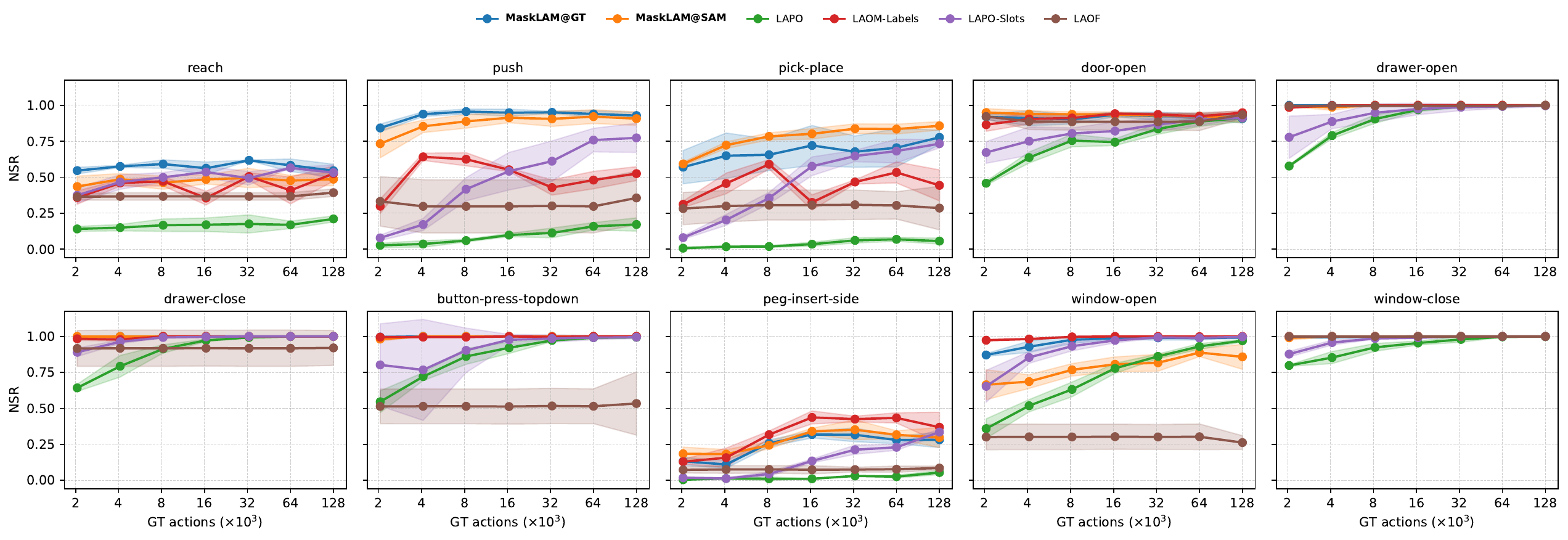}
    \caption{\textbf{Per-task NSR on DMW (distractor) as a function of the Stage~3 label budget (log scale).} MaskLAM reaches peak performance with $2\text{k}$ labels on six of ten tasks; LAPO and LAPO-Slots require nearly two orders of magnitude more labels to close the gap.}
    \label{fig:q2_meta_world_environments_normalized}
\end{figure}

\textbf{DMW.} MaskLAM@GT reaches near peak NSR with $2\text{k}$ labels on six of ten tasks (reach, push, pick-place, drawer-open, drawer-close, window-open), while LAPO and LAPO-Slots need nearly two orders of magnitude more labels to close the gap (\Cref{fig:q2_meta_world_environments_normalized}).

\begin{figure}[h]
    \centering
    \includegraphics[width=\textwidth]{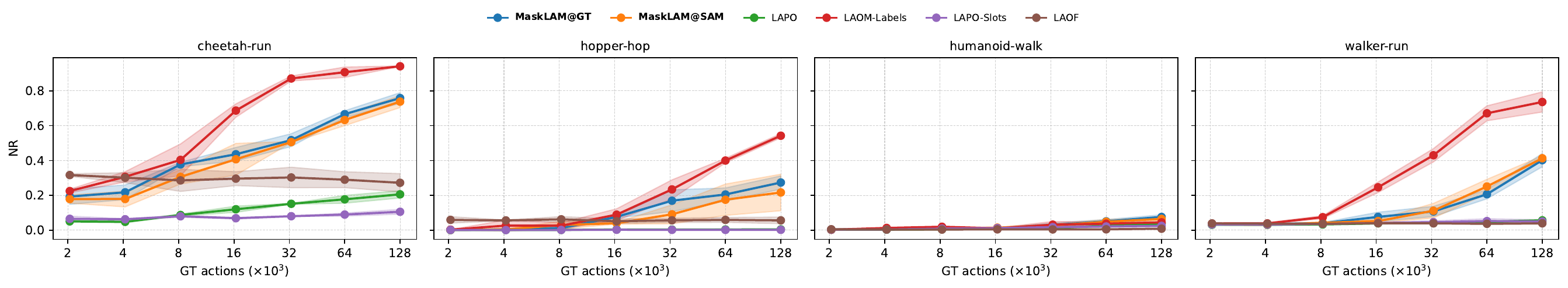}
    \caption{\textbf{Per-task NR on DCS (distractor) as a function of the Stage~3 label budget (log scale).} MaskLAM@GT saturates earlier than every label-free baseline; LAPO and LAPO-Slots fail to recover useful policies at any budget.}
    \label{fig:q2_dm_control_environments_normalized}
\end{figure}

\textbf{DCS.} LAPO and LAPO-Slots fail to recover useful policies at any budget we tested; MaskLAM@GT reaches competitive returns from $16\text{k}$--$32\text{k}$ labels (\Cref{fig:q2_dm_control_environments_normalized}). LAOM-Labels remains the ceiling on hopper-hop and walker-run, where well-aligned latents still require more decoder capacity to translate into stable gaits.

\textbf{Mask source.} MaskLAM@SAM tracks MaskLAM@GT across all environments and label budgets, so swapping ground-truth for SAM-predicted masks costs no sample efficiency.

\section{MaskLAM is Robust to Occlusions}
\label{sec:appendix_q5}

Real videos rarely keep the full agent in frame: limbs leave the camera view, foreground objects pass in front of the agent, and contact-rich motion regularly occludes joints behind other parts of the body. A practical latent action model has to keep producing well-aligned latents under this regime. We probe this on the DCS (distractor) by occluding a controlled fraction of the agent at training and evaluation time, and reporting the action-probe NMSE of \eqref{eq:nmse} per task.

\begin{figure}[H]
    \centering
    \includegraphics[width=\textwidth]{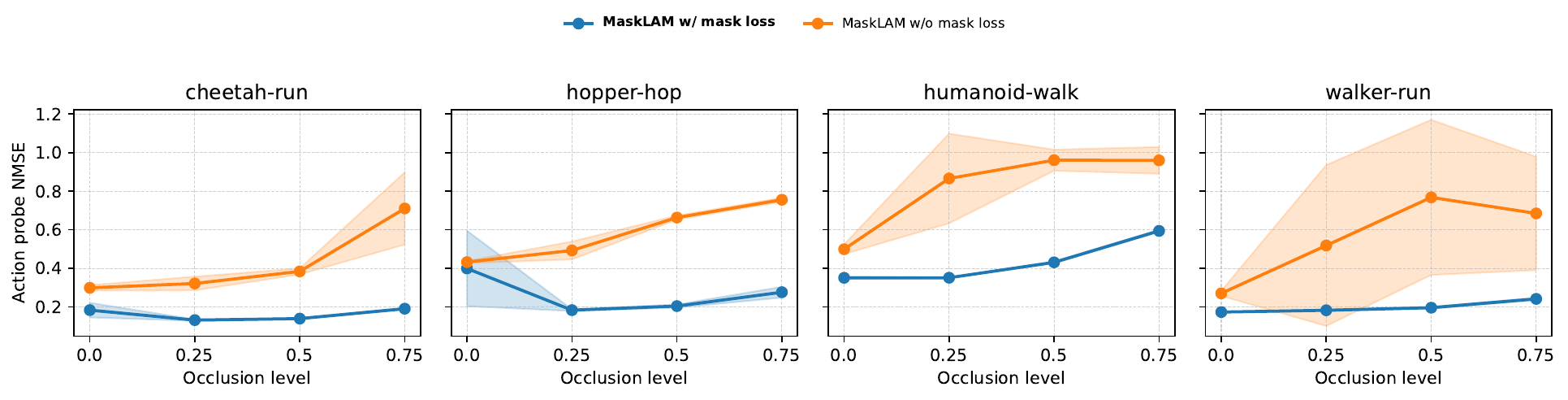}
    \caption{\textbf{NMSE as a function of agent occlusion fraction on the DCS ($d_z = 128$).} The $x$-axis is the fraction $\rho$ of the agent's bounding-box area occluded by a random gray rectangle at training time (0\% = no occlusion, 75\% = three-quarters of agent pixels removed). Blue lines: MaskLAM with masked loss; orange lines: same architecture with the masked loss removed. With the masked loss, NMSE stays nearly flat and tightly concentrated across the full occlusion range on all four tasks; without it, error more than doubles and variance increases sharply, because the growing occlusion patch reduces the action-driven fraction of the unmasked reconstruction target while distractor variance remains unchanged.}
    \label{fig:q5_per_env_normalized}
\end{figure}

\textbf{Occlusion procedure.} For every video sample we synthesize a single random axis-aligned occlusion rectangle anchored to the agent. The construction is as follows:
\begin{enumerate}
    \item Take the union of the ground-truth foreground mask over the frames of the video sample, $U = \bigvee_{t} M_t$, and compute its tight axis-aligned bounding box $B$ of area $A_B$.
    \item Sample a target occlusion area $A = \rho \, A_B$, where $\rho \in [0, 1]$ is the occlusion fraction reported on the x-axis of \Cref{fig:q5_per_env_normalized}.
    \item Sample a log-uniform aspect ratio $r \sim \exp(\mathcal{U}(-\tfrac{1}{2}, \tfrac{1}{2}))$ and derive height $h = \sqrt{A r}$ and width $w = A / h$, both clamped to $B$ so the rectangle never leaves the agent footprint.
    \item Sample a uniformly random top-left corner inside the remaining slack of $B$ to obtain the final rectangle $R$.
    \item Apply $R$ identically to every frame of the video sample: overwrite the observation pixels inside $R$ with a constant gray fill (corresponding to a $0.0$ fill value after normalization) and zero the corresponding entries of the segmentation mask.
\end{enumerate}

The rectangle is resampled per trajectory and per training step, so the model never memorizes a fixed occlusion. Zeroing the mask inside $R$ mirrors the behavior of any off-the-shelf segmenter, which simply fails to label pixels it cannot see; the IDM and FDM therefore receive exactly the inputs and supervisory signal they would receive in a deployment where occluded body parts disappear from the predicted segmentation.

\textbf{Action-probe NMSE under occlusion.} \Cref{fig:q5_per_env_normalized} sweeps $\rho$ from $0\%$ to $75\%$ for both MaskLAM and an ablation that removes the masked loss. With the masked loss the curve stays nearly flat across the full range on every environment, while removing it more than doubles the probe error on cheetah-run and walker-run and degrades the other two environments by a similar margin. The mask-loss curve is also visibly tighter across seeds.

\begin{figure}[h]
    \centering
    \setlength{\tabcolsep}{1pt}
    \begin{tabular*}{\linewidth}{@{\extracolsep{\fill}}lc@{}}
        \makebox[0.04\linewidth][c]{\rotatebox{90}{\scriptsize \shortstack[c]{\hspace*{2em}Occlusion 0.00}}} &
        \includegraphics[width=0.94\linewidth]{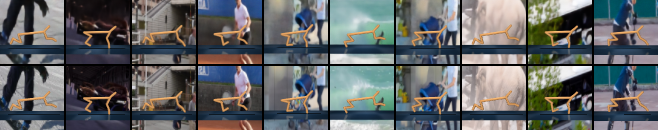} \\
        \makebox[0.04\linewidth][c]{\rotatebox{90}{\scriptsize \shortstack[c]{\hspace*{2em}Occlusion 0.25}}} &
        \includegraphics[width=0.94\linewidth]{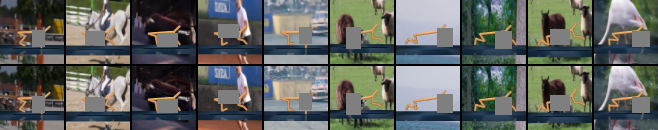} \\
        \makebox[0.04\linewidth][c]{\rotatebox{90}{\scriptsize \shortstack[c]{\hspace*{2em}Occlusion 0.50}}} &
        \includegraphics[width=0.94\linewidth]{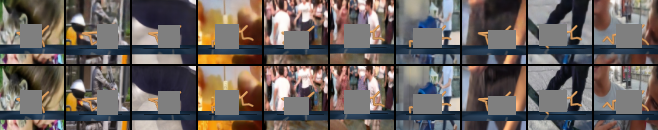} \\
        \makebox[0.04\linewidth][c]{\rotatebox{90}{\scriptsize \shortstack[c]{\hspace*{2em}Occlusion 0.75}}} &
        \includegraphics[width=0.94\linewidth]{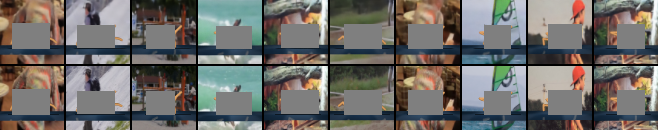} \\
    \end{tabular*}
    \caption{\textbf{FDM reconstructions of MaskLAM \emph{without} the masked loss objective on cheetah-run (DCS, distractor) under increasing agent occlusion.} Rows correspond to occlusion fractions $0\%$, $25\%$, $50\%$, and $75\%$ from top to bottom. Within each row, the top sub-row shows the predicted next observation $\hat{o}_{t+1}$ and the bottom sub-row shows the ground-truth $o_{t+1}$ with the gray occlusion patch overlaid on the cheetah. Without the masked loss, the world model is forced to reconstruct occlusion pixels.}
    \label{fig:occlusion_slapo_no_mask_loss}
\end{figure}

\begin{figure}[h]
    \centering
    \setlength{\tabcolsep}{1pt}
    \begin{tabular*}{\linewidth}{@{\extracolsep{\fill}}lc@{}}
        \makebox[0.04\linewidth][c]{\rotatebox{90}{\scriptsize \shortstack[c]{\hspace*{2em}Occlusion 0.00}}} &
        \includegraphics[width=0.94\linewidth]{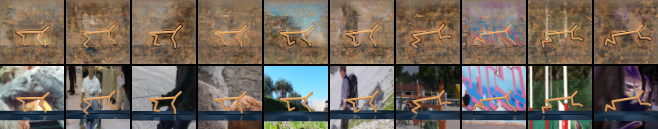} \\
        \makebox[0.04\linewidth][c]{\rotatebox{90}{\scriptsize \shortstack[c]{\hspace*{2em}Occlusion 0.25}}} &
        \includegraphics[width=0.94\linewidth]{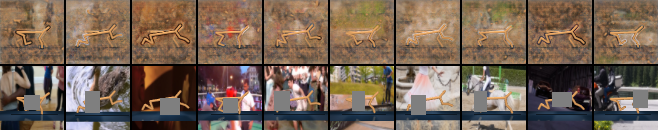} \\
        \makebox[0.04\linewidth][c]{\rotatebox{90}{\scriptsize \shortstack[c]{\hspace*{2em}Occlusion 0.50}}} &
        \includegraphics[width=0.94\linewidth]{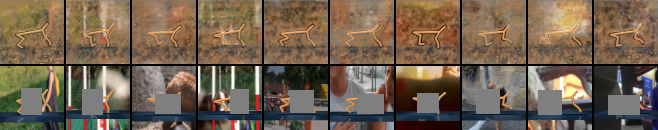} \\
        \makebox[0.04\linewidth][c]{\rotatebox{90}{\scriptsize \shortstack[c]{\hspace*{2em}Occlusion 0.75}}} &
        \includegraphics[width=0.94\linewidth]{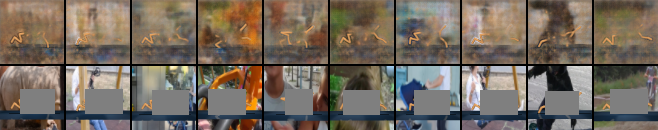} \\
    \end{tabular*}
    \caption{\textbf{FDM reconstructions of MaskLAM \emph{with} the masked loss objective on cheetah-run (DCS, distractor) under increasing agent occlusion.} Layout matches \Cref{fig:occlusion_slapo_no_mask_loss}: rows correspond to occlusion fractions $0\%$, $25\%$, $50\%$, and $75\%$, and within each row the top sub-row is the predicted next observation $\hat{o}_{t+1}$ while the bottom sub-row is the ground-truth $o_{t+1}$ with the gray occlusion patch on the cheetah. Restricting reconstruction to the supervisory mask keeps the prediction inside the mask robust to occlusion, recovering the visible cheetah pixels cleanly up to $75\%$ occlusion and mirroring the action-probe NMSE trend in \Cref{fig:q5_per_env_normalized}.}
    \label{fig:occlusion_slapo_mask_loss}
\end{figure}

\textbf{Why masking helps under occlusion.} Occluding the agent does not add new exogenous variance, but it shrinks the visible agent footprint and therefore the action-driven component of the FDM target, while distractor variance behind and around the agent stays the same. Under the unmasked loss the action-to-noise ratio of the supervisory target degrades as $\rho$ grows, and the optimization increasingly pressures $z_t$ to encode the distractor variance that now dominates the residual. This is the linearized PCA picture of \citet{zhang_what_2025} applied to a target whose endogenous content has been thinned: as the agent signal in the target shrinks, the principal direction of frame-to-frame change tilts toward the distractor, and so does $z_t$. Under the masked loss the occluded rectangle is removed from $M_{t+1}$ before the residual is computed, so the target contains only visible agent pixels at every $\rho$. The action-to-noise ratio of the supervisory signal stays effectively at one regardless of how much of the agent is hidden, and $z_t$ keeps encoding only action-driven variance. \Cref{fig:occlusion_slapo_no_mask_loss,fig:occlusion_slapo_mask_loss} show this at the pixel level. Without the masked loss the FDM continues to reproduce the full scene, occlusion patch, cheetah, and distractor video, pixel-accurately at every $\rho$, so $z_t$ keeps carrying all of that variance even as the action-relevant share of the target shrinks. With the masked loss the FDM is asked only to reconstruct visible agent pixels and does so cleanly up to $75\%$ occlusion, while everything outside $M_{t+1}$ collapses to the same low-frequency texture seen in \Cref{fig:wm_slapo_dcs}. Because zeroing the mask inside $R$ is exactly what an off-the-shelf segmenter does on partially visible agents, this experiment also demonstrates MaskLAM's robustness to real-world segmentation behavior.

\section{Masking Enables Compact Latent Action Spaces}
\label{sec:appendix_q3}

Per-environment breakdown of \crefpanel{fig:q3_q5_aggregate_normalized}{a}: NMSE on DCS (distractor) as a function of latent action dimension $d_z \in \{32, 64, 128, 256, 512, 1024\}$, with and without the masked loss (all other hyperparameters held fixed).

\begin{figure}[H]
    \centering
    \includegraphics[width=\textwidth]{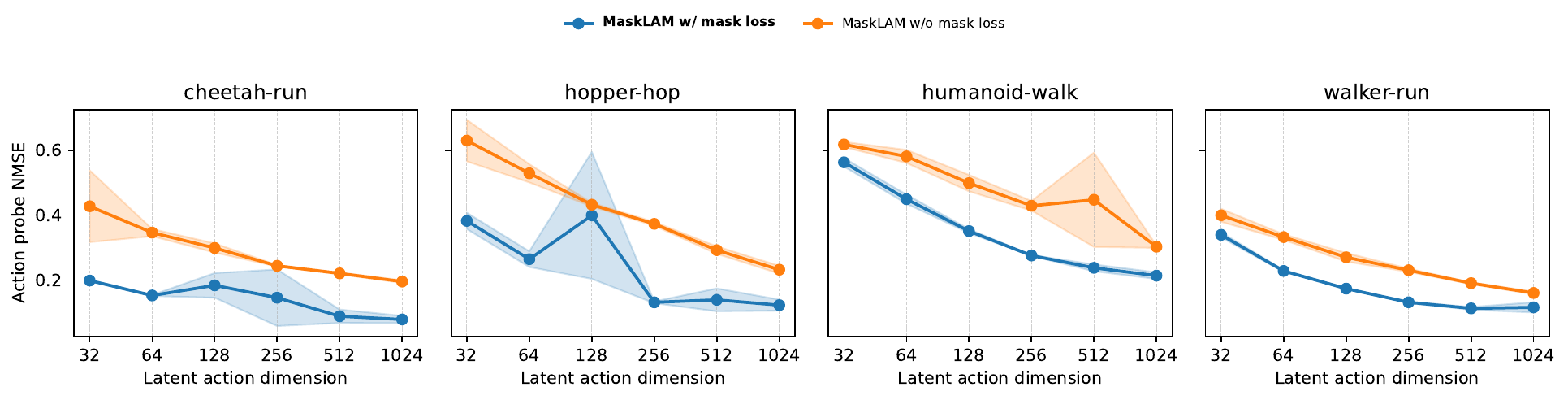}
    \caption{\textbf{Per-task NMSE on DCS (distractor) as a function of latent action dimension (log scale).} Blue: MaskLAM@GT; orange: same architecture with the masked loss removed. The masked loss yields lower NMSE at every $d_z$. A $256$-dim masked model matches a $1024$-dim unmasked one on all four tasks.}
    \label{fig:q3_per_env_normalized}
\end{figure}

\textbf{Compression at fixed alignment.} A $256$-dim masked latent matches or exceeds a $1024$-dim unmasked latent on all four tasks (\Cref{fig:q3_per_env_normalized}), a $4\times$ reduction with no loss in alignment. The unmasked loss must allocate dimensions to both agent and distractor dynamics; the masked loss restricts the supervisory variance to agent pixels, shrinking the effective rank of the prediction target.

\textbf{Low-dimensional regime.} The gap widens at $d_z \leq 64$, where encoding exogenous content under the unmasked loss starves the action signal entirely.

\section{Influence of Mask IoU on Latent Action Quality}
\label{sec:appendix_q6}

In practice, supervisory masks come from an off-the-shelf segmentation model whose boundaries are never pixel-perfect. To quantify how mask quality affects latent action alignment, we perturb ground-truth masks via morphological dilation (expansion) and erosion (shrinkage) with structuring element radii of 0 to 3 pixels. This drops the mask mIoU from 1.0 down to approximately 0.4--0.6 depending on the task and perturbation direction. We retrain MaskLAM@GT from scratch with each perturbed mask set and report the resulting NMSE on the DMW benchmark. We also report baselines as dashed horizontal reference lines to contextualize the absolute error level.

We do not separately perturb MaskLAM@SAM because SAM~2.1 masks already serve as the ``imperfect mask'' condition: the comparison between MaskLAM@GT (radius 0) and MaskLAM@SAM in \Cref{tab:q1_combined_aggregate} shows they are within noise, establishing that real segmentation errors from a foundation model are already tolerable (that too on datasets that do not exactly represent the kind of real-world data those models were trained on). The morphological perturbation experiment isolates a controlled, interpretable axis of degradation beyond what SAM produces.

\textbf{Dilation vs.\ erosion.} Mask expansion (dilation) includes a growing border of background pixels in the supervisory target. This introduces a small amount of exogenous signal into the loss, but the impact on NMSE is mild. Mask shrinkage (erosion) removes agent boundary pixels, reducing the informative area without adding exogenous noise; here the effect is similarly benign. Across all 10 DMW tasks and both perturbation directions, MaskLAM@GT remains below all label-free baselines at every radius tested, confirming that pixel-perfect segmentation is not required. Mask extraction methods with systematic boundary bias (e.g., coarser bounding-box prompts that over-segment by 1--2 pixels) are therefore safe to use with MaskLAM.

\begin{figure}[h]
    \centering
    \includegraphics[width=\textwidth]{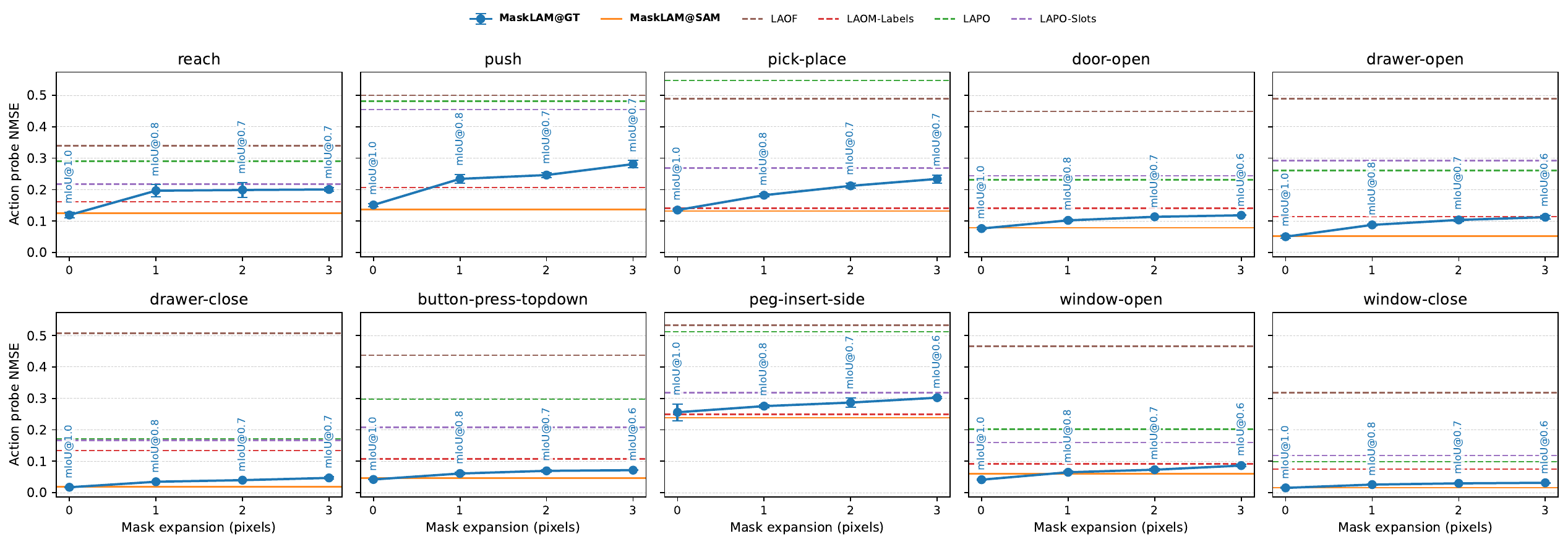}
    \caption{\textbf{NMSE as a function of morphological dilation radius (0--3 pixels) applied to ground-truth masks before MaskLAM@GT training on the DMW.} Each subplot is one task; the $x$-axis is the dilation radius in pixels (0 = unperturbed ground-truth mask); the $y$-axis is NMSE. Dashed horizontal lines show baselines. MaskLAM@GT remains below all label-free baselines across the entire perturbation range, indicating that including a thin border of background pixels in the loss does not meaningfully degrade latent action quality.}
    \label{fig:q6_dilation_environments_normalized}
\end{figure}

\begin{figure}[h]
    \centering
    \includegraphics[width=\textwidth]{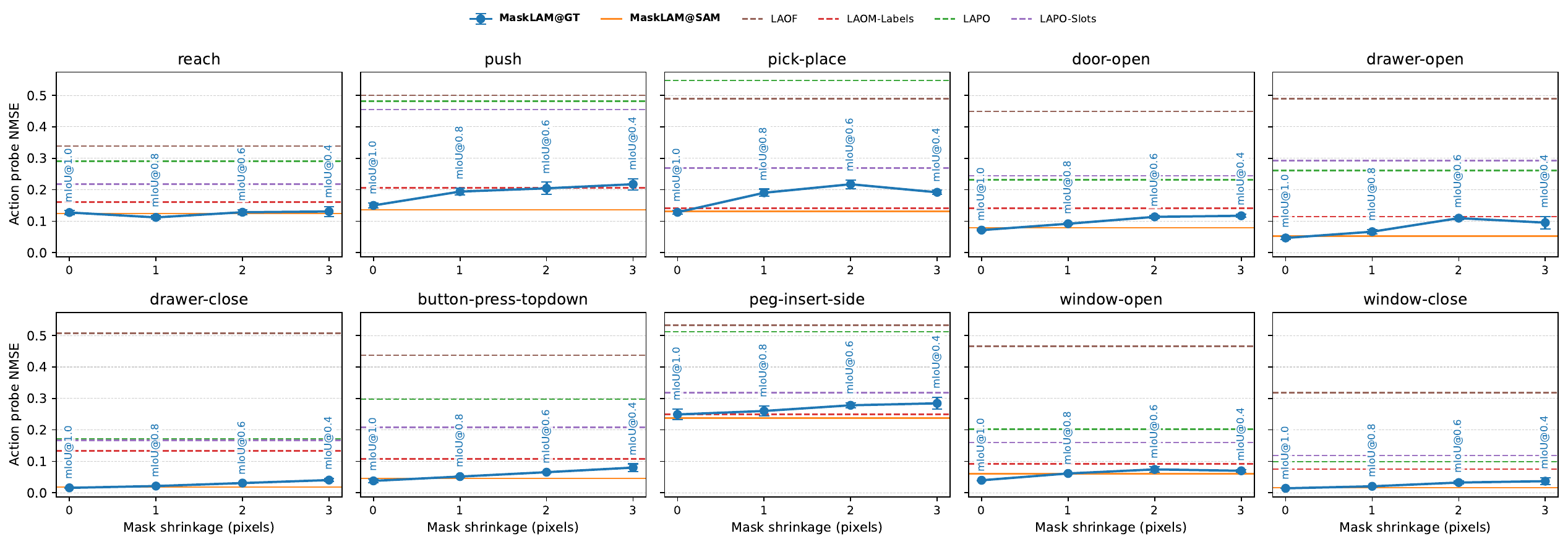}
    \caption{\textbf{NMSE as a function of morphological erosion radius (0--3 pixels) applied to ground-truth masks before MaskLAM@GT training on the DMW.} Layout matches \Cref{fig:q6_dilation_environments_normalized}. Erosion removes agent boundary pixels, reducing the supervisory area without introducing exogenous signal. MaskLAM@GT degrades gracefully and stays below all label-free baselines even at radius~3, where mIoU drops below 0.5 on several tasks.}
    \label{fig:q6_erosion_environments_normalized}
\end{figure}

\begin{figure}[H]
    \centering
    \setlength{\tabcolsep}{1pt}
    \begin{tabular*}{\linewidth}{@{\extracolsep{\fill}}lccccc@{}}
        \makebox[0.035\linewidth][c]{\rotatebox{90}{\scriptsize \hspace*{3em}Dilation}} &
        \includegraphics[width=0.18\linewidth]{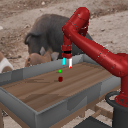} &
        \includegraphics[width=0.18\linewidth]{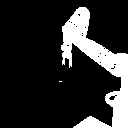} &
        \includegraphics[width=0.18\linewidth]{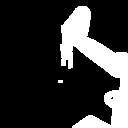} &
        \includegraphics[width=0.18\linewidth]{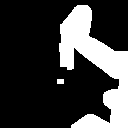} &
        \includegraphics[width=0.18\linewidth]{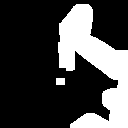} \\
        &
        \makebox[0.18\linewidth][c]{\scriptsize Observation} &
        \makebox[0.18\linewidth][c]{\scriptsize Mask expansion 0 pixel} &
        \makebox[0.18\linewidth][c]{\scriptsize Mask expansion 1 pixel} &
        \makebox[0.18\linewidth][c]{\scriptsize Mask expansion 2 pixel} &
        \makebox[0.18\linewidth][c]{\scriptsize Mask expansion 3 pixel} \\[4pt]
        \makebox[0.035\linewidth][c]{\rotatebox{90}{\scriptsize \hspace*{3em}Erosion}} &
        \includegraphics[width=0.18\linewidth]{figures/q6/q6_dmw_original.png} &
        \includegraphics[width=0.18\linewidth]{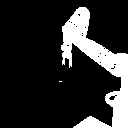} &
        \includegraphics[width=0.18\linewidth]{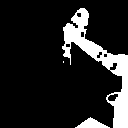} &
        \includegraphics[width=0.18\linewidth]{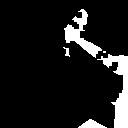} &
        \includegraphics[width=0.18\linewidth]{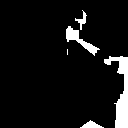} \\
        &
        \makebox[0.18\linewidth][c]{\scriptsize Observation} &
        \makebox[0.18\linewidth][c]{\scriptsize Mask shrinkage 0 pixel} &
        \makebox[0.18\linewidth][c]{\scriptsize Mask shrinkage 1 pixel} &
        \makebox[0.18\linewidth][c]{\scriptsize Mask shrinkage 2 pixel} &
        \makebox[0.18\linewidth][c]{\scriptsize Mask shrinkage 3 pixel}
    \end{tabular*}
    \caption{\textbf{Qualitative visualization of morphological mask perturbations on a representative DMW frame (window-open task, distractor setting).} Top row: progressive dilation of the ground-truth mask by 0--3 pixels, which expands the white supervisory region into the background. Bottom row: progressive erosion by 0--3 pixels, which shrinks the supervisory region inward from the boundary. Even at radius~3 the mask retains the coarse agent silhouette, explaining the graceful degradation observed in \Cref{fig:q6_dilation_environments_normalized,fig:q6_erosion_environments_normalized}.}
    \label{fig:q6_dmw_morphology_examples}
\end{figure}

\section{Extended Ablation Results}
\label{sec:appendix_ablation}

\begin{figure}[H]
    \centering
    \includegraphics[width=\textwidth]{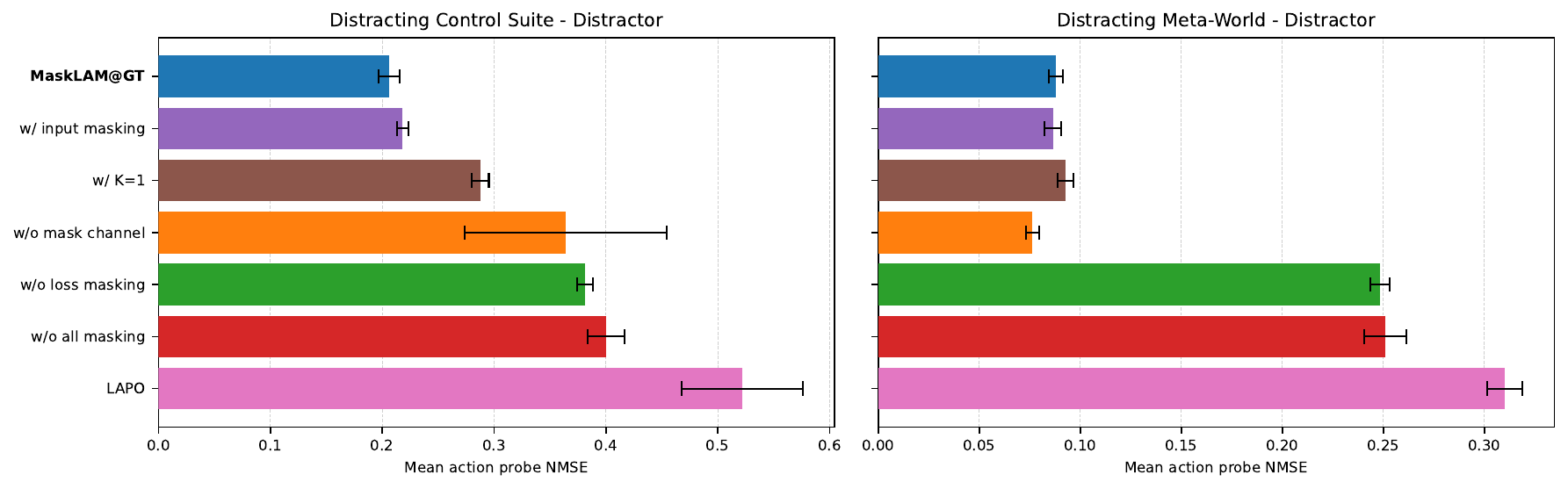}
    \caption{\textbf{Ablation of MaskLAM components aggregated per benchmark suite (DCS left, DMW right).} Loss masking is the principal contributor on both suites; the mask channel helps on DCS but slightly hurts on DMW; multi-step IDM adds a smaller orthogonal gain.}
    \label{fig:ablation_aggregate_suites_normalized}
\end{figure}

\begin{figure}[H]
    \centering
    \includegraphics[width=\textwidth]{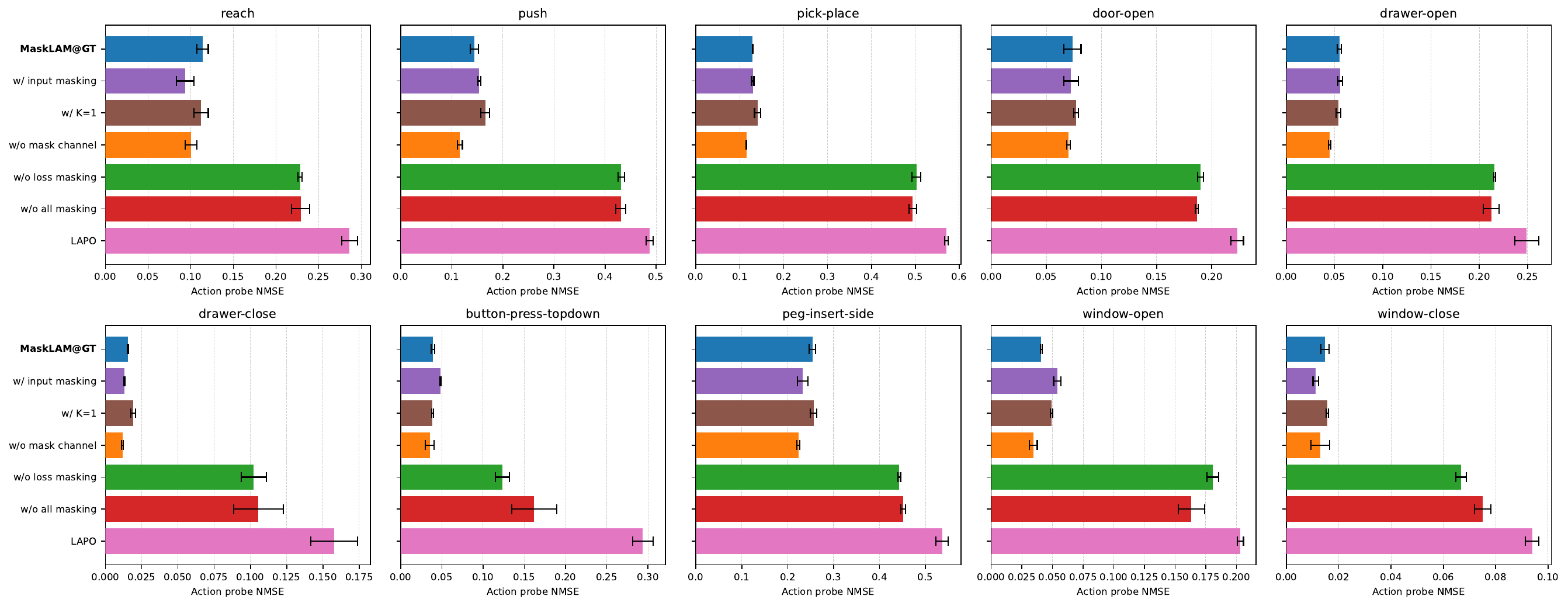}
    \caption{\textbf{Per-task ablation on DMW (distractor).} Loss masking is the principal contributor across all ten tasks; the mask channel slightly hurts performance, as the compact arm silhouettes already provide a sufficient spatial prior; multi-step IDM adds a smaller orthogonal gain.}
    \label{fig:ablation_meta_world_environments_normalized}
\end{figure}

\begin{figure}[H]
    \centering
    \includegraphics[width=\textwidth]{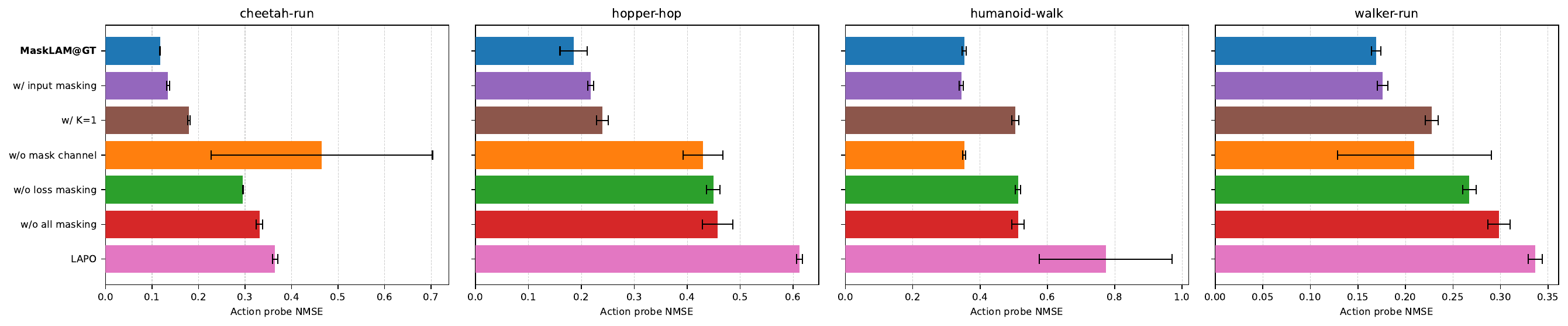}
    \caption{\textbf{Per-task ablation on DCS (distractor).} The hierarchy is consistent across cheetah-run, hopper-hop, humanoid-walk, and walker-run: loss masking is the principal contributor, followed by the mask channel, followed by multi-step IDM. The mask channel's contribution is largest on cheetah-run and hopper-hop, where the thin limb geometry makes an explicit spatial prior most useful.}
    \label{fig:ablation_dm_control_environments_normalized}
\end{figure}

Per-environment breakdown of the aggregated ablation in \Cref{fig:ablation_aggregate_dcs_dmw_combined_normalized}; variant definitions in \Cref{sec:ablation}. The findings of \Cref{sec:ablation} hold at the per-task level on every DCS task (\Cref{fig:ablation_dm_control_environments_normalized}) and every DMW task (\Cref{fig:ablation_meta_world_environments_normalized}), as well as when aggregated separately by benchmark suite (\Cref{fig:ablation_aggregate_suites_normalized}).

\textbf{Loss masking is the principal contributor on every task.} Removing the masked loss more than doubles NMSE on nearly all DCS and DMW tasks.

\textbf{The mask channel helps on DCS but slightly hurts on DMW.} With loss masking active, removing the mask channel increases NMSE on every DCS task, with the largest effect on cheetah-run and hopper-hop. We attribute the per-task variation to the thin limb geometry of these agents, where an explicit spatial prior helps the model isolate the agent from the dynamic background. On DMW, the channel slightly hurts performance instead, as the larger and more compact arm silhouettes already provide a sufficient prior. Without loss masking, the channel makes no meaningful difference on any task: it is useful only when the loss already restricts gradients to the masked region.

\textbf{Multi-step IDM contributes a roughly additive gain.} Setting $k{=}1$ degrades NMSE by a similar margin whether or not masking is present, consistent with the decorrelation argument of \citet{lamb_guaranteed_2022} and confirming that multi-step prediction complements rather than substitutes for the spatial intervention.

\textbf{Input masking matches loss masking on every task.} \textit{w/ input masking} is within noise of MaskLAM@GT on every DCS and DMW task, mirroring the aggregate result in \Cref{sec:ablation}. Both benchmarks feature self-driven agents whose motion is generated entirely within $M_t$, so the interaction surface outside the mask carries no action-relevant signal beyond the agent pixels. Loss masking still preserves the IDM's access to the full scene, and \Cref{fig:eigencam_cheetah_video_clip} confirms it attends to ground-contact pixels outside $M_{t+1}$ in practice. In real video where the agent's motion depends on external contact (a rider on a scooter, a commuter on an escalator), that surface carries action-relevant signal that input masking would discard at the input.

\section{Extended Eigen-CAM Results}
\label{sec:appendix_eigencam}

Eigen-CAM~\citep{muhammad_eigen-cam_2020} produces a saliency map by taking the principal component of the activations of a chosen convolutional layer, highlighting the spatial regions that drive the network's prediction without requiring class labels or gradient signals. We apply it to the final convolutional block of the IDM and overlay the resulting heatmap on the input frame; warmer colors mark pixels that contribute most to the predicted latent action $z_t$. Since the IDM is the only module that consumes observations to produce $z_t$, its saliency directly exposes which pixels shape the latent action.

\begin{figure}[H]
    \centering
    \setlength{\tabcolsep}{1pt}
    \begin{tabular*}{\linewidth}{@{\extracolsep{\fill}}lcccccccccc@{}}
        \makebox[0.025\linewidth][c]{\rotatebox{90}{\scriptsize \hspace*{0.8em}Original}} &
        \includegraphics[width=0.09\linewidth]{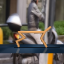} &
        \includegraphics[width=0.09\linewidth]{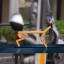} &
        \includegraphics[width=0.09\linewidth]{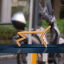} &
        \includegraphics[width=0.09\linewidth]{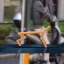} &
        \includegraphics[width=0.09\linewidth]{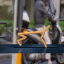} &
        \includegraphics[width=0.09\linewidth]{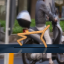} &
        \includegraphics[width=0.09\linewidth]{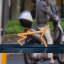} &
        \includegraphics[width=0.09\linewidth]{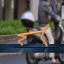} &
        \includegraphics[width=0.09\linewidth]{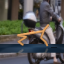} &
        \includegraphics[width=0.09\linewidth]{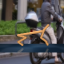} \\
        \makebox[0.025\linewidth][c]{\rotatebox{90}{\scriptsize \hspace*{1.5em}LAPO}} &
        \includegraphics[width=0.09\linewidth]{figures/eigencam_analysis/lapo_eigencam_0.png} &
        \includegraphics[width=0.09\linewidth]{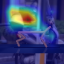} &
        \includegraphics[width=0.09\linewidth]{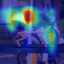} &
        \includegraphics[width=0.09\linewidth]{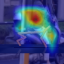} &
        \includegraphics[width=0.09\linewidth]{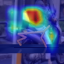} &
        \includegraphics[width=0.09\linewidth]{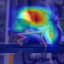} &
        \includegraphics[width=0.09\linewidth]{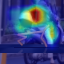} &
        \includegraphics[width=0.09\linewidth]{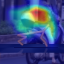} &
        \includegraphics[width=0.09\linewidth]{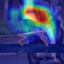} &
        \includegraphics[width=0.09\linewidth]{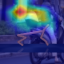} \\
        \makebox[0.025\linewidth][c]{\rotatebox{90}{\scriptsize \hspace*{0.25em}MaskLAM}} &
        \includegraphics[width=0.09\linewidth]{figures/eigencam_analysis/slapo_eigencam_0.png} &
        \includegraphics[width=0.09\linewidth]{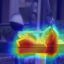} &
        \includegraphics[width=0.09\linewidth]{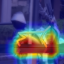} &
        \includegraphics[width=0.09\linewidth]{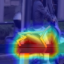} &
        \includegraphics[width=0.09\linewidth]{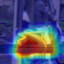} &
        \includegraphics[width=0.09\linewidth]{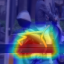} &
        \includegraphics[width=0.09\linewidth]{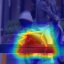} &
        \includegraphics[width=0.09\linewidth]{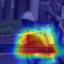} &
        \includegraphics[width=0.09\linewidth]{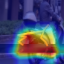} &
        \includegraphics[width=0.09\linewidth]{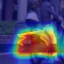}
    \end{tabular*}
    \caption{\textbf{Eigen-CAM saliency on ten consecutive frames of a cheetah-run rollout on DCS (distractor).} Warmer colors indicate higher influence on the predicted latent action. Top: Original frames. Middle: LAPO drifts across the distractor background. Bottom: MaskLAM stays on the cheetah and the ground-plane region under its feet, despite the supervisory mask covering only the cheetah.}
    \label{fig:eigencam_cheetah_video_clip}
\end{figure}

\Cref{fig:eigencam_cheetah_video_clip} extends the single-frame comparison of \Cref{fig:eigencam_cheetah_frame_0} to ten consecutive frames of a cheetah-run rollout in the distracting setting. LAPO's saliency drifts across the distractor background throughout the trajectory and rarely overlaps with the agent, confirming that without the masked loss the IDM grounds $z_t$ in exogenous variation. MaskLAM's saliency stays locked onto the cheetah body and onto the ground-plane region directly under its feet across all ten frames, even though the supervisory mask covers only the cheetah itself (see \Cref{fig:mask_generation_examples}). The ground-plane response is informative: it is endogenous context, the surface the agent interacts with to produce its motion, and the IDM has learned to attend to it without ever being told to.

This is the behavior we designed for in \Cref{sec:method}. The endogenous context varies from video to video and from frame to frame: the floor under a cheetah, a cup approached by a gripper, a door touched by a hand, a tool brought into contact with a workpiece. Specifying these regions a priori for every observation would require dense, task-aware annotation that defeats the purpose of label-free latent action learning, and any conservative mask drawn at input time would either erase useful interaction surfaces or admit distractors back in. Loss masking is therefore the only viable option: the IDM still sees the full observation and is free to attend wherever reconstruction demands, while $z_t$ is shaped only by reconstruction errors inside $M_{t+1}$. The boundary of endogenous context emerges from data rather than from a hand-crafted mask.

\section{Extended UMAP Results}
\label{sec:appendix_umap}

\begin{figure}[H]
    \centering
    \setlength{\tabcolsep}{1pt}
    \begin{tabular*}{\linewidth}{@{\extracolsep{\fill}}lcccccc@{}}
        \makebox[0.02\linewidth][c]{\rotatebox{90}{\scriptsize \hspace*{3em}LAPO}} &
        \includegraphics[width=0.14\linewidth]{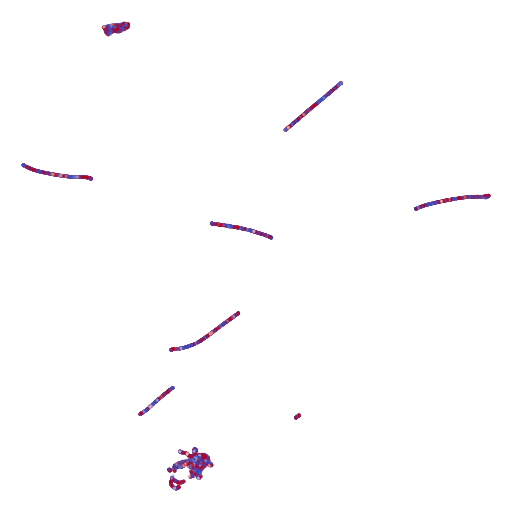} &
        \includegraphics[width=0.14\linewidth]{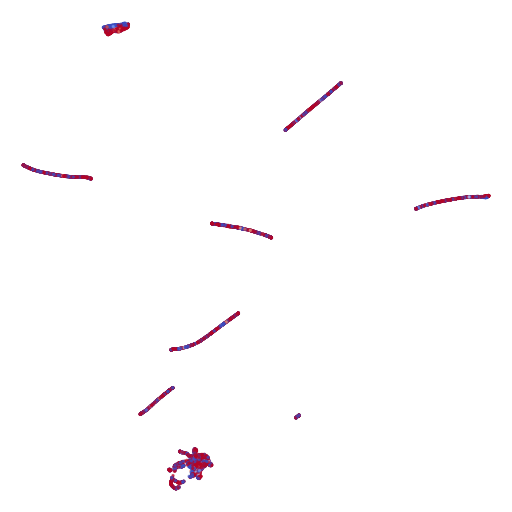} &
        \includegraphics[width=0.14\linewidth]{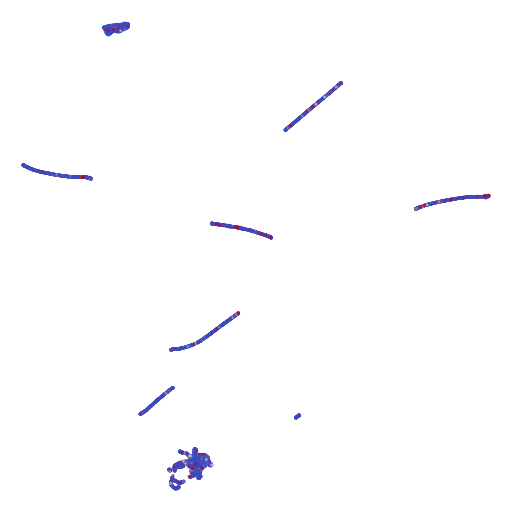} &
        \includegraphics[width=0.14\linewidth]{figures/umap_analysis/umap_lapo_4.png} &
        \includegraphics[width=0.14\linewidth]{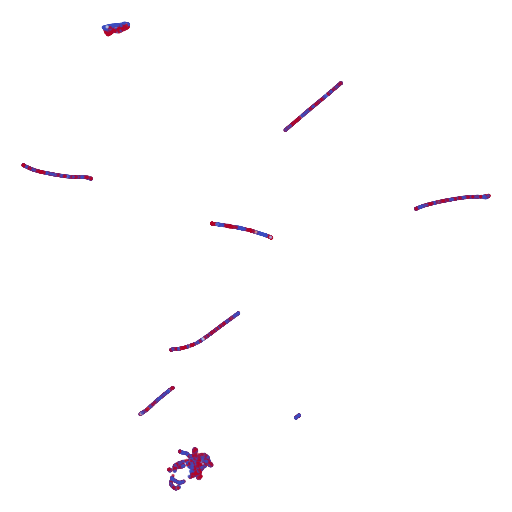} &
        \includegraphics[width=0.14\linewidth]{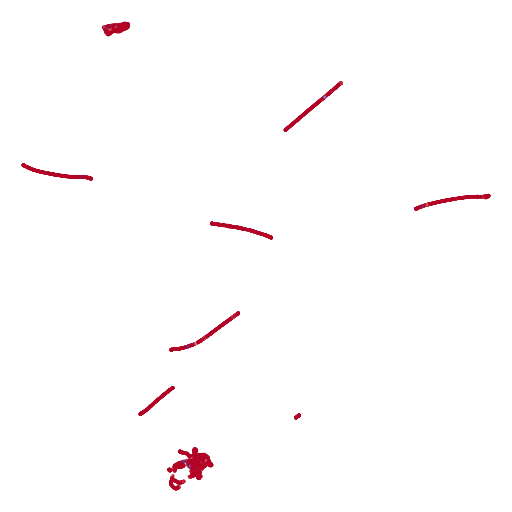} \\
        \makebox[0.02\linewidth][c]{\rotatebox{90}{\scriptsize \hspace*{1em}LAOM-Labels}} &
        \includegraphics[width=0.14\linewidth]{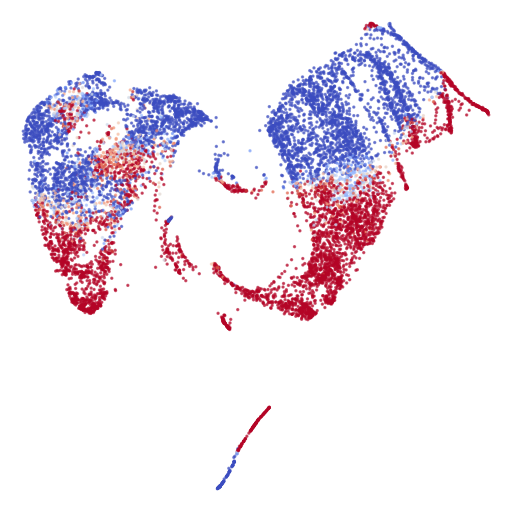} &
        \includegraphics[width=0.14\linewidth]{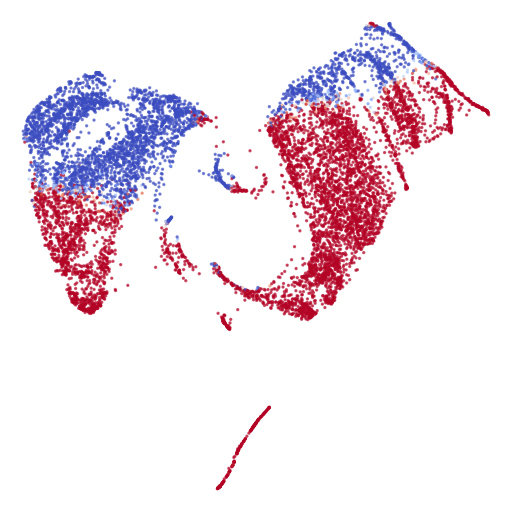} &
        \includegraphics[width=0.14\linewidth]{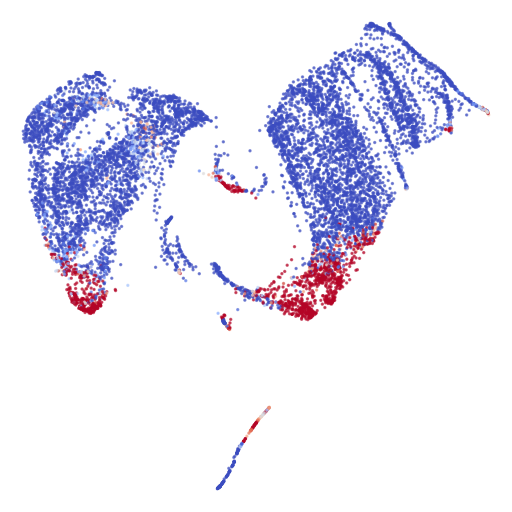} &
        \includegraphics[width=0.14\linewidth]{figures/umap_analysis/umap_laom_labels_4.png} &
        \includegraphics[width=0.14\linewidth]{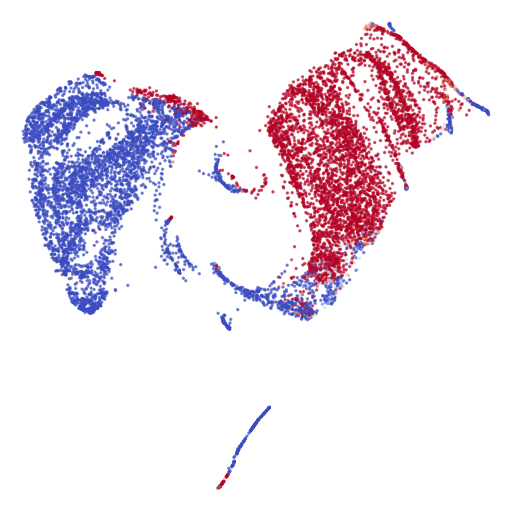} &
        \includegraphics[width=0.14\linewidth]{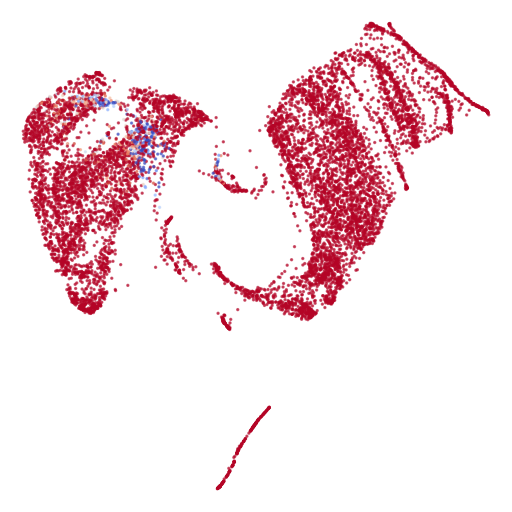} \\
        \makebox[0.02\linewidth][c]{\rotatebox{90}{\scriptsize \hspace*{3em}LAOF}} &
        \includegraphics[width=0.14\linewidth]{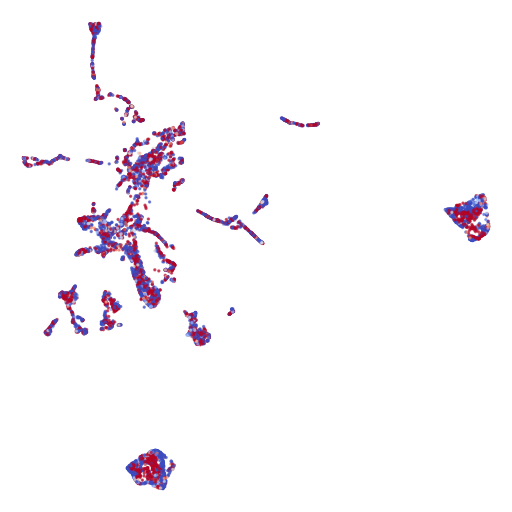} &
        \includegraphics[width=0.14\linewidth]{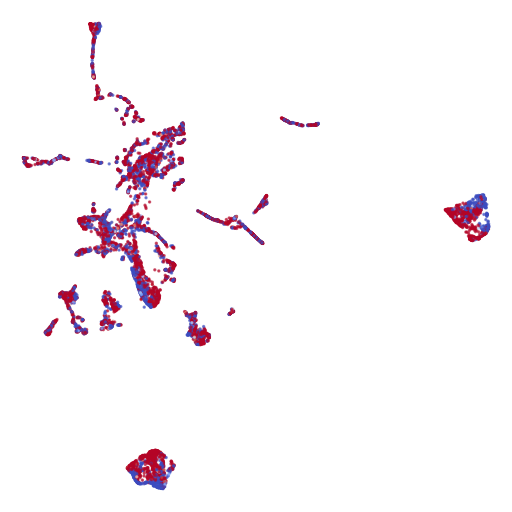} &
        \includegraphics[width=0.14\linewidth]{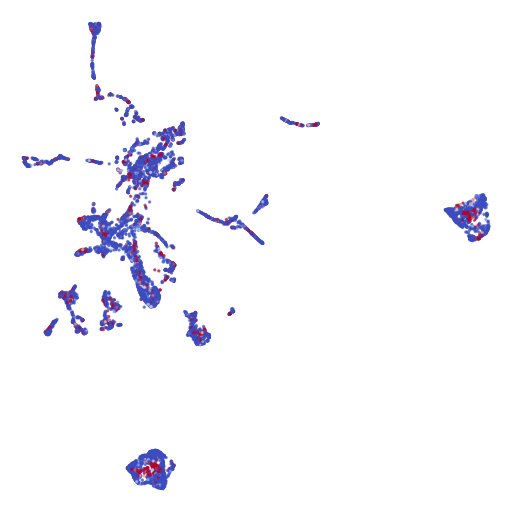} &
        \includegraphics[width=0.14\linewidth]{figures/umap_analysis/umap_laof_4.png} &
        \includegraphics[width=0.14\linewidth]{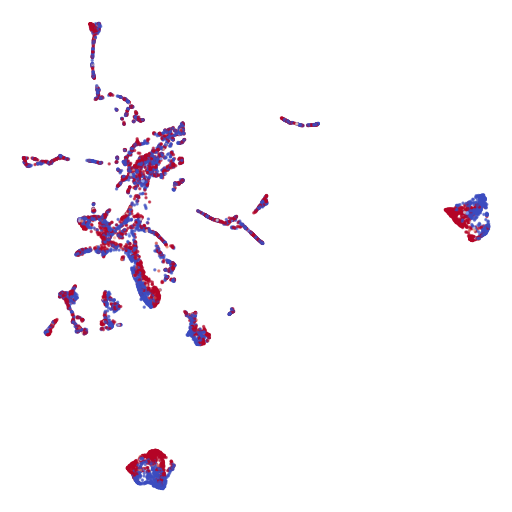} &
        \includegraphics[width=0.14\linewidth]{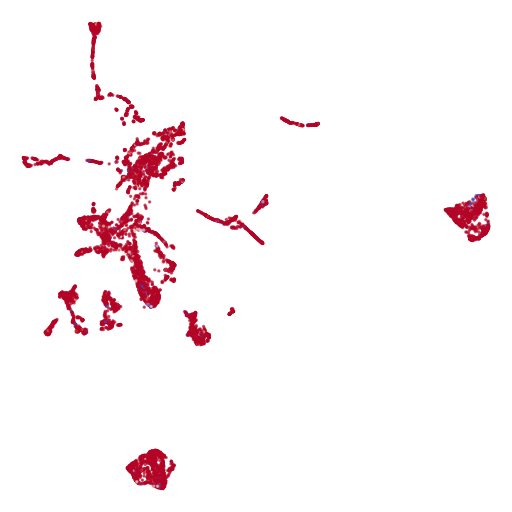} \\
        \makebox[0.02\linewidth][c]{\rotatebox{90}{\scriptsize \hspace*{1.5em}LAPO-Slots}} &
        \includegraphics[width=0.14\linewidth]{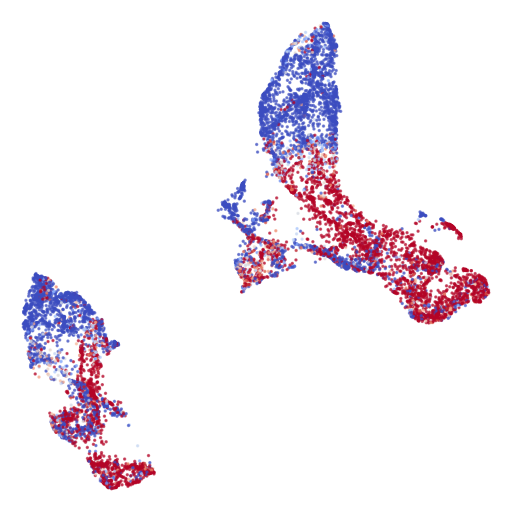} &
        \includegraphics[width=0.14\linewidth]{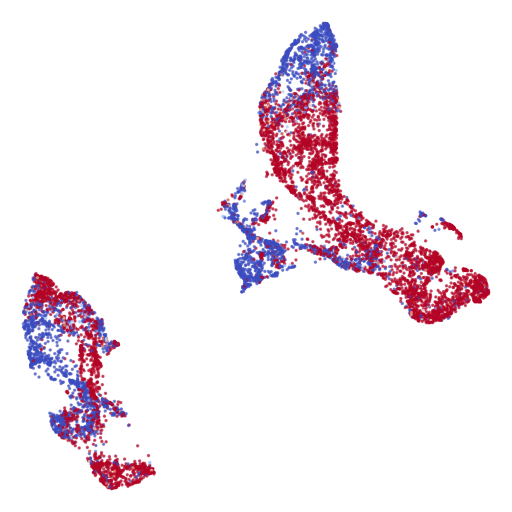} &
        \includegraphics[width=0.14\linewidth]{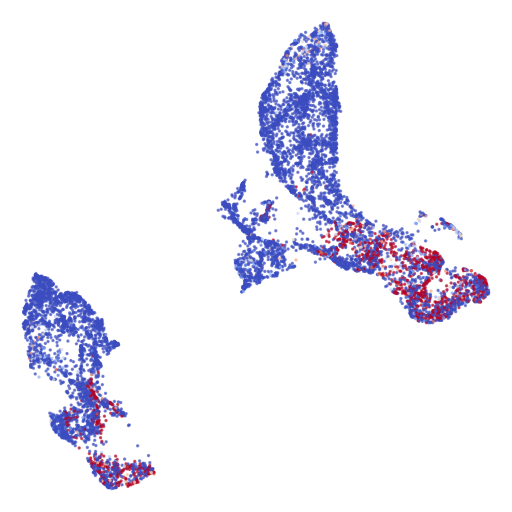} &
        \includegraphics[width=0.14\linewidth]{figures/umap_analysis/umap_oc-lapo_4.png} &
        \includegraphics[width=0.14\linewidth]{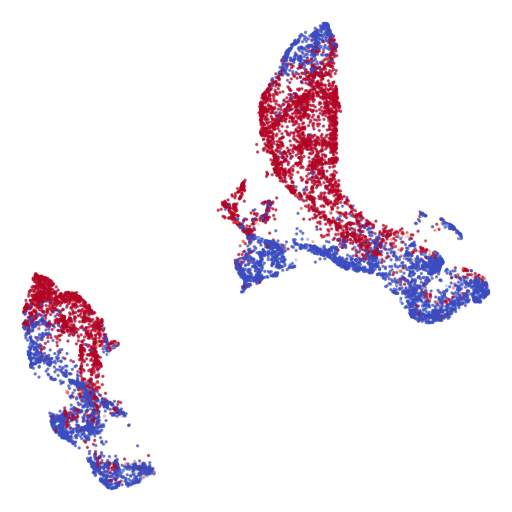} &
        \includegraphics[width=0.14\linewidth]{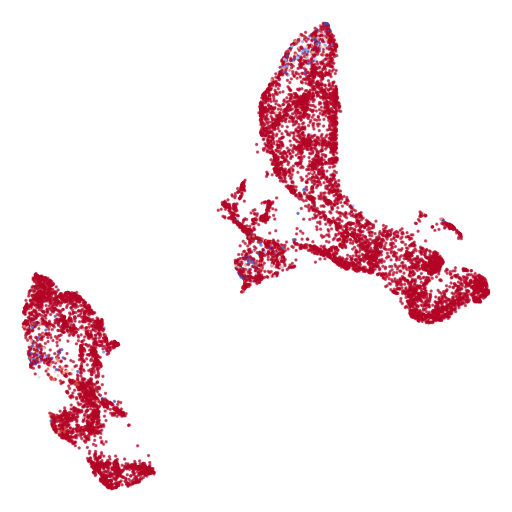} \\
        \makebox[0.02\linewidth][c]{\rotatebox{90}{\scriptsize \hspace*{2em}MaskLAM}} &
        \includegraphics[width=0.14\linewidth]{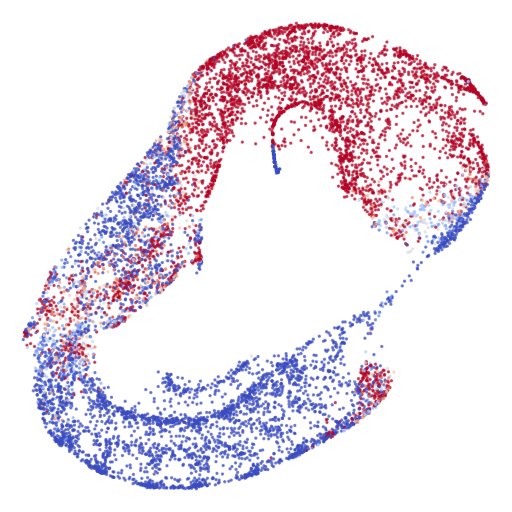} &
        \includegraphics[width=0.14\linewidth]{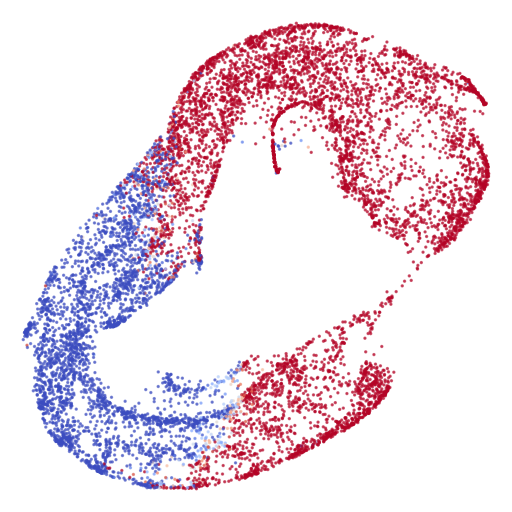} &
        \includegraphics[width=0.14\linewidth]{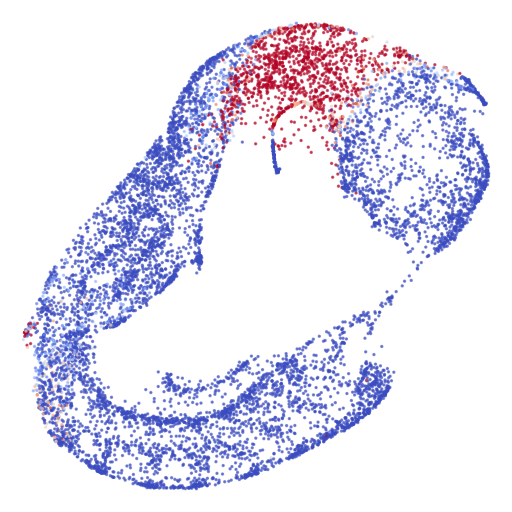} &
        \includegraphics[width=0.14\linewidth]{figures/umap_analysis/umap_slapo_4.png} &
        \includegraphics[width=0.14\linewidth]{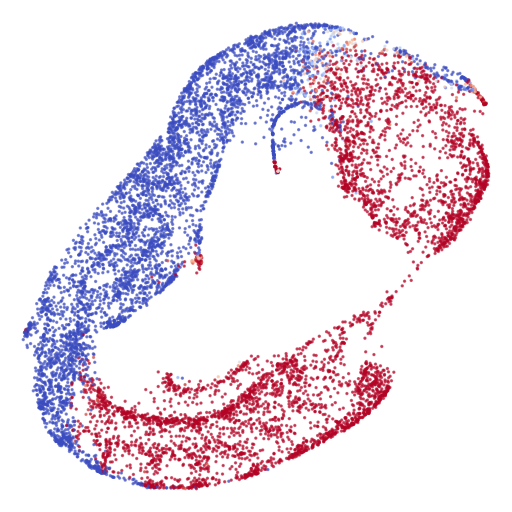} &
        \includegraphics[width=0.14\linewidth]{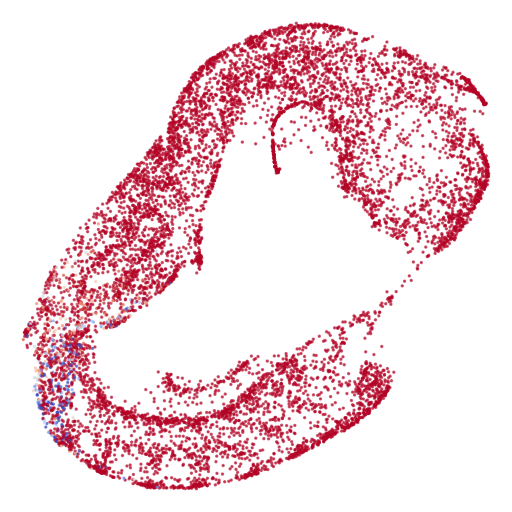} \\
        &
        \makebox[0.14\linewidth][c]{\scriptsize Action dimension 0} &
        \makebox[0.14\linewidth][c]{\scriptsize Action dimension 1} &
        \makebox[0.14\linewidth][c]{\scriptsize Action dimension 2} &
        \makebox[0.14\linewidth][c]{\scriptsize Action dimension 3} &
        \makebox[0.14\linewidth][c]{\scriptsize Action dimension 4} &
        \makebox[0.14\linewidth][c]{\scriptsize Action dimension 5}
    \end{tabular*}
    \vspace{2pt}

    \makebox[\linewidth][c]{%
    \begin{tikzpicture}
        \shade[left color=blue!70,right color=red!70,middle color=white] (0,0) rectangle (0.42\linewidth,0.018\linewidth);
        \node[font=\scriptsize] at (0.02\linewidth,0.009\linewidth) {-1};
        \node[font=\scriptsize] at (0.21\linewidth,0.009\linewidth) {0};
        \node[font=\scriptsize] at (0.40\linewidth,0.009\linewidth) {1};
    \end{tikzpicture}
    }

    \caption{\textbf{UMAP projections of learned latent actions on the DCS (distractor) dataset.} Rows show LAPO, LAOM-Labels, LAOF, LAPO-Slots, and MaskLAM, respectively; columns show action dimensions 0--5.}
    \label{fig:umap_dcs_distractor}
\end{figure}

\Cref{fig:umap_dcs_distractor} extends the single-dimension comparison of \Cref{fig:umap_action_dim4_methods} to all six action coordinates of cheetah-run on DCS (distractor). Each row corresponds to one method and each column to the ground-truth action dimension used to color the projection; a well-aligned latent exhibits a smooth color gradient along the manifold for every column.

\textbf{MaskLAM recovers a coherent gradient on every action dimension.} MaskLAM's manifold shows a clean red-to-blue transition for all six coordinates, with the gradient direction rotating from column to column as expected from a latent that is linearly aligned with the action vector. The per-dimension consistency confirms that the single-dimension view in the main text is representative rather than cherry-picked.

\textbf{LAOM-Labels matches the structure using action supervision.} LAOM-Labels also forms a connected manifold with visible color separation on every dimension, consistent with its use of ground-truth actions as an auxiliary supervisory signal during pre-training (\Cref{sec:experiments}). MaskLAM reaches comparable per-dimension structure without consuming any action labels at the latent action learning stage.

\textbf{LAPO and LAOF collapse into entangled fragments with no action structure.} LAPO's projection breaks into thin disjoint strands with no visible gradient on any dimension, and LAOF mirrors this pattern at finer fragmentation. Under distractors, the unmasked reconstruction loss steers both latents toward exogenous variance instead of agent-controlled dynamics (\Cref{sec:results}), and the per-dimension view shows this happens uniformly across the action vector rather than on isolated coordinates.

\textbf{LAPO-Slots recovers partial structure on a subset of dimensions.} LAPO-Slots produces a single connected manifold with locally visible red-blue regions on some coordinates, but the gradient is non-monotonic or washes out on the remaining ones. Object-centric decomposition isolates the agent at the input, yet the unmasked reconstruction loss still allocates the latent's capacity unevenly across action coordinates.

The per-dimension picture matches the aggregate probe error reported in \Cref{tab:q1_combined_aggregate}: only MaskLAM and LAOM-Labels recover a latent action space that is linearly decodable to the full action vector, and MaskLAM does so without any ground-truth action supervision during pre-training.

\section{Dataset Generation}
\label{sec:appendix_dataset}

We construct one offline dataset per benchmark and reuse it for every method in this work. To isolate the effect of visual distractors, we generate \emph{synchronized} pairs of distractor-free and distracting trajectories: the underlying agent behavior and environment state are identical in both settings, so any gap in latent action quality or downstream return must come from how a method handles the visual distractors rather than from a different action distribution.

\textbf{Expert policies.} On DCS we reuse the released expert checkpoints of LAOM~\citep{nikulin_latent_2025} rather than retraining our own. Briefly, those checkpoints are PPO~\citep{schulman_proximal_2017} from CleanRL~\citep{huang2022cleanrl} for cheetah-run, walker-run, and hopper-hop, and SAC~\citep{haarnoja_soft_2018} from Stable-Baselines3~\citep{stable-baselines3} for humanoid-walk, all trained on proprioceptive states with default hyperparameters; we refer the reader to~\citep{nikulin_latent_2025} for the full training schedule. On DMW we use the released single-task SAC experts of Meta-World~\citep{yu_meta-world_2020}, one per task, trained with the Garage configuration of $500$ epochs $\times$ $40$ epoch cycles $\times$ $500$ gradient steps per cycle, corresponding to roughly $10$M environment steps per task; full hyperparameters are reported in~\citep{yu_meta-world_2020}. Across both benchmarks the experts act on proprioceptive states, not on rendered images, so visual distractors cannot influence their decisions and the resulting action sequences are deterministic functions of the seed and initial state.

\textbf{Trajectory collection.} We roll out each expert checkpoint and render every state twice: once with a clean background and once with the distractor configuration of~\Cref{sec:experiments}, sharing the camera pose, agent state, and time step across both renders. Because the expert ignores pixels, the two renders differ only in the observation channel and not in the recorded action or the next state, yielding the synchronized distractor-free / distracting pair described above. The collected datasets contain 9M training and 1M held-out transitions on DCS, and 1M training and 100k held-out transitions on DMW (\Cref{sec:experiments}). For each transition we store the rendered observation, the ground-truth simulator mask, the proprioceptive state, and the executed action.

\textbf{Distractor videos.} The dynamic backgrounds are sampled from DAVIS~\citep{pont-tuset_2017_2018}. We assign DAVIS clips to our train and test splits disjointly: training trajectories sample backgrounds only from the DAVIS train split, and held-out evaluation trajectories sample backgrounds only from the DAVIS test split. This is intentionally stricter than the standard protocol in LAOM, which reuses the DAVIS clips seen during training in its in-distribution evaluation and reports a separate, harder ``OOD'' number on held-out clips. By construction, every evaluation in this paper corresponds to that OOD setting, so our reported metrics measure generalization to unseen distractor videos rather than memorization of the training backgrounds.

\textbf{Masks.} Ground-truth segmentation masks are read directly from the simulator at render time, so they are tightly aligned with the rendered observation and require no additional processing. SAM~2.1 masks~\citep{ravi_sam_2024} are produced in a separate post-processing pass over the released observations using the prompt-based extraction protocol detailed in~\Cref{sec:mask_generation}. All datasets, including ground-truth and SAM masks, will be released upon acceptance.

\section{Mask Generation}
\label{sec:mask_generation}

\begin{figure}[H]
    \centering
    \setlength{\tabcolsep}{1pt}
    \begin{tabular*}{\linewidth}{@{\extracolsep{\fill}}lcccc@{}}
        \makebox[0.035\linewidth][c]{\rotatebox{90}{\scriptsize \hspace*{5em}Original}} &
        \includegraphics[width=0.22\linewidth]{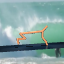} &
        \includegraphics[width=0.22\linewidth]{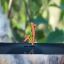} &
        \includegraphics[width=0.22\linewidth]{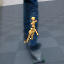} &
        \includegraphics[width=0.22\linewidth]{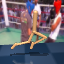} \\
        \makebox[0.035\linewidth][c]{\rotatebox{90}{\scriptsize  \hspace*{5em}Prompt}} &
        \includegraphics[width=0.22\linewidth]{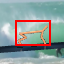} &
        \includegraphics[width=0.22\linewidth]{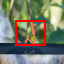} &
        \includegraphics[width=0.22\linewidth]{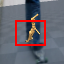} &
        \includegraphics[width=0.22\linewidth]{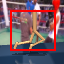} \\
        \makebox[0.035\linewidth][c]{\rotatebox{90}{\scriptsize  \hspace*{5em}Mask}} &
        \includegraphics[width=0.22\linewidth]{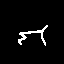} &
        \includegraphics[width=0.22\linewidth]{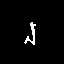} &
        \includegraphics[width=0.22\linewidth]{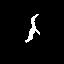} &
        \includegraphics[width=0.22\linewidth]{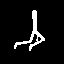} \\
        &
        \makebox[0.22\linewidth][c]{\scriptsize cheetah-run} &
        \makebox[0.22\linewidth][c]{\scriptsize hopper-hop} &
        \makebox[0.22\linewidth][c]{\scriptsize humanoid-walk} &
        \makebox[0.22\linewidth][c]{\scriptsize walker-run}
    \end{tabular*}
    \caption{\textbf{SAM~2.1 mask generation on DCS.} Each column shows one task (cheetah-run, hopper-hop, humanoid-walk, walker-run); rows show, top to bottom, the rendered observation, the same observation with the fixed per-task bounding-box prompt overlaid, and the segmentation mask propagated by the SAM~2.1 video tracker from that single first-frame prompt. A single coarse box per task is sufficient to recover a tight agent mask across all four environments despite dynamic backgrounds, agent color randomization, and camera shake.}
    \label{fig:mask_generation_examples}
\end{figure}

We pre-compute SAM segmentation masks for every trajectory in both benchmarks using the SAM~2.1-hiera-tiny video model~\citep{ravi_sam_2024}. To avoid per-frame annotation, we exploit the deterministic initialization of each environment: the agent always starts at a known location in the camera frame, so a single coarse, fixed-size bounding box on the first frame of a trajectory is sufficient to identify it. We feed this bounding box as the initial prompt to SAM~2.1 and let the video tracker propagate the mask through the remaining frames. The same procedure is applied to DCS and DMW; per-task bounding boxes are reused across all trajectories of that task. \Cref{fig:mask_generation_examples} illustrates the prompt and the resulting segmentation for the four DCS environments.

At the reported throughput of $91.5$ FPS for the 38.9M-parameter tiny variant on an NVIDIA A100~\citep{ravi_sam_2024}, annotating the full corpus takes approximately $309$ GPU-hours on a single A100. SAM masks are stored alongside the simulator ground-truth masks in the released datasets so that any method can be evaluated under either mask source without re-running segmentation.

%%%%%%%%%%%%%%%%%%%%%%%%%%%%%%%%%%%%%%%%%%%%%%%%%%%%%%%%%%%%
% \FloatBarrier
% \newpage
% \input{checklist.tex}

\end{document}